\def\year{2022}\relax
\documentclass[letterpaper]{article} 
\usepackage{aaai22}  
\usepackage{times}  
\usepackage{helvet}  
\usepackage{courier}  
\usepackage[hyphens]{url}  
\usepackage{graphicx} 
\usepackage{natbib}  
\usepackage{caption} 
\DeclareCaptionStyle{ruled}{labelfont=normalfont,labelsep=colon,strut=off} 
\usepackage{algorithm}
\usepackage{algorithmic}

\usepackage{newfloat}
\usepackage{listings}
\floatstyle{ruled}
\newfloat{listing}{tb}{lst}{}
\floatname{listing}{Listing}
\usepackage{amsmath,amssymb}		
\usepackage{subfigure}		
\usepackage{hyperref}

\usepackage{graphicx}		
\usepackage{wrapfig}		
\usepackage{subfigure}		
\usepackage{color}			
\usepackage{multicol}		
\usepackage{multirow} 		
\usepackage{booktabs} 		
\usepackage{color}			
\usepackage{xcolor}         

\newtheorem{theorem}{Theorem}
\newtheorem{assumption}{Assumption}

\title{Learn Goal-Conditioned Policy with Intrinsic Motivation \\for Deep Reinforcement Learning}
\author{
    Jinxin Liu\textsuperscript{\rm 123} \ \  Donglin Wang\textsuperscript{\rm 23}\thanks{Corresponding author.} \ \ Qiangxing Tian\textsuperscript{\rm 123} \ \  Zhengyu Chen\textsuperscript{\rm 123} 
}
\affiliations{
    \textsuperscript{\rm 1}Zhejiang University. \textsuperscript{\rm 2}Westlake University. \textsuperscript{\rm 3}Westlake Institute for Advanced Study. \\
    \{liujinxin, wangdonglin, tianqiangxing, chenzhengyu\}@westlake.edu.cn 
}

\usepackage{bibentry}

\begin{document}

\maketitle

\begin{abstract}
It is of significance for an agent to autonomously explore the environment and learn a widely applicable and general-purpose goal-conditioned policy that can achieve diverse goals including images and text descriptions.  Considering such perceptually-specific goals, one natural approach is to reward the agent with a prior non-parametric distance over the embedding spaces of states and goals. However, this may be infeasible in some situations, either because it is unclear how to choose suitable measurement, or because embedding (heterogeneous) goals and states is non-trivial. The key insight of this work is that we introduce a latent-conditioned policy to provide goals and intrinsic rewards for learning the goal-conditioned policy. 
As opposed to directly scoring current states with regards to goals, we obtain rewards by scoring current states with associated latent variables. We theoretically characterize the connection between our unsupervised objective and the multi-goal setting, and empirically demonstrate the effectiveness of our proposed  method which substantially outperforms prior techniques in a variety of tasks. 
\end{abstract}

\section{Introduction}
Deep reinforcement learning (RL) makes it possible to drive agents to achieve sophisticated goals in complex and uncertain environments, from computer games~\citep{badia2020agent57, berner2019dota} to real robot control~\citep{lee2018composing, lowrey2018plan, vecerik2019practical, popov2017data}, which usually 
involves learning a specific policy for individual task relying on hand-specifying reward function. 
However, autonomous agents are expected to exist persistently in the world and have the ability to reach diverse goals.  
To achieve this, one needs to design a mechanism to spontaneously generate diverse goals and the associated rewards, over which the goal-conditioned policy is trained.

Based on the space of goal manifold, previous works can be divided into two categories: \emph{perceptually-specific goal based approaches} and \emph{latent variable based methods}. 
In the former, previous approaches normally assume the spaces of perceptual goals and states are same, and sample goals from the historical trajectories of the policy to be trained. 
It is convenient to use a prior non-parametric measure function, such~as L2 norm, to provide rewards (current states vs. goals) over the state space or the embedding space~\citep{higgins2017darla, nair2018visual, sermanet2018time, warde2018unsupervised}. 
However, these approaches taking the prior non-parametric measure function may limit the repertoires of behaviors and impose manual engineering burdens. 

On the contrary, 
\emph{latent variable based methods} assume that goals (latent variables) and states come from different spaces and the distribution of goals (latent variables) is known a priori. 
In parallel, such methods autonomously learn a reward function and {a latent-conditioned policy} through the lens of empowerment~\cite{salge2014empowerment, eysenbach2018diversity, sharma2019dynamics}. 
However, such policy is conditioned on latent variables rather than perceptually-specific goals. 
Applying this procedure  to goal-reaching tasks, similar to the parameter initialization or hierarchical RL, needs an external reward function for new tasks; 
otherwise the learned latent-conditioned policy cannot be applied directly to perceptually-specific goals.

In this paper, we incorporate a latent variable based objective into the perceptual goal-reaching tasks. Specifically, we decouple the task generation (including perceptual goals and associated reward functions) and goal-conditioned policy optimization, which are often intertwined in prior approaches. 
For the task generation, we employ a latent variable based objective~\citep{eysenbach2018diversity} to learn a latent-conditioned policy, run to generate goals, and a discriminator, serve as the reward function. 
Then our goal-conditioned policy is rewarded by the discriminator to imitate the trajectories, relabeled as goals, induced by the latent-conditioned policy. This procedure enables the acquired discriminator as a proxy to reward the 
goal-conditioned policy for various relabeled goals. \emph{In essence, the latent-conditioned policy can reproducibly influence the environment, and the goal-conditioned policy perceptibly imitates these influences.}

The main contribution of our work is an unsupervised RL method that can learn a perceptual goal-conditioned policy via intrinsic motivation (GPIM). 
Our training procedure decouples~the~task~(goals and rewards) generation and policy optimization, which makes the obtained reward~function universal and effective for various relabeled goals, including images and texts. 
We formally analyze the effectiveness of our relabeling procedure,{\tiny ~}and empirically find that our intrinsic reward is well shaped 
by the environment's dynamics and as a result benefits the training efficiency on extensive~tasks. 

\section{Preliminaries}

The goal in a reinforcement learning  problem is to maximize the expected return in a Markov decision process (MDP) $\mathcal{M}$, defined by the tuple $(S, A, p, r, \gamma)$, where $S$ and $A$ are state and action spaces, $p(s_{t+1} | {s_t}, {a_t})$ gives the next-state  distribution upon taking action ${a_t}$ in state ${s_t}$, $r({s_t}, {a_t}, {s_{t+1}})$ is the reward received at transition ${s_t} \stackrel{{a_t}}{\to} {s_{t+1}}$, and $\gamma$ is a discount factor. The objective is to learn the policy $\pi_\theta(a_t|s_t)$ by maximizing $\mathbb{E}_{p(\tau;\theta)}\left[ R(\tau) \right] = \mathbb{E}_{p(\tau;\theta)}\left[ \sum_t \gamma^t r(s_t,a_t,s_{t+1}) \right]$, where $p(\tau;\theta)$ denotes the induced trajectories by policy $\pi_\theta$ in the environment: $p(\tau;\theta) = p(s_0)\cdot \prod_{t=0}^{t=T-1}\pi_\theta(a_t|s_t)p(s_{t+1}|s_t,a_t)$.

Multi-goal RL augments the above optimization with a goal $g$ by learning a goal-conditioned policy $\pi_\theta(a_t|s_t,g)$ and optimizing  $\mathbb{E}_{p(g)}\mathbb{E}_{p(\tau|g;\theta)}\left[ R(\tau) \right]$ with reward  $r(s_t,a_t,s_{t+1},g)$. Such optimization can also be interpreted as a form of mutual information between the goal $g$ and agent's trajectory $\tau$ \citep{warde2018unsupervised}:
\begin{equation}
\max \mathcal{I}(\tau;g) = \mathbb{E}_{p(g)p(\tau|g;\theta)} \left[ \log p({g}|{\tau}) - \log p({g})\right].
\end{equation}
If $\log p(g|{\tau})$ is unknown and the goal $g$ is a latent variable, latent variable based models normally maximize the mutual information between the latent variable $\omega$ and agent's behavior $b$, and lower-bound this mutual information by approximating the posterior $p(\omega|b)$ with a learned  $q_\phi(\omega|b)$:  
$\mathcal{I}(b;\omega) \geq \mathbb{E}_{p(\omega, b;\mu)} \left[ \log q_\phi({\omega}|{b}) - \log p({\omega})\right]$, 
where the specific manifestation of agent's behavior $b$ can be an entire trajectory $\tau$, an individual state $s$ or a final state $s_T$. It is thus applicable to train $\pi_\mu(a_t|s_t,\omega)$ with learned $q_\phi$  (as~reward).

Several prior works have sought to incorporate the latent-conditioned $\pi_\mu(a_t|s_t,\omega)$ (as low-level skills) into hierarchical RL \citep{zhang2021hierarchical} or reuse the learned $q_\phi$ (as predefined tasks) in meta-RL~\citep{gupta2018unsupervised}, 
while we claim to reuse \emph{both} $\pi_\mu$ and $q_\phi$ with our relabeling procedure.

\begin{figure}[t]
	\centering
	\includegraphics[width=\columnwidth]{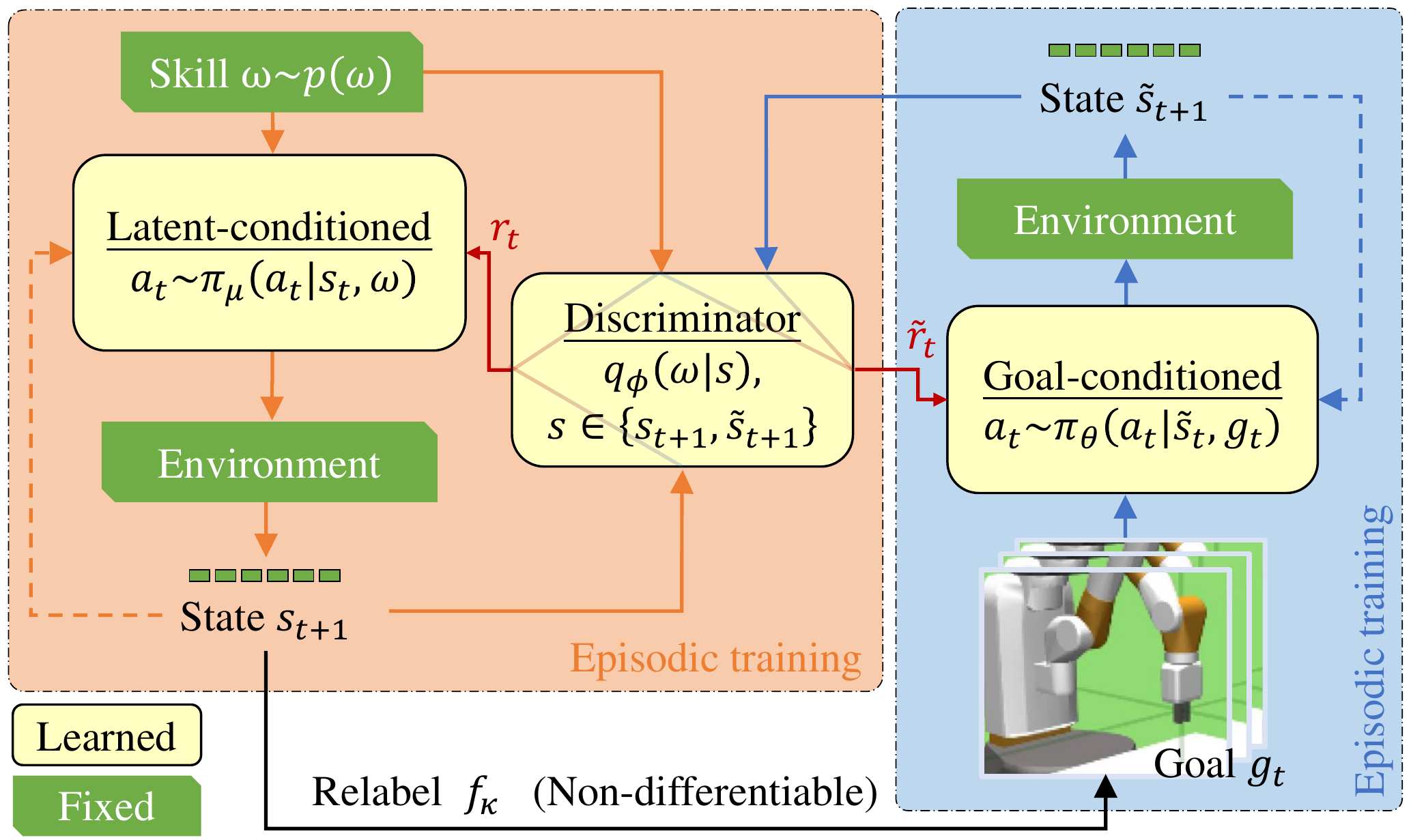}
	\caption{Framework of GPIM. We jointly train 
		the latent-conditioned policy $\pi_\mu$ and the discriminator $q_\phi$ to understand skills which specify task objectives (e.g., trajectories, the final goal state), and use such understanding to reward the  goal-conditioned policy $\pi_\theta$ for completing such tasks (relabeled states).
		In this diagram, state ${s}_{t+1}$ (e.g., joints)~induced by $\pi_\mu$ is converted into the perceptually-specific goal $g_t$ (e.g., images or text descriptions) for~$\pi_\theta$. 
		Note that the two environments above are same, and the initial states ${s}_0$ of $\pi_\mu$ and $\tilde{s}_0$ of $\pi_\theta$ are sampled from the same (fixed) distribution.
	}
	\label{framework}
\end{figure}

\section{The Method}
In this section, we first formalize the problem and introduce the framework. 
Second, we illustrate our~GPIM objective and elaborate on the process of how to jointly learn the latent-conditioned policy and a goal-conditioned policy. 
Third, we formally verify our (unsupervised) objective and understand how GPIM relates to the standard multi-goal RL. 


\subsection{Overview}
\label{ps}

As shown in Figure~\ref{framework} (right), our objective is to learn a goal-conditioned policy $\pi_\theta \left( a |\tilde{s}, g \right)$ that inputs state $\tilde{s}$ and perceptually-specific goal $g$ and outputs action $a$. 
To efficiently generate tasks for training the goal-conditioned policy $\pi_\theta$, we introduce another latent-conditioned policy $\pi_\mu(a|s, {\omega})$, which takes as input a state $s$ and a latent variable $\omega$ and outputs action $a$ to generate goals, and the associated discriminator $q_\phi$ (i.e., generating tasks). 
Additionally, we assume that we have access to a procedural relabeling function $f_\kappa$ (we will discuss this assumption latter), which can relabel states $s$ as goals $g$ for training $\pi_\theta$. 
On this basis, $\pi_\theta \left( a |\tilde{s}, g \right)$ conditioned on the relabeled goal $g$ interacts with the reset environment under the instruction of the associated $q_\phi$.  
We use the non-tilde $s$ and the tilde $\tilde{s}$ to distinguish between the states of two policies respectively. 
Actually, $\tilde{s}$ and $s$ come from the same state space. 
In the following, if not specified, goal $g$ refers to the perceptually-specific goal, and no longer includes the case that goal is a latent variable. 

To ensure the generated tasks (by $\pi_\mu$) are reachable for $\pi_\theta$, we explicitly make the following assumption\footnote{In appendix, we empirically find the assumption can be lifted.}: 
\begin{assumption}
	\label{assumption-relabel-given}
	{The initial state of the environment is fixed.} 
\end{assumption}

\subsection{Proposed GPIM Method}
\label{trcp}

\begin{figure}[h]
	\centering
	\begin{center}
		\includegraphics[width=0.678\columnwidth]{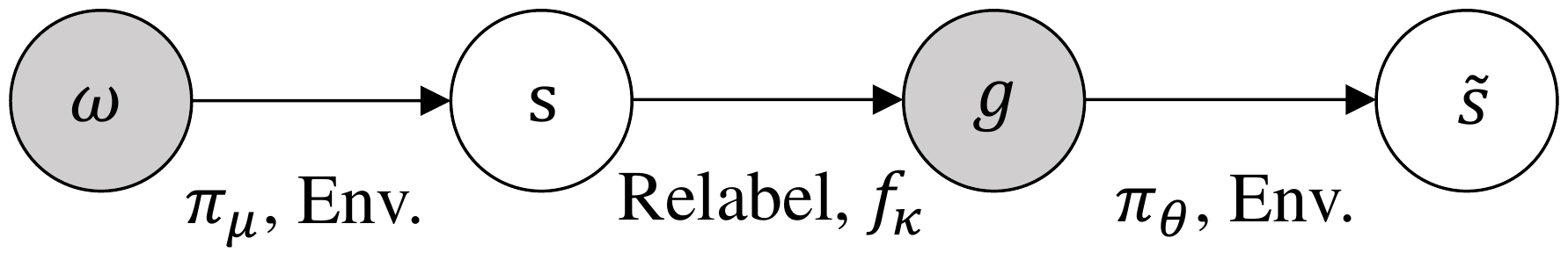}
	\end{center}
	\caption{Latent-conditioned policy $\pi_\mu$ provides goals and the associated reward for the goal-conditioned policy $\pi_\theta$.}
	\label{graphicalmodel}
\end{figure}

In order to jointly learn the latent-conditioned  $\pi_\mu(a|s, {\omega})$ and goal-conditioned  $\pi_\theta \left( a |\tilde{s}, g \right)$, we maximize the mutual information between the state $s$ and latent variable $w$ for $\pi_\mu$, and simultaneously maximize the mutual information between the state $\tilde{s}$ and goal $g$ for $\pi_\theta$. Consequently, the overall objective  to be maximized can be expressed as follows
\footnote{To further clarify the motivation, we conduct the ablation study to compare our method (maximizing $\mathcal{I}({s}; {\omega}) + \mathcal{I}({\tilde{s}}; g)$) with that just maximizing $\mathcal{I}(\tilde{s};\omega)$ and that maximizing $\mathcal{I}(\tilde{s};g)$ in  appendix.} 
\begin{equation}\label{eq3}
\mathcal{F}(\mu, \theta) = \mathcal{I}({s}; {\omega}) + \mathcal{I}({\tilde{s}}; g).
\end{equation}
For clarification, Figure~\ref{graphicalmodel} depicts the graphical model for the latent variable $\omega$, state $s$ induced by  $\pi_\mu$, goal $g$ relabeled from $s$, and state $\tilde{s}$ induced by $\pi_\theta$.
As seen, the latent variable ${\omega} \sim p({\omega})$ is firstly used to generate state ${s}$ via the policy $\pi_\mu$ interacting with the environment. 
Then, we relabel the generated state $s$ to goal $g$. 
After that, $\pi_\theta$ conditioned on $g$ interacts with the environment to obtain the state $\tilde{s}$ at another episode. 
In particular, $\pi_\mu$ is expected to generate diverse behavior modes by maximizing $\mathcal{I}({s}; {\omega})$,  while $\pi_\theta$ behaving like $\tilde{s}$ is to "imitate" (see next $q_\phi$) these behaviors by taking as input the relabeled goal (indicated by  Figure~\ref{graphicalmodel}). 

Based on the context, the correlation between $\tilde{s}$ and $g$ is no less than that between  $\tilde{s}$ and $w$: $ \mathcal{I}({\tilde{s}}; g) \geq \mathcal{I}({\tilde{s}}; {\omega})$ \citep{beaudry2011intuitive}. 
Thus, we can obtain the lower bound:
\begin{align}\label{eqs45}
\mathcal{F}(\mu, \theta) &\geq \mathcal{I}({s}; {\omega}) + \mathcal{I}({\tilde{s}}; {\omega})  \\
&= 2\mathcal{H}({\omega}) + \mathbb{E}_{p_{m}(\cdot)} \left[ \log p({\omega}|{s}) + \log p({\omega}|{\tilde{s}})\right], \nonumber
\end{align}
where $p_{m}(\cdot) \triangleq p(\omega, s, g, \tilde{s}; \mu, \kappa, \theta)$ denotes the joint distribution of $\omega$, $s$, $g$ and $\tilde{s}$ specified by the graphic model in Figure~\ref{graphicalmodel}.  
Since it is difficult to exactly compute the posterior distributions $p({\omega}|{s})$ and $p({\omega}|{\tilde{s}})$, Jensen's Inequality~\citep{barber2003algorithm} is further applied for approximation by using a learned discriminator network $q_\phi({\omega}|\cdot)$. Thus, we have $\mathcal{F}(\mu, \theta)  \geq 
\mathcal{J}(\mu, \phi, \theta)$, where 
\begin{align}
\mathcal{J}(\mu, \phi, \theta) 
\triangleq 
2\mathcal{H}({\omega}) + 
\mathbb{E}_{p_{m}(\cdot)} \left[\log q_\phi({\omega}|{s})  + \log q_\phi({\omega}|{\tilde{s}}) \right]. \nonumber
\end{align}
It is worth noting that the identical discriminator $q_\phi$ is used for the variational approximation of $p({\omega}|{s})$ and $p({\omega}|{\tilde{s}})$. 
For the state $s$ induced by skill $\pi_\mu(\cdot|\cdot,\omega)$ and $\tilde{s}$ originating from $\pi_\theta(\cdot|\cdot,g)$, the shared discriminator $q_\phi$ assigns a similarly high probability on $\omega$ for both states $s$ and $\tilde{s}$ associated with the same $\omega$. 
Therefore, $q_\phi$ can be regarded as a reward network shared by the latent-conditioned $\pi_\mu$ and goal-conditioned $\pi_\theta$. 
Intuitively, we factorize acquiring the goal-conditioned policy $\pi_\theta$ and learn it purely in the space of the agent's embodiment (i.e., the latent $\omega$) --- separate from the perceptually-specific goal $g$ (e.g., states, images and texts), where the latent $\omega$ and the perceptual goal $g$ have different characteristics due to the underlying manifold~spaces.

According to the surrogate objective $\mathcal{J}(\mu, \phi, \theta)$
, we propose an alternating optimization between $\pi_\mu$, $q_\phi$ and $\pi_\theta$: 

\noindent 
\textbf{Step I:} Fix $\pi_\theta$ and update $\pi_\mu$ and $q_\phi$. In this case, $\theta$ is not a variable to update and thus $\mathcal{J}(\mu, \phi, \theta)$ becomes
\begin{align}
\label{eq7}
\mathcal{J}(\mu, \phi)  =  &\  \mathbb{E}_{p(\omega, s; \mu)} \left[\log q_\phi({\omega}|{s})\right]  \nonumber \\
&+ \underbrace{\mathbb{E}_{p_{m}(\cdot)} \left[\log q_\phi({\omega}|{\tilde{s}}) - 2\log p({\omega}) \right]}_{\text{Variable independent term}}.
\end{align}
According to Equation~\ref{eq7}, $\pi_\mu$ can be thus optimized by setting the intrinsic reward at time step $t$ as
\begin{equation}\label{eq8}
r_t = \log q_\phi({\omega}|{s}_{t+1}) - \log p({\omega}), 
\end{equation}
where the term $- \log p({\omega})$ is added for agents to avoid artificial termination and reward-hacking issues~\citep{amodei2016concrete, eysenbach2018diversity}. We implement this optimization with SAC.
In parallel, the reward network (discriminator) $q_\phi$ can be updated with SGD by maximizing
\begin{equation}\label{eq-sgd-discriminator}
\mathbb{E}_{p(\omega)p(s|\omega; \mu)} \left[\log q_\phi({\omega}|{s})\right].
\end{equation}

\noindent
\textbf{Step II:} Fix $\pi_\mu$ and $q_\phi$ to update $\pi_\theta$. In this case, $\mu$ and $\phi$ are not variables to update,  and $\mathcal{J}(\mu, \phi, \theta)$ can be simplified~as
\begin{align}
\label{eq9}
\mathcal{J}(\theta)  = &\  \mathbb{E}_{p_{m}(\cdot)} \left[\log q_\phi({\omega}|\tilde{s})\right]  \nonumber \\ 
&+ \underbrace{\mathbb{E}_{p(\omega, s; \mu)} \left[\log q_\phi({\omega}|s) - 2\log p({\omega}) \right]}_{\text{Variable independent term}}.
\end{align}
According to Equation~\ref{eq9}, $\pi_\theta$ can thus be optimized  by setting the intrinsic reward at time step $t$ as
\begin{equation}\label{eq10}
\tilde{r}_t = \log q_\phi({\omega}|\tilde{{s}}_{t+1}) - \log p({\omega}), 
\end{equation}
where the term $- \log p({\omega})$ is added for the same reason as above and we also implement this~optimization with SAC. Note that we do not update $q_\phi$ with the data induced~by~$\pi_\theta$. 

\begin{algorithm}[t]
	\caption{Learning process of our proposed GPIM}
	\small  
	\label{alg:algorithm1}
		\begin{algorithmic}[1]
			\WHILE {not converged}
			\STATE {\emph{\# Step I:} generate goals and reward functions.}
			\STATE Sample the latent variable: ${\omega} \sim p({\omega})$. 
			\STATE Reset Env. \& sample initial state: ${s}_0 \sim p_0({s})$.
			\FOR { $t = 0, 1, ..., T-1$ steps }
			\STATE Sample action: $ {a}_t \sim \pi_\mu ({a}_t | {s}_t, {\omega}) $. 
			\STATE Step environment: $ {s}_{t+1} \sim p({s}_{t+1} | {s}_t,{a}_t)$.
			\STATE Relabel: ${g}_{t} = f_\kappa({s}_{t+1})$. \ \ \ \  {\emph{$\rhd$ Record.}} 
			\STATE Compute reward $r_t$ for policy $\pi_\mu$ using (\ref{eq8}).
			\STATE Update policy $\pi_\mu$ to maximize $r_t$ with SAC.
			\STATE Update  discriminator ($q_\phi$) to maximize (\ref{eq-sgd-discriminator}) with SGD.
			\ENDFOR
			\STATE {\emph{\# Step II:} $\pi_\theta$ imitates $\pi_\mu$ with the relabeled goals and the associated rewards (for the same $\omega$).}
			\STATE Reset Env. \& sample initial state: $\tilde{s}_0 \sim p_0(\tilde{s})$.
			\FOR { $t = 0, 1, ..., T-1$ steps }
			\STATE Recap dynamic (time-varying) goal $g_t$ from the recorded goals in \emph{line 8}. {\ \ \#  \emph{Note: $g_t = g_T$ for static (\emph{fixed}) goals.}}
			\STATE Sample action: $ {a}_t \sim \pi_\theta ({a}_t | \tilde{s}_t, {g}_{t})$.
			\STATE Step environment: $ {\tilde{s}}_{t+1} \sim p({\tilde{s}}_{t+1} | {\tilde{s}}_t,{a}_t)$.
			\STATE Compute reward $\tilde{r}_t$ for policy $\pi_\theta$ using (\ref{eq10}).
			\STATE Update policy $\pi_\theta$  to maximize $ \tilde{r}_t$ with SAC. 
			\ENDFOR
			\ENDWHILE
		\end{algorithmic}
\end{algorithm}

These two steps are performed alternately until convergence (see Algorithm~\ref{alg:algorithm1}). 
In summary, we train the goal-conditioned $\pi_\theta$ along with an extra latent-conditioned  $\pi_\mu$ an a procedural relabel function $f_\kappa$, which explicitly decouples the procedure of unsupervised RL into task generation (including goals and reward functions) and policy optimization. 

For clarity, we state three different settings for $f_\kappa$: 
(1) if $f_\kappa = q_\phi$, our objective is identical to the latent variable based models, maximizing $\mathcal{I}(\omega;s)$ to obtain the \emph{latent-conditioned} policy~\cite{eysenbach2018diversity}; 
(2) if $f_\kappa(s) = s$, this procedure is consistent with the hindsight relabeling~\cite{andrychowicz2017hindsight}; 
(3) if $f_\kappa(s)$ and $s$ have different spaces (not latent spaces), this relabeling is also a reasonable belief under the semi-supervised setting, e.g., the social partner in \citet{Colas2020LanguageAA}. 
In our experiment, we will consider (2) and (3) to learn $\pi_\theta(\cdot|\cdot,g)$ that is conditioned \emph{perceptually-specific} goals. 
For (3), it is easy to procedurally generate the image-based goals from (joint-based) states with the MuJoCo Physics Engine’s~\citep{emanuel2012mujoco} built-in renderer. 
Facing high-dimensional goals, we also incorporate a self-supervised loss over the perception-level~\citep{hafner2020action, xingyu2020dynamcis} for $\pi_\theta$. 



\subsection{Theoretical Analysis} 


Normally, multi-goal RL seeks the goal-conditioned policy $\pi_\theta(a|\tilde{s},g)$ that maximizes $\mathcal{I}(\tilde{s};g)$ with prior goal-distribution $p'(g)$ and the associated reward $p'(g|\tilde{s})$, while  GPIM learns $\pi_\theta(a|\tilde{s},g)$ by maximizing $\mathcal{I}({s}; {\omega}) + \mathcal{I}({\tilde{s}}; \omega)$ without any prior {goals and rewards}. Here, we~characterize~the~theoretical connection of returns between the two objectives {under deterministic $\pi_\mu$ and Assumption \ref{assumption-relabel-env}}.
\begin{assumption}
	\label{assumption-relabel-env}
	{The relabeling function $f_\kappa$ is bijective and the environment is deterministic.}
\end{assumption}
Let $\eta(\pi_\theta) \triangleq \mathcal{I}(\tilde{s};g) = \mathbb{E}_{p'(g,\tilde{s}; \theta)}\left[ \log p'(g|\tilde{s}) - \log p'(g) \right]$, 
where the expectation is taken over the rollout $p'(g,\tilde{s}; \theta)=p'(g)p(\tilde{s}|g;\theta)$,
and $\hat{\eta}(\pi_\theta) \triangleq \mathbb{E}_{{p^*_m(\cdot)}}\left[ \log p(\omega|\tilde{s}) - \log p(\omega) \right]$, 
where the joint distribution ${p^*_m(\cdot)} \triangleq p(\omega, s, g, \tilde{s}; {\mu^*}, \kappa, \theta)=p(\omega)p(s|\omega;{\mu^*})p(g|s;\kappa)p(\tilde{s}|g;\theta)$, {$\mu^* = \arg \max_{\mu} \mathcal{I}({s}; {\omega})$, and $\mathbb{E}_{p(\omega,s;\mu^*)}\log p(s|\omega;\mu^*)=0$.}
According our training procedure ($\pi_\mu$ is not affected by $\pi_\theta$) in Algorithm~\ref{alg:algorithm1}, it is trival to show that $\hat{\eta}(\pi_\theta)$ is a surrogate for our $\mathcal{I}({s}; {\omega}) + \mathcal{I}({\tilde{s}}; \omega)$ in Equation~\ref{eqs45}.  
We start by deriving that the standard multi-goal RL objective ${\eta}(\pi_\theta)$ and our (unsupervised) objective  ${\hat{\eta}(\pi_\theta)}$ {are equal} under some mild assumptions 
and then generalize this connection to a~general~case. 

\textbf{Special case:} We first assume the prior goal distribution $p'(g)$ for optimizing $\eta(\pi_\theta)$ matches the goal distribution {$\mathbb{E}_{\omega}\left[ p(g|\omega; \mu^*, \kappa) \right]$} induced by {$\pi_{\mu^*}$} and $f_\kappa$ for optimizing $\hat{\eta}(\pi_\theta)$. Then, we obtain: 
\begin{align}
&\ \quad \hat{\eta}(\pi_\theta) - \eta(\pi_\theta)  \nonumber \\
&= \mathbb{E}_{{p^*_{m}(\cdot)}}\left[ \log p(\omega|\tilde{s}) - \log p(\omega) - \log p'(g|\tilde{s}) + \log p'(g) \right]  \nonumber \\
&= \mathbb{E}_{{p^*_{m}(\cdot)}}\left[ \log p(\tilde{s}|\omega; {\mu^*}, \kappa, \theta) - \log p(\tilde{s}|g; \theta) \right] 
{= 0}. \label{equ-analysis}
\end{align}
Equation~\ref{equ-analysis} comes from our relabeling procedure {(with deterministic $\pi_{\mu^*}$ and Assumption~\ref{assumption-relabel-env})}, {specifying that $p(\tilde{s}|\omega; \mu^*, \kappa, \theta) = \mathbb{E}_{s,g}\left[ p(s|\omega; \mu^*) p(g|s;\kappa) p(\tilde{s}|g;\theta)\right] $, 
	$\mathbb{E}_{p^*_m(\cdot)}\left[ \log p(s|\omega; \mu^*) \right]=0$ and 	$\mathbb{E}_{p^*_m(\cdot)}\left[ \log p(g|s;\kappa) \right]=0$. 
	Essentially, this special case shows that without inductive bias on the self-generated goal distribution, our learning procedure leads to the desired goal-conditioned~policy~$\pi_\theta$.}


\textbf{General case:} 
Suppose that we do not have any prior connection between the goal distributions $p'(g)$ wrt optimizing $\eta(\pi_\theta)$ and the self-generated {$\mathbb{E}_{\omega}\left[ p(g|\omega; \mu^*, \kappa) \right]$} wrt optimizing $\hat{\eta}(\pi_\theta)$. 
The following theorem provides such a performance guarantee (Please see appendix for a full derivation): 
\begin{theorem}
	\label{theorem-generalized-case}
	Let ${\eta}(\pi_\theta)$ and $\hat{\eta}(\pi_\theta)$ be as defined above, and assume  relabeling $f_\kappa$ is bijective, then, 
	\begin{align*}
	{\hat{\eta}(\pi_\theta) - \eta(\pi_\theta) \leq 2 R_{max} \sqrt{\epsilon/2}}, 
	\end{align*}
	where $R_{max} = \max_{p'(g)p(\tilde{s}|g;\theta)} \log p'(g|\tilde{s}) - \log p'(g) $ 
	and $\epsilon = \mathbb{E}_{p(\omega)}\left[ D_{\text{KL}}(p(s|\omega;\mu^*) \Vert p'(s)) \right]$. 
\end{theorem}
This theorem implies that as long as we improve the return wrt $\hat{\eta}(\pi_\theta)$ by more than $2 R_{max} \sqrt{\epsilon/2}$, we can guarantee improvement wrt the return $\eta(\pi_\theta)$ of standard multi-goal RL. 



Note that the analysis presented above makes an implicit requirement that the goal distribution is valid for training $\pi_\theta$ (i.e., there is the corresponding target in the environment for the agent to pursue). This requirement is well satisfied for the multi-goal RL. However, ambiguity appears when the goal distribution {$\mathbb{E}_{\omega}\left[ p(g|\omega; \mu^*, \kappa) \right]$}, induced by {$\pi_{\mu^*}$} and $f_\kappa$, and the existed target $p'(g)$ in the reset environment for training $\pi_\theta$ have different supports. For example, the generated goal is "reaching red square", while such target does not exist in the reset environment for training $\pi_\theta$ (Algorithm~\ref{alg:algorithm1}~\emph{line~14}). 
Thus, we introduce relabeling over the environment~(see appendix) for granting valid training.

\section{Related Work}

\emph{Investigating the goal distribution:} 
For goal-reaching tasks, 
many prior methods \cite{schaul2015universal, andrychowicz2017hindsight, levy2017learning, pong2018temporal, hartikainen2019dynamical} assume an available distribution of goals during the exploration. 
In the unsupervised RL setting, the agent needs~to~automatically explore the environment and discover potential goals for learning the goal-conditioned policy. 
Several works 
\citep{colas2018gep, pere2018unsupervised, warde2018unsupervised, pong2019skew,  kovavc2020grimgep} also adopt heuristics to acquire the goal distribution based on previously~visited states, which is orthogonal to our relabeling~procedure. 

\emph{Learning the goal-achievement reward function:}  
Building on prior works in standard RL algorithms \cite{schaul2015universal, schulman2017proximal, haarnoja2018soft} that learn policies with prior goals and rewards, unsupervised RL faces another challenge --- automatically learning the goal-achievement reward function. 
Two common approaches to obtain rewards are (1) applying \emph{the pre-defined function} on the learned goal representations, and (2) \emph{directly learning a reward function}. 
Estimating the reward with \emph{the pre-defined function} typically assumes the goal space is the same as the state space, and learns the embeddings of states and goals with various auxiliary tasks (self-supervised loss in perception-level~\citep{hafner2020action}):  
\citet{sermanet2018time, warde2018unsupervised, liu2021return} employ the contrastive loss to acquire embeddings for high-dimensional inputs, and \citet{nair2018visual, florensa2019self, nair2019contextual, pong2019skew} elicit the features with the generative models. 
Over the learned representations, these approaches apply a~prior non-parametric measure function (e.g., the cosine similarity) to provide rewards. This contrasts with 
our decoupled training procedure, where we acquire rewards by scoring current states with their associated latent variables, instead of the perceptual goals. 
Such procedure provides more flexibility in training the goal-conditioned policy than using pre-defined measurements, especially for the heterogeneous states and~goals.

Another approach, \emph{directly learning a reward function}, aims to pursues skills (the latent-conditioned policy) by maximizing the empowerment \citep{salge2014empowerment}, which draws a connection between option discovery and information theory. 
This procedure \citep{achiam2018variational, eysenbach2018diversity, gregor2016variational, campos2020ExploreDA, sharma2019dynamics, tian2020learning} typically maximizes the mutual information between a latent variable and the induced behaviors (states or trajectories), which is optimized by introducing a latent-conditioned reward function. 
We explicitly relabels the states induced by the latent-conditioned policy and reuses the learned discriminator for~instructing~"imitation". 



\emph{Hindsight, self-play and knowledge distillation:} 
Our method is similar in spirit to goal relabeling methods like hindsight experience replay (HER)~\citep{andrychowicz2017hindsight} which replays each episode with a different goal in 
addition to the one the agent was trying to achieve. 
By contrast, our GPIM relabels the task for 
another policy while keeping behavior invariant. 
The self-play~\citep{sukhbaatar2017intrinsic, sukhbaatar2018learning} and knowledge~distillation~\citep{xu2020knowledge} are also related to our relabeling scheme, aiming to refine the training of one task with another associated task. 



\section{Experiments}

Extensive experiments are conducted to evaluate our proposed GPIM method,
where the following four questions will be considered in the main paper: 
(1) By using the "archery" task, we clarify whether $q_\phi$ can provide an effective reward function on learning the goal-conditioned policy $\pi_\theta$. 
Furthermore, more complex tasks including navigation, object manipulation, atari games, and mujoco tasks are introduced to answer:
(2) Does our model learn effective behaviors conditioned on a variety of goals (with different procedural relabeling $f_\kappa$), including high-dimensional images and text descriptions that are heterogeneous to states? 
(3) Does the proposed GPIM on learning the goal-conditioned policy outperform baselines? 
(4) Does the learned reward function produce better expressiveness of tasks, compared to the prior non-parametric function in the embedding space?  
For more experimental questions, analysis and results, see Appendix~\ref{appendix_additional_exp} and  \href{https://sites.google.com/view/gpim}{https://sites.google.com/view/gpim} (video). 




\begin{figure}[h]
	\centering
	\includegraphics[width=0.95\linewidth]{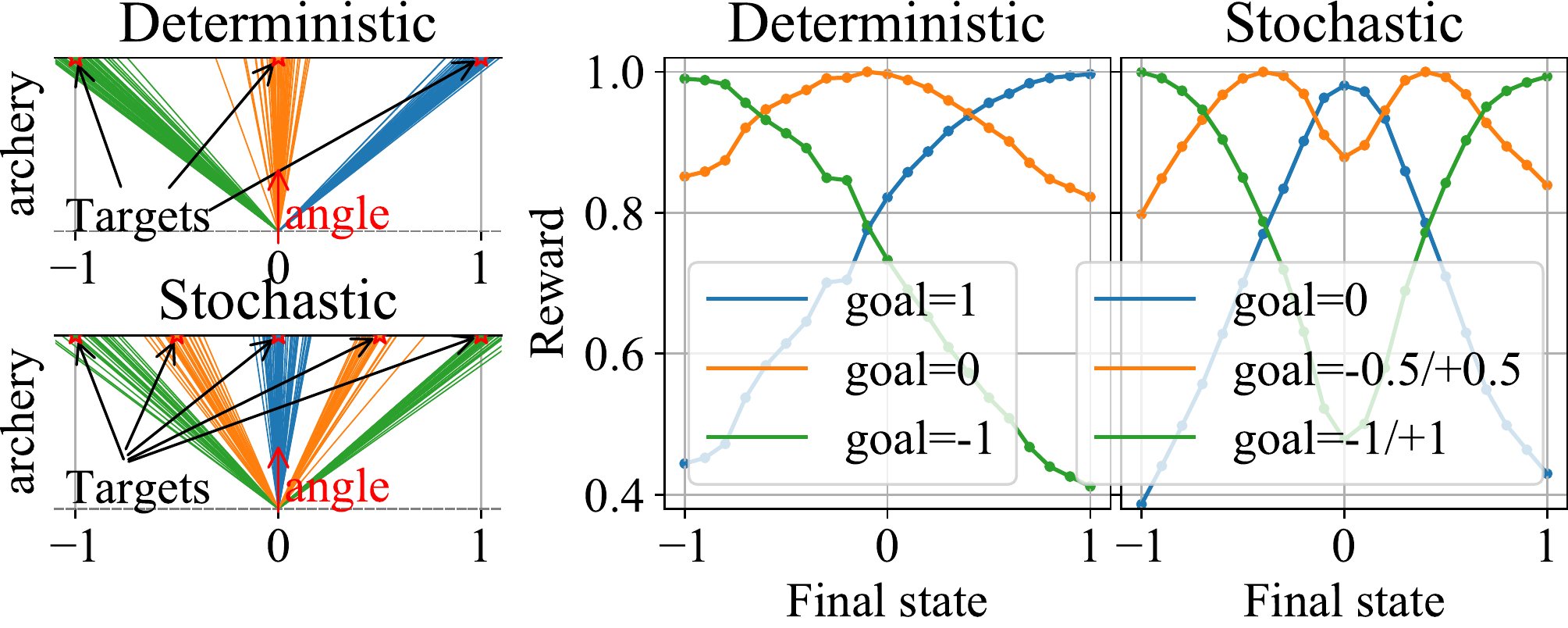}
	\caption{"Archery" tasks (left) and the learned rewards (right) on both deterministic and stochastic environments.}
	\label{arrow}
\end{figure}

\textbf{Visualizing the learned reward function.} 
We start with simple "archery" task to visualize how the learned reward function (discriminator $q_\phi$) accounts for goal-conditioned behaviors in environment. The 
task shown in Figure~\ref{arrow} requires choosing an angle at which we shoot an arrow to the target. The left upper subfigure shows that in a deterministic environment, given three different but fixed targets (with different colors), the arrow reaches the corresponding target successfully under the learned reward function $q_\phi$. 
The reward as a function of the final location of arrows in three tasks is shown on the right. We can find that the learned reward functions resemble convex in terms of the distance between final states to~targets.
Specifically, the maximum value of the learned reward function is achieved when the final state is close to the given target. The farther away the agent's final state is from the target, the smaller this reward value is. 
Similarly, the same conclusion can be drawn from the stochastic environment in the left lower subfigure, where the angle of the arrow has a $50\%$ chance to become a mirror symmetric angle.  
We see that the learned reward function substantially describes the environment's dynamics and the corresponding tasks, both in deterministic and stochastic environments. This answers our \emph{first} question.

\begin{figure*}[t]
	\centering
	\subfigure[2D navigation]{
		\begin{minipage}{0.185\linewidth}
			\centering
			\includegraphics[scale=0.11]{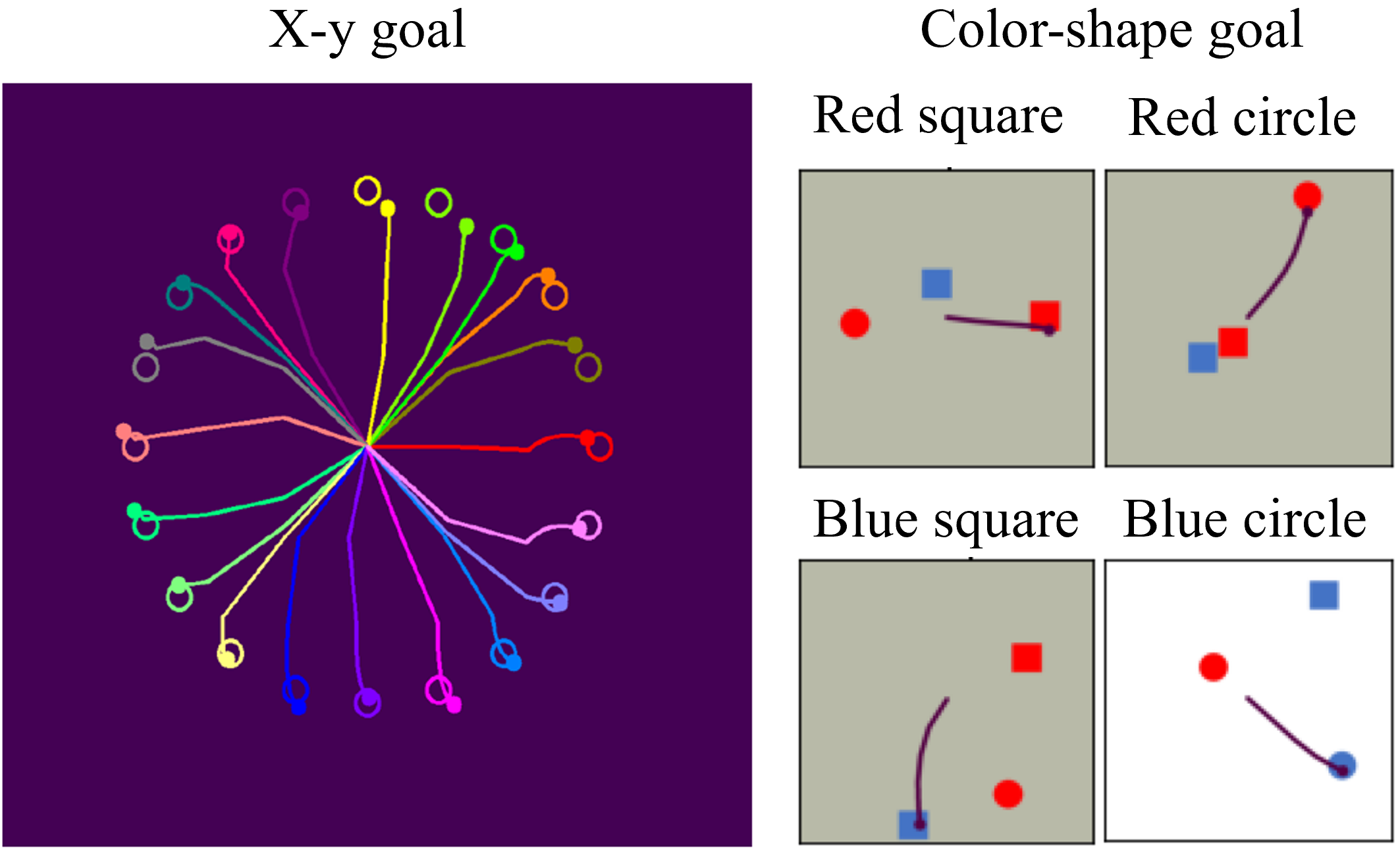}
		\end{minipage}
		\label{fig-exp1-1}
	}
	\subfigure[Object manipulation]{
		\begin{minipage}{0.185\linewidth}
			\centering
			\includegraphics[scale=0.100]{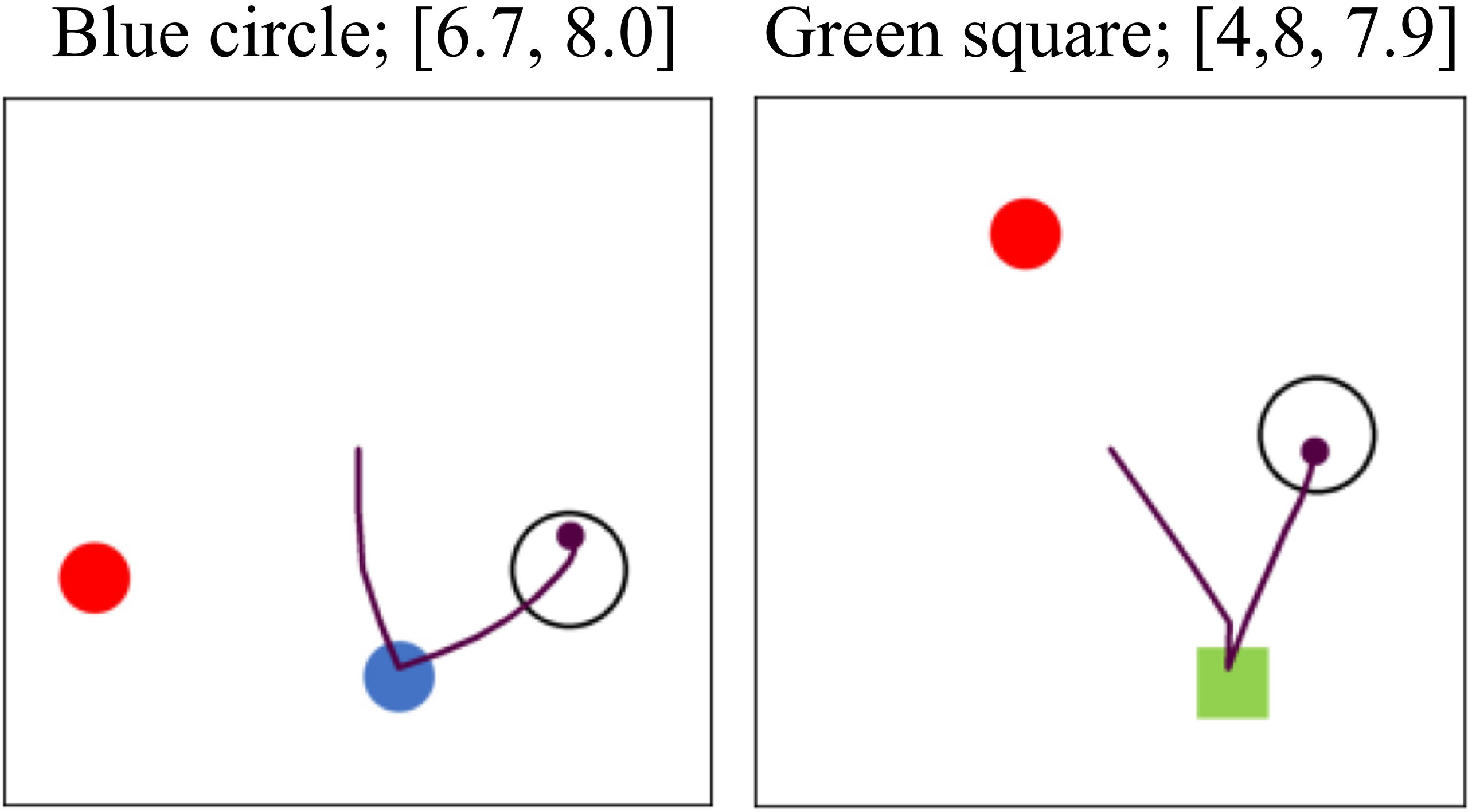}
		\end{minipage}
		\label{fig-exp1-2}
	}
	\subfigure[Atari games]{
		\centering
		\begin{minipage}{0.185\linewidth}
			\centering
			\includegraphics[scale=0.27]{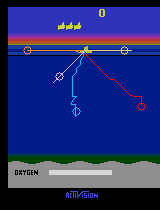}
			\includegraphics[scale=0.27]{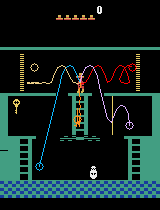}
		\end{minipage}
		\label{fig-exp1-3}
	}
	\subfigure[Swimmer]{
		\begin{minipage}{0.185\linewidth}
			\centering
			\includegraphics[scale=0.140]{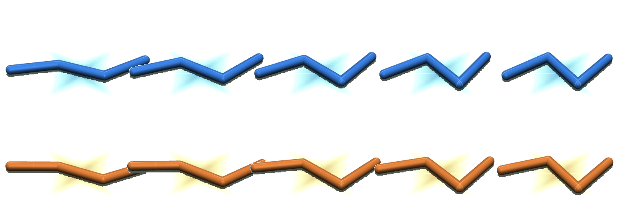}
			\includegraphics[scale=0.140]{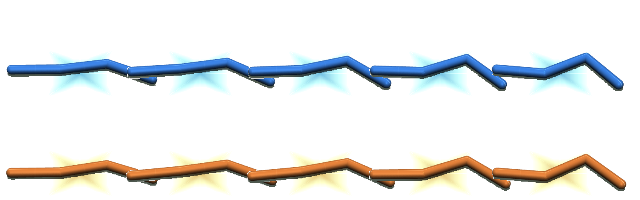}
		\end{minipage}
		\label{fig-exp1-4}
	}
	\subfigure[Half cheetah]{
		\begin{minipage}{0.185\linewidth}
			\centering
			\includegraphics[scale=0.140]{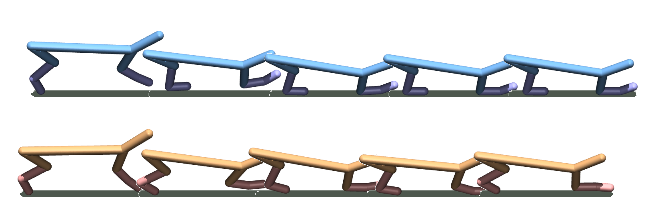}
			\includegraphics[scale=0.140]{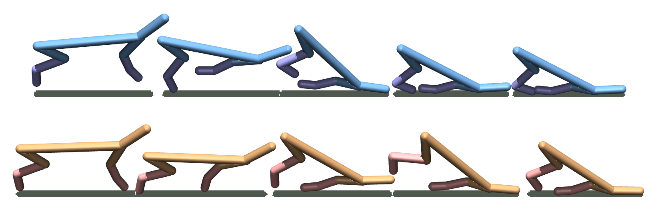}
		\end{minipage}
		\label{fig-exp1-5}
	}
	\caption{Goals and learned behaviors: Dots in 2D navigation (x-y goal)~and atari games denote different final (\emph{static}) goal states, and curves with same color represent corresponding trajectories; Goals in 2D navigation (color-shape goal) and object manipulation~are described using the text at the top of the diagram, where the purple lines imply the behaviors; In the swimmer and half~cheetah tasks, the first and third rows represent the \emph{dynamic} goals, and each row below represents the learned behaviors.} 
	\label{fig-exp1}
\end{figure*}

\textbf{Scaling to more complex tasks.} 
To answer our \emph{second} question, 
we now consider more complex tasks as shown in Figure~\ref{fig-exp1}. 
(1) In \emph{2D navigation tasks}, an agent can move in each of the four cardinal directions. 
We consider the following two tasks: moving the agent to a specific coordinate named \emph{x-y goal} (see appendix for details) and moving the agent to a specific object with certain color and shape named \emph{color-shape goal}. 
(2) \emph{Object manipulation} considers a moving agent in 2D environment with one block for manipulation, and the other block as a distractor. 
The agent first needs to reach the block and then move the block to the target location, where the block is described using color and shape. 
In other words, the description of the goal contains the \emph{color-shape goal} of the true block and the \emph{x-y goal} of the target coordinate. 
(3) Three \emph{atari games} including seaquest, berzerk and montezuma revenge require an agent to reach the given final states. 
(4) We use three \emph{mujoco tasks} (swimmer, half cheetah, and fetch) taken from OpenAI GYM~\citep{openaigym} to fast imitate given expert trajectories. 
Specifically, the \emph{static} goals for $\pi_\theta$ in 2D navigation, object manipulation and atari games are the relabeled final state $s_T$ induced by the latent-conditioned policy $\pi_\mu$: $g_t = f_\kappa(s_T)$, and the \emph{dynamic} goals for $\pi_\theta$ in mujoco tasks are the relabeled states induced by $\pi_\mu$ at each time step: $g_t= f_\kappa(s_{t+1})$ for $0 \leq t \leq T-1$.

The left subfigure of Figure~\ref{fig-exp1-1} shows the learned behavior of navigation in continuous action space given the x-y goal which is denoted as the small circle, and the right subfigure shows the trajectory of behavior with the given color-shape goal.  As observed, the agent manages to learn navigation tasks by using GPIM. 
Further, 2D navigation with color-shape goal (Figure~\ref{fig-exp1-1} \emph{right}) and object manipulation tasks (Figure~\ref{fig-exp1-2}) show the effectiveness of our model facing heterogeneous goals and states. Specifically, Figure~\ref{fig-exp1-2} shows the behaviors of the agent on object manipulation, where the agent is asked to first arrive at a block (i.e., blue circle and green square respectively) and then push it to the target location inside a dark circle (i.e., [6.7, 8.0] and [4.8, 7.9] respectively), where the red object exists as a distractor.
Figure~\ref{fig-exp1-3} shows the behaviors of agents that reach the final states in a higher dimensional (action, state and goal) space on \emph{seaquest} and \emph{montezuma revenge} respectively. 
Figure~\ref{fig-exp1}(d-e) shows how the agent imitates expert trajectories (dynamic goals) of \emph{swimmer} and \emph{half cheetah}.  We refer the reader to appendix for more results (task \emph{berzerk} and \emph{fetch}).

By learning to reach diverse goals generated by the latent-conditioned policy and employing the self-supervised loss over the perception-level to represent goals, the agent learns the ability to infer new goals later encountered by the agent. 
For example, as in Figure~\ref{fig-exp1-1} (\emph{right}), learning three behaviors with the goal of red-square, red-circle or blue-square (gray background) makes the agent accomplish the new goal of blue-circle (white background). 
In appendix, we also conduct the ablation study to show how the self-supervised loss (in the perception-level) affects behaviors, and provide more experiments to show the generalization to unseen~goals.




\begin{figure}[t]
	\centering
	\includegraphics[scale=0.55]{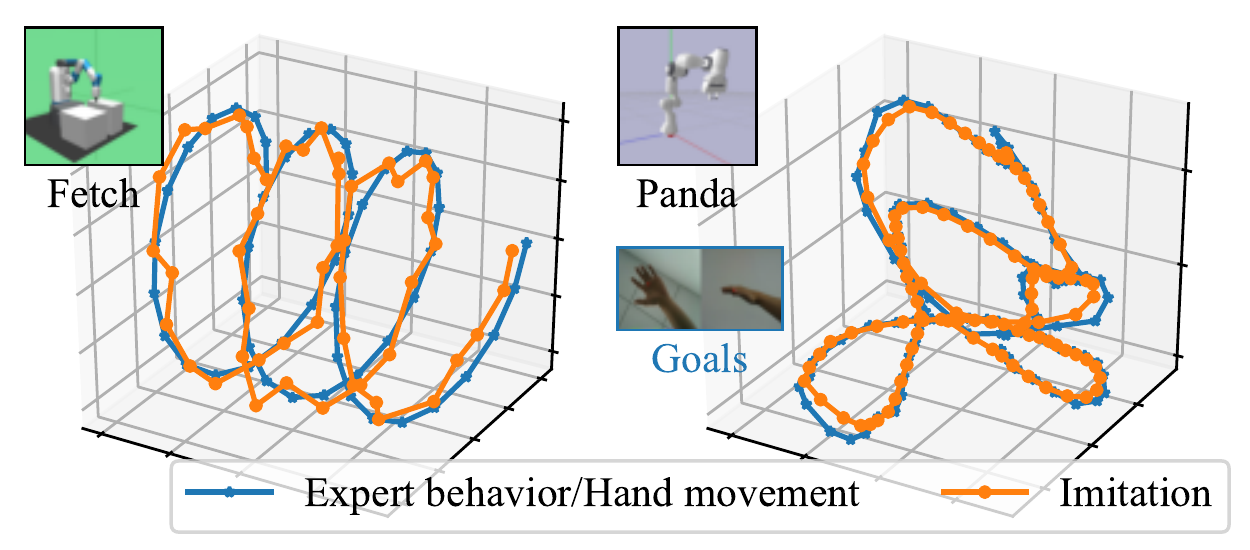}
	\caption{Dynamic goals on Fetch and Panda.}
	\label{temporally-extended-tasks}
	\vspace{-2pt}
\end{figure}

Compared to the usual relabeling procedure (e.g., HER with static goals) or latent variable base methods (e.g., DIAYN, equivalent to $f_\kappa = q_\phi$ in our GPIM), our approach scales to dynamic goals. 
Considering the setting of dynamic relabeling in \emph{fetch} task, we further demonstrate the ability of GPIM on temporally-extended tasks, where the 3D-coordinates of the gripper of the robotic arm is relabeled as goals for "imitation" in the training phase. 
During test, we employ a parameterized complex curve, $(x, y, z)=(t/5, \cos(t)/5-1/5, \sin(t)/5)$, for the gripper to follow and show their performance in Figure~\ref{temporally-extended-tasks} (left). 
It is worth noting that during training the agent is required to imitate a large number of simple behaviors and has never seen such complex goals before testing. 
We also validate GPIM on a Franka Panda robot (Figure~\ref{temporally-extended-tasks} right)), purposing tracking of hand movement, with MediaPipe~\citep{campillo2019mediapipe} to capturing features of images. 
It is observed from Figure~\ref{temporally-extended-tasks} that the imitation curves are almost overlapping with the given desired trajectories, indicating that the agent using GPIM framework has the potential to learn such compositional structure of goals during training and generalize to new composite goals during~test. 


\begin{figure*}[t]
	\centering
	\includegraphics[scale=0.310]{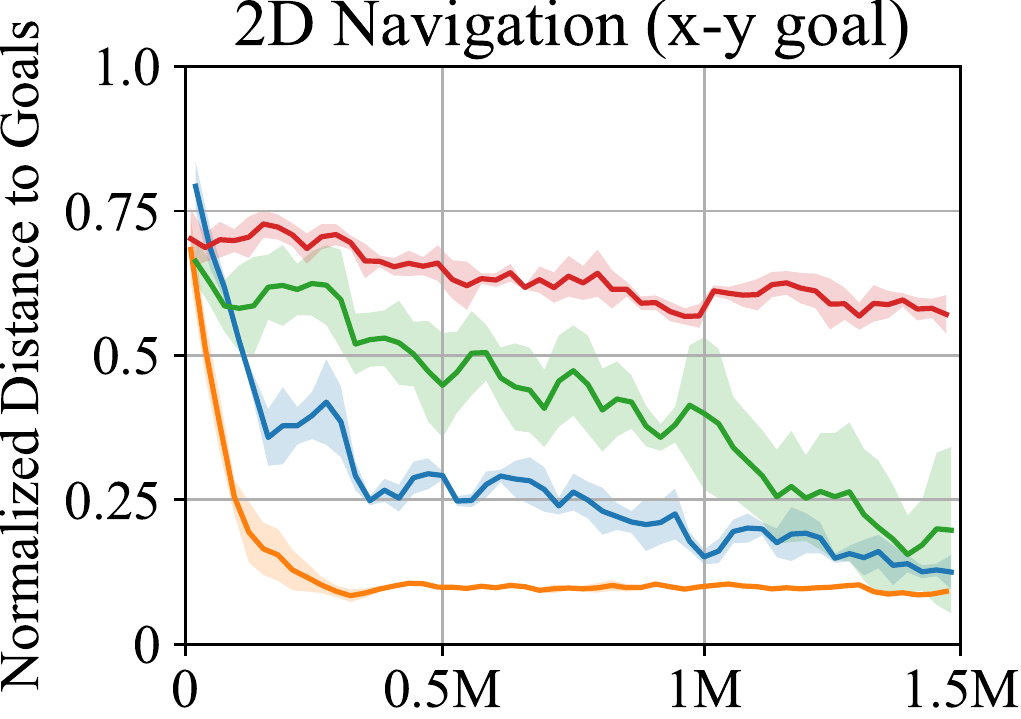}
	\includegraphics[scale=0.310]{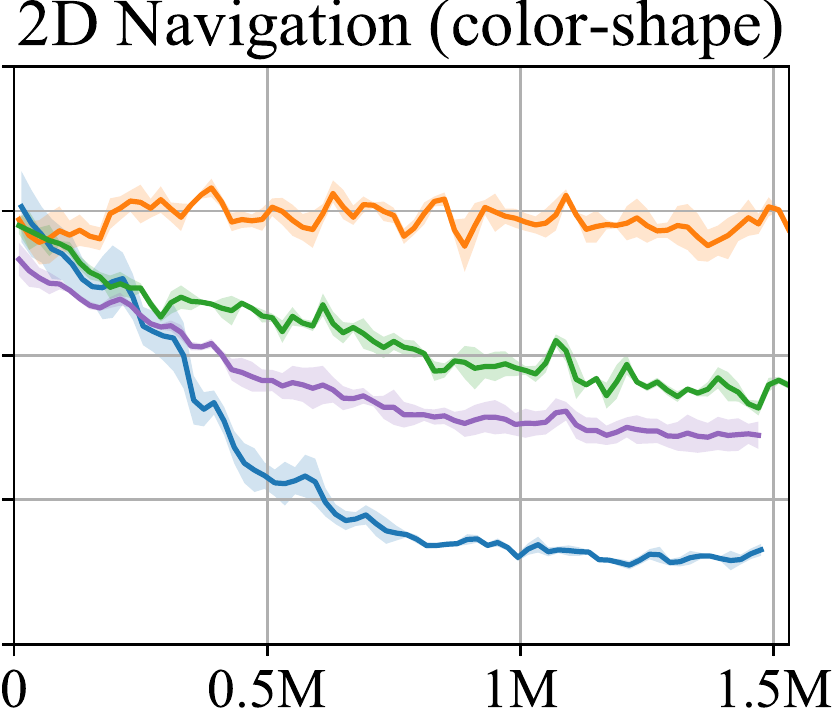}
	\includegraphics[scale=0.310]{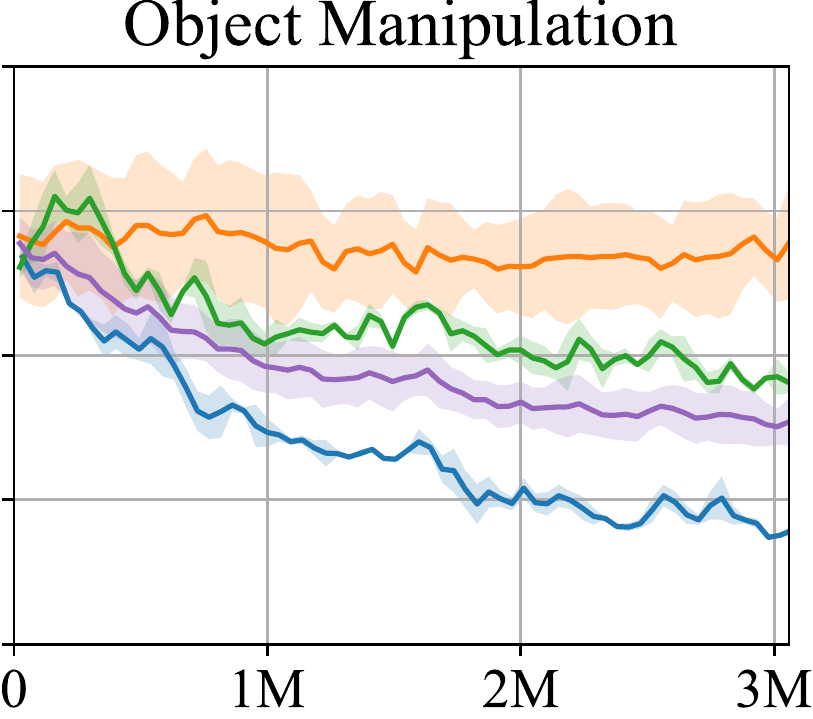}
		\includegraphics[scale=0.310]{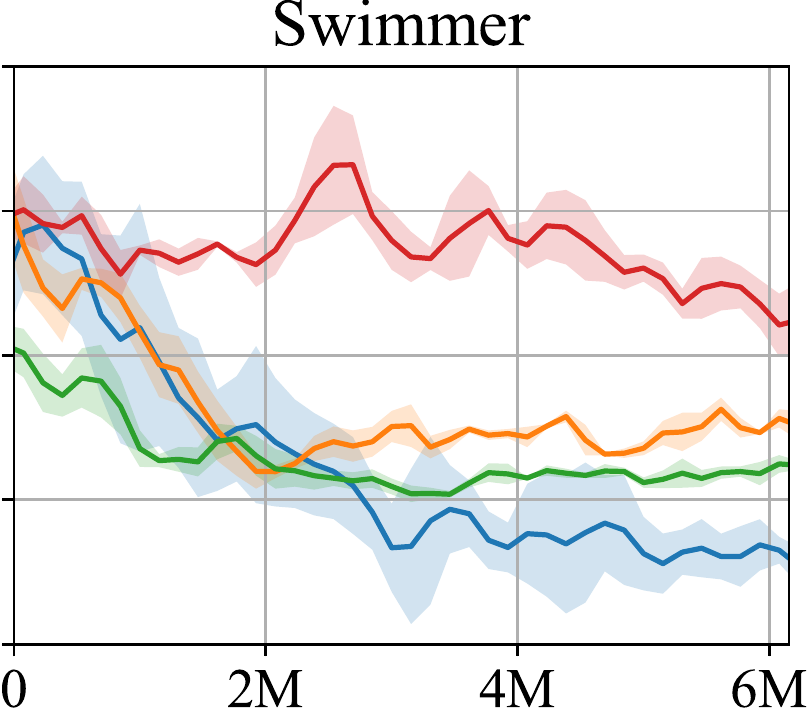}
		\includegraphics[scale=0.310]{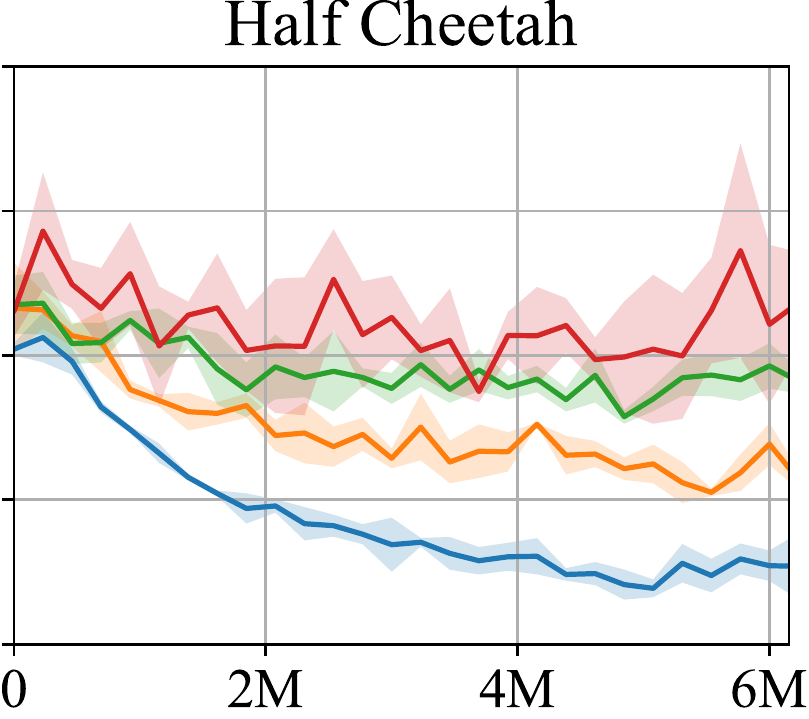}
		\includegraphics[scale=0.310]{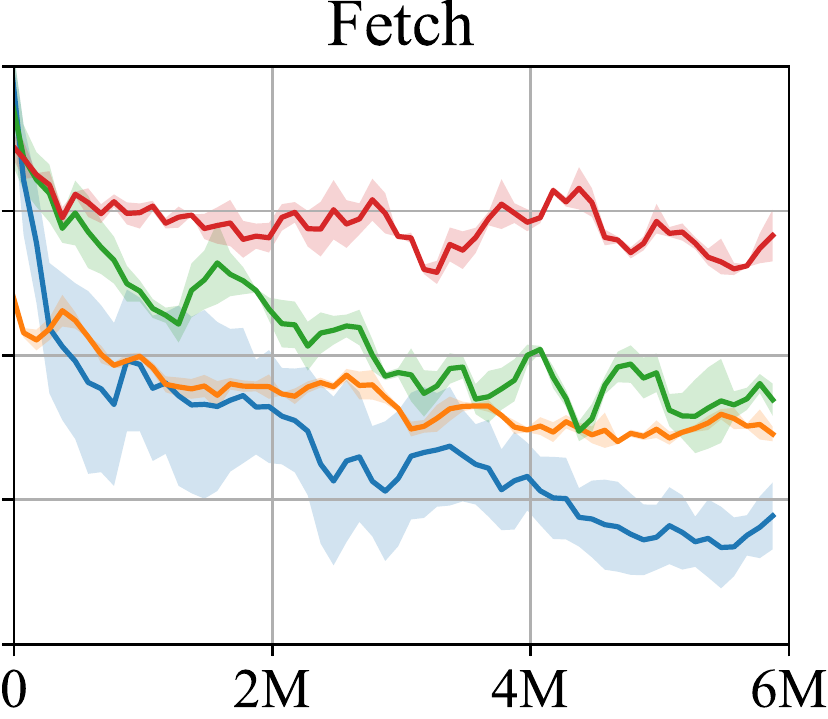}
	\includegraphics[scale=0.321]{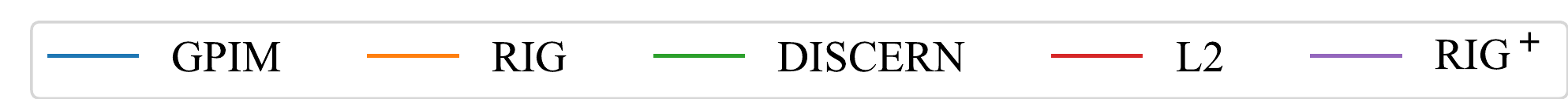}
	\caption{Performance (normalized distance to goals vs. actor steps) of our GPIM and baselines (RIG, DISCERN, L2, RIG$^+$).}
	\label{fig-exp3}
\end{figure*}

\textbf{Comparison with baselines.} 
For the \emph{third} question, we mainly compare our method to three~baselines: \textbf{RIG} \citep{nair2018visual}, \textbf{DISCERN} \citep{warde2018unsupervised}, and \textbf{L2 Distance}. L2 Distance measures the distance between~states and goals, where the $L2$ distance $-||\tilde{s}_{t+1} - g_t||^2/\sigma_{pixel}$ is considered with a hyperparameter $\sigma_{pixel}$. 
Note that 2D navigation with the color-shape goal and object manipulation using text description makes the dimensions of states and goals different, so L2 cannot be used in these two tasks. 
In RIG,~we obtain rewards by using the distances in two embedding spaces and learning two independent VAEs, where one VAE is to encode states and the other is to encode goals. For this~heterogeneous setting, we also conduct baseline \textbf{RIG$^+$} by training one VAE only on goals and then reward~agent~with the distance between the embeddings of goals and relabeled states (i.e., $g_t$ vs. $f_\kappa(\tilde{s}_{t+1})$). 
We use~the~normalized distance to goals as the evaluation metric, where we generate 50 goals (tasks)~as~validation.  

We show the results in Figure~\ref{fig-exp3} by plotting the normalized distance to goals as a function of the number of actor's steps, where each curve considers 95\% confidence interval in terms of the mean value across three seeds. 
As observed, our GPIM consistently outperforms baselines in almost all tasks except for the RIG in 2D navigation (x-y goal) due to the simplicity of this task. 
Particularly, as the task complexity increases from 2D navigation (x-y goal) to 2D navigation (color-shape goal) and eventually object manipulation (mixed x-y goal and color-shape goal), GPIM converges faster than baselines and the performance gap between our GPIM and baselines becomes larger. 
Moreover, although RIG learns fast on navigation with x-y goal, it fails to accomplish complex navigation with color-shape goal because the embedding distance between two independent VAEs has difficulty in capturing the correlation of heterogeneous states and goals. Even with a stable VAE, RIG$^+$ can be poorly suited for training the goal-reaching policy. 
Especially in high-dimensional action space~and~on more exploratory tasks (atari and mujoco tasks), our method substantially outperforms the baselines. 

\begin{figure}[t]
	\centering
	\includegraphics[scale=0.58]{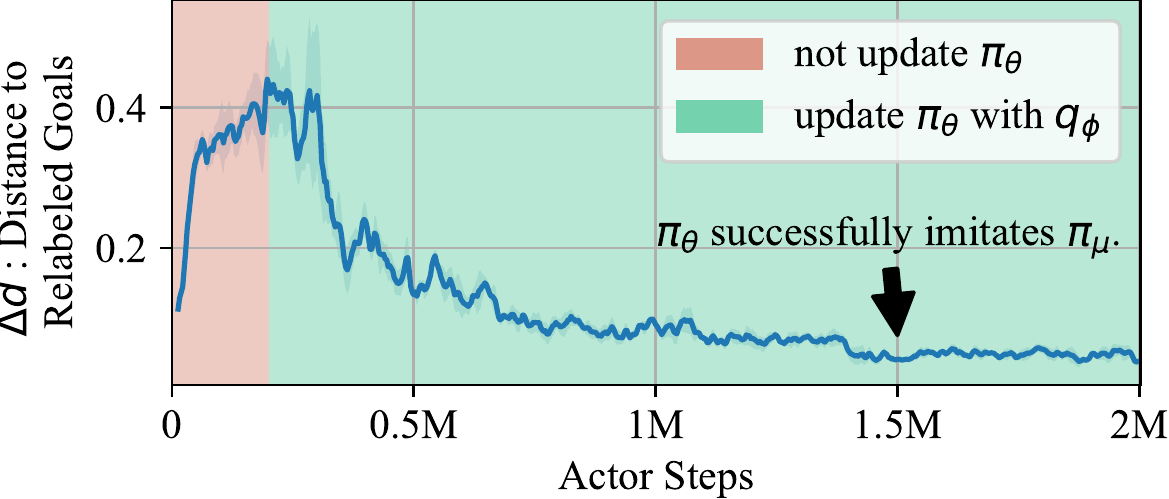}
	\caption{\emph{Pink slice}: Latent-conditioned $\pi_\mu$ gradually explores environment, generating more difficult goals. \emph{Mint green}: Learned discriminator $q_\phi$ encourages $\pi_\theta$ to mimic~$\pi_\mu$.}
	\label{arrow-middle}
\end{figure}

To gain more intuition for our method, we record the distance ($\Delta d$) between the goal induced by $\pi_\mu$ and the final state induced by $\pi_\theta$ throughout the training process of the 2D navigation (x-y goal). 
In this specific experiment, we update $\pi_\mu$ and $q_\phi$ but ignore the update of $\pi_\theta$ before $200~\text{k}$ steps to show the exploration of $\pi_\mu$ at the task generation phase. 
As shown in Figure~\ref{arrow-middle}, $\Delta d$ steadily increases during the first $200~\text{k}$ steps, indicating that the latent-conditioned policy $\pi_\mu$ explores the environment (i.e., goal space) to distinguish skills more easily (with $q_\phi$), and as a result, generates diverse goals for training  
goal-conditioned policy $\pi_\theta$.  
After around $1.5~\text{M}$ steps, $\Delta d$ almost comes to $0$, indicating that goal-conditioned $\pi_\theta$ has learned a good strategy to reach the relabeled goals. In appendix, we visually show the generated goals in more complex tasks, which shows that our straightforward~framework can effectively explore without additional sophisticated exploration strategies. 

\begin{figure}[t]
	\centering
	\includegraphics[width=\linewidth]{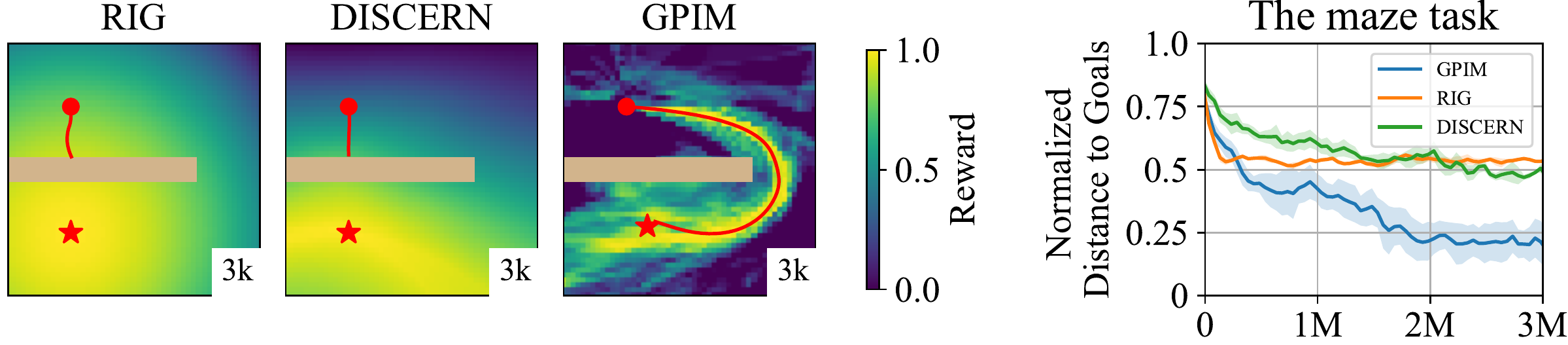}
	\caption{(Left) Reward functions, where heatmaps depict the reward conditioned on the bottom-left goal, reaching the left-bottom star. 
		(Right) Learning curves on the left maze. 
	}
	\label{fig-exp-reward}
\end{figure}

\textbf{Expressiveness of the reward function.} 
Particularly, the performance of unsupervised~RL methods depend on the diversity of generated goals and the expressiveness of the learned reward functions that are conditioned on the goals. 
We show that our straightforward framework can effectively explore environments in appendix (though it is not our focus). 
The next question is that: 
with the same exploration capability to generate goals, does our model achieve competitive performance against the baselines? Said another way, will the obtained rewards (over embedding space) of baselines taking prior non-parametric functions limit the repertoires of learning tasks in some environments? 
For better graphical interpretation and comparison with baselines, we simplify the complex Atari games to a maze environment shown in Figure~\ref{fig-exp-reward}, where the middle wall poses a bottleneck state. 
At the same time, as an example to show the compatibility of our objective with existing exploration strategies~\citep{jabri2019unsupervised, lee2019efficient}, we set the reward for the latent-conditioned policy $\pi_\mu$ as $r'_t = \lambda r_t + (\lambda - 1) \log q_\nu(s_{t+1})$,~where~$q_\nu$ is a density model, and $\lambda \in [0, 1]$ can be interpreted as trade off between discriminability of skills and task-specific exploration (here we set $\lambda = 0.5$). Note that we modify $r'_t$ for improving the exploration on generating goals and we do not change the reward for 
training goal-conditioned $\pi_\theta$. To guarantee the generation of same diverse goals for goal-conditioned 
policies of baselines, we adopt $\pi_\mu$ taking the modified $r'_t$ to generate goals for RIG and DISCERN. 

In Figure~\ref{fig-exp-reward}, we visualize the learned reward on a specific task reaching the left-bottom star, and the learning curves on the maze task, where the testing-goals are random sampled. We can see that the learned rewards of RIG and DISCERN produce poor signal for the goal-conditioned policy, which makes learning vulnerable to local optima. Our method acquires the reward function $q_\phi$~after~exploring the environment, dynamics of which itself further shapes the reward function. In Figure~\ref{fig-exp-reward}~(left), we can see that our model provides the reward function better expressiveness of the task by compensating~for the dynamics of training environment. 
This produces that, even with the same exploration capability to generate diverse goals, our model sufficiently outperforms the baselines, as shown in Figure~\ref{fig-exp-reward}~(right).


\section{Conclusion}
\label{conclusion}

In this paper, we propose GPIM to learn a goal-conditioned policy in an unsupervised manner. 
The core idea of GPIM lies in that we introduce a latent-conditioned policy with a procedural  relabeling procedure to generate tasks (goals and the associated reward functions) for training the goal-conditioned policy. 
For goal-reaching tasks, we theoretically describe the performance guarantee of our (unsupervised) objective compared with the standard multi-goal RL.   
We also conduct extensive experiments on a variety of tasks to demonstrate the effectiveness and efficiency~of our method. 

There are several potential directions for future in our unsupervised relabeling framework. One promising direction would be developing a domain adaptation mechanism when the interaction environments (action/state spaces, dynamics, or initial states) wrt learning $\pi_\mu$ and $\pi_\theta$ are different. Additionally, GPIM can get benefits from more extensive exploration strategies to control the exploration-exploitation trade-off. 
Finally, latent-conditioned $\pi_\mu$ (generating goals and reward functions) is not affected by the goal-conditioned $\pi_\theta$ in GPIM. One can develop self-paced (curriculum) learning over the two policies under the unsupervised RL setting.

\section*{Acknowledgments}
The authors would like to thank Yachen Kang for helpful discussions and the anonymous reviewers for the valuable comments. This work is supported by NSFC General Program (62176215).

\bibliography{references}

\begin{thebibliography}{51}
\providecommand{\natexlab}[1]{#1}

\bibitem[{Achiam et~al.(2018)Achiam, Edwards, Amodei, and
  Abbeel}]{achiam2018variational}
Achiam, J.; Edwards, H.; Amodei, D.; and Abbeel, P. 2018.
\newblock Variational Option Discovery Algorithms.
\newblock \emph{CoRR}, abs/1807.10299.

\bibitem[{Amodei et~al.(2016)Amodei, Olah, Steinhardt, Christiano, Schulman,
  and Man{\'{e}}}]{amodei2016concrete}
Amodei, D.; Olah, C.; Steinhardt, J.; Christiano, P.~F.; Schulman, J.; and
  Man{\'{e}}, D. 2016.
\newblock Concrete problems in AI safety.
\newblock \emph{CoRR}, abs/1606.06565.

\bibitem[{Andrychowicz et~al.(2017)Andrychowicz, Wolski, Ray, Schneider, Fong,
  Welinder, McGrew, Tobin, Abbeel, and Zaremba}]{andrychowicz2017hindsight}
Andrychowicz, M.; Wolski, F.; Ray, A.; Schneider, J.; Fong, R.; Welinder, P.;
  McGrew, B.; Tobin, J.; Abbeel, O.~P.; and Zaremba, W. 2017.
\newblock Hindsight experience replay.
\newblock In \emph{Advances in neural information processing systems},
  5048--5058.

\bibitem[{Badia et~al.(2020)Badia, Piot, Kapturowski, Sprechmann, Vitvitskyi,
  Guo, and Blundell}]{badia2020agent57}
Badia, A.~P.; Piot, B.; Kapturowski, S.; Sprechmann, P.; Vitvitskyi, A.; Guo,
  D.; and Blundell, C. 2020.
\newblock Agent57: Outperforming the atari human benchmark.
\newblock \emph{arXiv preprint arXiv:2003.13350}.

\bibitem[{Barber and Agakov(2003)}]{barber2003algorithm}
Barber, D.; and Agakov, F.~V. 2003.
\newblock The IM algorithm: a variational approach to information maximization.
\newblock In \emph{Advances in neural information processing systems}, None.

\bibitem[{Beaudry and Renner(2011)}]{beaudry2011intuitive}
Beaudry, N.~J.; and Renner, R. 2011.
\newblock An intuitive proof of the data processing inequality.
\newblock \emph{arXiv preprint arXiv:1107.0740}.

\bibitem[{Berner et~al.(2019)Berner, Brockman, Chan, Cheung, Debiak, Dennison,
  Farhi, Fischer, Hashme, Hesse et~al.}]{berner2019dota}
Berner, C.; Brockman, G.; Chan, B.; Cheung, V.; Debiak, P.; Dennison, C.;
  Farhi, D.; Fischer, Q.; Hashme, S.; Hesse, C.; et~al. 2019.
\newblock Dota 2 with large scale deep reinforcement learning.
\newblock \emph{arXiv preprint arXiv:1912.06680}.

\bibitem[{Brockman et~al.(2016)Brockman, Cheung, Pettersson, Schneider,
  Schulman, Tang, and Zaremba}]{openaigym}
Brockman, G.; Cheung, V.; Pettersson, L.; Schneider, J.; Schulman, J.; Tang,
  J.; and Zaremba, W. 2016.
\newblock OpenAI Gym.
\newblock .

\bibitem[{Campos et~al.(2020)Campos, Trott, Xiong, Socher, i~Nieto, and
  Torres}]{campos2020ExploreDA}
Campos, V.~A.; Trott, A.; Xiong, C.; Socher, R.; i~Nieto, X.~G.; and Torres, J.
  2020.
\newblock Explore, Discover and Learn: Unsupervised Discovery of State-Covering
  Skills.
\newblock \emph{ArXiv}, abs/2002.03647.

\bibitem[{Colas et~al.(2020)Colas, Karch, Lair, Dussoux, Moulin{-}Frier,
  Dominey, and Oudeyer}]{Colas2020LanguageAA}
Colas, C.; Karch, T.; Lair, N.; Dussoux, J.; Moulin{-}Frier, C.; Dominey,
  P.~F.; and Oudeyer, P. 2020.
\newblock Language as a Cognitive Tool to Imagine Goals in Curiosity Driven
  Exploration.
\newblock In Larochelle, H.; Ranzato, M.; Hadsell, R.; Balcan, M.; and Lin, H.,
  eds., \emph{Advances in Neural Information Processing Systems 33: Annual
  Conference on Neural Information Processing Systems 2020, NeurIPS 2020,
  December 6-12, 2020, virtual}.

\bibitem[{Colas, Sigaud, and Oudeyer(2018)}]{colas2018gep}
Colas, C.; Sigaud, O.; and Oudeyer, P.-Y. 2018.
\newblock Gep-pg: Decoupling exploration and exploitation in deep reinforcement
  learning algorithms.
\newblock \emph{arXiv preprint arXiv:1802.05054}.

\bibitem[{Eysenbach et~al.(2018)Eysenbach, Gupta, Ibarz, and
  Levine}]{eysenbach2018diversity}
Eysenbach, B.; Gupta, A.; Ibarz, J.; and Levine, S. 2018.
\newblock Diversity is all you need: Learning skills without a reward function.
\newblock \emph{arXiv preprint arXiv:1802.06070}.

\bibitem[{Florensa et~al.(2019)Florensa, Degrave, Heess, Springenberg, and
  Riedmiller}]{florensa2019self}
Florensa, C.; Degrave, J.; Heess, N.; Springenberg, J.~T.; and Riedmiller, M.
  2019.
\newblock Self-supervised learning of image embedding for continuous control.
\newblock \emph{arXiv preprint arXiv:1901.00943}.

\bibitem[{Gregor, Rezende, and Wierstra(2017)}]{gregor2016variational}
Gregor, K.; Rezende, D.~J.; and Wierstra, D. 2017.
\newblock Variational Intrinsic Control.

\bibitem[{Gupta et~al.(2018)Gupta, Eysenbach, Finn, and
  Levine}]{gupta2018unsupervised}
Gupta, A.; Eysenbach, B.; Finn, C.; and Levine, S. 2018.
\newblock Unsupervised Meta-Learning for Reinforcement Learning.
\newblock \emph{CoRR}, abs/1806.04640.

\bibitem[{Haarnoja et~al.(2018)Haarnoja, Zhou, Abbeel, and
  Levine}]{haarnoja2018soft}
Haarnoja, T.; Zhou, A.; Abbeel, P.; and Levine, S. 2018.
\newblock Soft actor-critic: Off-policy maximum entropy deep reinforcement
  learning with a stochastic actor.
\newblock \emph{arXiv preprint arXiv:1801.01290}.

\bibitem[{Hafner et~al.(2020)Hafner, Ortega, Ba, Parr, Friston, and
  Heess}]{hafner2020action}
Hafner, D.; Ortega, P.~A.; Ba, J.; Parr, T.; Friston, K.~J.; and Heess, N.
  2020.
\newblock Action and Perception as Divergence Minimization.
\newblock \emph{CoRR}, abs/2009.01791.

\bibitem[{Hartikainen et~al.(2019)Hartikainen, Geng, Haarnoja, and
  Levine}]{hartikainen2019dynamical}
Hartikainen, K.; Geng, X.; Haarnoja, T.; and Levine, S. 2019.
\newblock Dynamical Distance Learning for Semi-Supervised and Unsupervised
  Skill Discovery.
\newblock In \emph{International Conference on Learning Representations}.

\bibitem[{Higgins et~al.(2017)Higgins, Pal, Rusu, Matthey, Burgess, Pritzel,
  Botvinick, Blundell, and Lerchner}]{higgins2017darla}
Higgins, I.; Pal, A.; Rusu, A.; Matthey, L.; Burgess, C.; Pritzel, A.;
  Botvinick, M.; Blundell, C.; and Lerchner, A. 2017.
\newblock Darla: Improving zero-shot transfer in reinforcement learning.
\newblock In \emph{Proceedings of the 34th International Conference on Machine
  Learning-Volume 70}, 1480--1490. JMLR. org.

\bibitem[{Jabri et~al.(2019)Jabri, Hsu, Gupta, Eysenbach, Levine, and
  Finn}]{jabri2019unsupervised}
Jabri, A.; Hsu, K.; Gupta, A.; Eysenbach, B.; Levine, S.; and Finn, C. 2019.
\newblock Unsupervised curricula for visual meta-reinforcement learning.
\newblock In \emph{Advances in Neural Information Processing Systems},
  10519--10531.

\bibitem[{Khirodkar, Yoo, and Kitani(2018)}]{khirodkar2018vadra}
Khirodkar, R.; Yoo, D.; and Kitani, K.~M. 2018.
\newblock VADRA: Visual adversarial domain randomization and augmentation.
\newblock \emph{arXiv preprint arXiv:1812.00491}.

\bibitem[{Kova{\v{c}}, Laversanne-Finot, and Oudeyer(2020)}]{kovavc2020grimgep}
Kova{\v{c}}, G.; Laversanne-Finot, A.; and Oudeyer, P.-Y. 2020.
\newblock Grimgep: learning progress for robust goal sampling in visual deep
  reinforcement learning.
\newblock \emph{arXiv preprint arXiv:2008.04388}.

\bibitem[{Laskin, Srinivas, and Abbeel(2020)}]{srinivas2020curl}
Laskin, M.; Srinivas, A.; and Abbeel, P. 2020.
\newblock {CURL:} Contrastive Unsupervised Representations for Reinforcement
  Learning.
\newblock 119: 5639--5650.

\bibitem[{Lee et~al.(2019)Lee, Eysenbach, Parisotto, Xing, Levine, and
  Salakhutdinov}]{lee2019efficient}
Lee, L.; Eysenbach, B.; Parisotto, E.; Xing, E.~P.; Levine, S.; and
  Salakhutdinov, R. 2019.
\newblock Efficient Exploration via State Marginal Matching.
\newblock \emph{CoRR}, abs/1906.05274.

\bibitem[{Lee et~al.(2018)Lee, Sun, Somasundaram, Hu, and
  Lim}]{lee2018composing}
Lee, Y.; Sun, S.-H.; Somasundaram, S.; Hu, E.~S.; and Lim, J.~J. 2018.
\newblock Composing complex skills by learning transition policies.
\newblock In \emph{International Conference on Learning Representations}.

\bibitem[{Levy et~al.(2017)Levy, Konidaris, Platt, and
  Saenko}]{levy2017learning}
Levy, A.; Konidaris, G.; Platt, R.; and Saenko, K. 2017.
\newblock Learning multi-level hierarchies with hindsight.
\newblock \emph{arXiv preprint arXiv:1712.00948}.

\bibitem[{Liu et~al.(2021)Liu, Zhang, Zhao, Qin, Zhu, Li, Yu, and
  Liu}]{liu2021return}
Liu, G.; Zhang, C.; Zhao, L.; Qin, T.; Zhu, J.; Li, J.; Yu, N.; and Liu, T.
  2021.
\newblock Return-Based Contrastive Representation Learning for Reinforcement
  Learning.
\newblock \emph{CoRR}, abs/2102.10960.

\bibitem[{Lowrey et~al.(2019)Lowrey, Rajeswaran, Kakade, Todorov, and
  Mordatch}]{lowrey2018plan}
Lowrey, K.; Rajeswaran, A.; Kakade, S.~M.; Todorov, E.; and Mordatch, I. 2019.
\newblock Plan Online, Learn Offline: Efficient Learning and Exploration via
  Model-Based Control.

\bibitem[{Lu et~al.(2020)Lu, Lee, Abbeel, and Tiomkin}]{xingyu2020dynamcis}
Lu, X.; Lee, K.; Abbeel, P.; and Tiomkin, S. 2020.
\newblock Dynamics Generalization via Information Bottleneck in Deep
  Reinforcement Learning.
\newblock \emph{CoRR}, abs/2008.00614.

\bibitem[{Lugaresi et~al.(2019)Lugaresi, Tang, Nash, McClanahan, Uboweja, Hays,
  Zhang, Chang, Yong, Lee, Chang, Hua, Georg, and
  Grundmann}]{campillo2019mediapipe}
Lugaresi, C.; Tang, J.; Nash, H.; McClanahan, C.; Uboweja, E.; Hays, M.; Zhang,
  F.; Chang, C.; Yong, M.~G.; Lee, J.; Chang, W.; Hua, W.; Georg, M.; and
  Grundmann, M. 2019.
\newblock MediaPipe: {A} Framework for Building Perception Pipelines.
\newblock \emph{CoRR}, abs/1906.08172.

\bibitem[{Nair et~al.(2019)Nair, Bahl, Khazatsky, Pong, Berseth, and
  Levine}]{nair2019contextual}
Nair, A.; Bahl, S.; Khazatsky, A.; Pong, V.; Berseth, G.; and Levine, S. 2019.
\newblock Contextual Imagined Goals for Self-Supervised Robotic Learning.
\newblock 100: 530--539.

\bibitem[{Nair et~al.(2018)Nair, Pong, Dalal, Bahl, Lin, and
  Levine}]{nair2018visual}
Nair, A.~V.; Pong, V.; Dalal, M.; Bahl, S.; Lin, S.; and Levine, S. 2018.
\newblock Visual reinforcement learning with imagined goals.
\newblock In \emph{Advances in Neural Information Processing Systems},
  9191--9200.

\bibitem[{P{\'e}r{\'e} et~al.(2018)P{\'e}r{\'e}, Forestier, Sigaud, and
  Oudeyer}]{pere2018unsupervised}
P{\'e}r{\'e}, A.; Forestier, S.; Sigaud, O.; and Oudeyer, P.-Y. 2018.
\newblock Unsupervised learning of goal spaces for intrinsically motivated goal
  exploration.
\newblock \emph{arXiv preprint arXiv:1803.00781}.

\bibitem[{Pitis et~al.(2020)Pitis, Chan, Zhao, Stadie, and
  Ba}]{Pitis2020MaximumEG}
Pitis, S.; Chan, H.; Zhao, S.; Stadie, B.~C.; and Ba, J. 2020.
\newblock Maximum Entropy Gain Exploration for Long Horizon Multi-goal
  Reinforcement Learning.
\newblock \emph{ArXiv}, abs/2007.02832.

\bibitem[{Pong et~al.(2018)Pong, Gu, Dalal, and Levine}]{pong2018temporal}
Pong, V.; Gu, S.; Dalal, M.; and Levine, S. 2018.
\newblock Temporal difference models: Model-free deep rl for model-based
  control.
\newblock \emph{arXiv preprint arXiv:1802.09081}.

\bibitem[{Pong et~al.(2019)Pong, Dalal, Lin, Nair, Bahl, and
  Levine}]{pong2019skew}
Pong, V.~H.; Dalal, M.; Lin, S.; Nair, A.; Bahl, S.; and Levine, S. 2019.
\newblock Skew-fit: State-covering self-supervised reinforcement learning.
\newblock \emph{arXiv preprint arXiv:1903.03698}.

\bibitem[{Popov et~al.(2017)Popov, Heess, Lillicrap, Hafner, Barth-Maron,
  Vecerik, Lampe, Tassa, Erez, and Riedmiller}]{popov2017data}
Popov, I.; Heess, N.; Lillicrap, T.; Hafner, R.; Barth-Maron, G.; Vecerik, M.;
  Lampe, T.; Tassa, Y.; Erez, T.; and Riedmiller, M. 2017.
\newblock Data-efficient deep reinforcement learning for dexterous
  manipulation.
\newblock \emph{arXiv preprint arXiv:1704.03073}.

\bibitem[{Salge, Glackin, and Polani(2014)}]{salge2014empowerment}
Salge, C.; Glackin, C.; and Polani, D. 2014.
\newblock Empowerment--an introduction.
\newblock In \emph{Guided Self-Organization: Inception}, 67--114. Springer.

\bibitem[{Schaul et~al.(2015)Schaul, Horgan, Gregor, and
  Silver}]{schaul2015universal}
Schaul, T.; Horgan, D.; Gregor, K.; and Silver, D. 2015.
\newblock Universal value function approximators.
\newblock In \emph{International conference on machine learning}, 1312--1320.

\bibitem[{Schulman et~al.(2017)Schulman, Wolski, Dhariwal, Radford, and
  Klimov}]{schulman2017proximal}
Schulman, J.; Wolski, F.; Dhariwal, P.; Radford, A.; and Klimov, O. 2017.
\newblock Proximal Policy Optimization Algorithms.
\newblock \emph{CoRR}, abs/1707.06347.

\bibitem[{Sermanet et~al.(2018)Sermanet, Lynch, Chebotar, Hsu, Jang, Schaal,
  Levine, and Brain}]{sermanet2018time}
Sermanet, P.; Lynch, C.; Chebotar, Y.; Hsu, J.; Jang, E.; Schaal, S.; Levine,
  S.; and Brain, G. 2018.
\newblock Time-contrastive networks: Self-supervised learning from video.
\newblock In \emph{2018 IEEE International Conference on Robotics and
  Automation (ICRA)}, 1134--1141. IEEE.

\bibitem[{Sharma et~al.(2020)Sharma, Gu, Levine, Kumar, and
  Hausman}]{sharma2019dynamics}
Sharma, A.; Gu, S.; Levine, S.; Kumar, V.; and Hausman, K. 2020.
\newblock Dynamics-Aware Unsupervised Discovery of Skills.

\bibitem[{Sukhbaatar et~al.(2018)Sukhbaatar, Denton, Szlam, and
  Fergus}]{sukhbaatar2018learning}
Sukhbaatar, S.; Denton, E.; Szlam, A.; and Fergus, R. 2018.
\newblock Learning goal embeddings via self-play for hierarchical reinforcement
  learning.
\newblock \emph{arXiv preprint arXiv:1811.09083}.

\bibitem[{Sukhbaatar et~al.(2017)Sukhbaatar, Lin, Kostrikov, Synnaeve, Szlam,
  and Fergus}]{sukhbaatar2017intrinsic}
Sukhbaatar, S.; Lin, Z.; Kostrikov, I.; Synnaeve, G.; Szlam, A.; and Fergus, R.
  2017.
\newblock Intrinsic motivation and automatic curricula via asymmetric
  self-play.
\newblock \emph{arXiv preprint arXiv:1703.05407}.

\bibitem[{Tian, Liu, and Wang(2020)}]{tian2020learning}
Tian, Q.; Liu, J.; and Wang, D. 2020.
\newblock Learning transitional skills with intrinsic motivation.

\bibitem[{Tobin et~al.(2017)Tobin, Fong, Ray, Schneider, Zaremba, and
  Abbeel}]{tobin2017domain}
Tobin, J.; Fong, R.; Ray, A.; Schneider, J.; Zaremba, W.; and Abbeel, P. 2017.
\newblock Domain randomization for transferring deep neural networks from
  simulation to the real world.
\newblock In \emph{2017 IEEE/RSJ International Conference on Intelligent Robots
  and Systems (IROS)}, 23--30. IEEE.

\bibitem[{Todorov, Erez, and Tassa(2012)}]{emanuel2012mujoco}
Todorov, E.; Erez, T.; and Tassa, Y. 2012.
\newblock MuJoCo: {A} physics engine for model-based control.
\newblock In \emph{2012 {IEEE/RSJ} International Conference on Intelligent
  Robots and Systems, {IROS} 2012, Vilamoura, Algarve, Portugal, October 7-12,
  2012}, 5026--5033. {IEEE}.

\bibitem[{Vecerik et~al.(2019)Vecerik, Sushkov, Barker, Roth{\"o}rl, Hester,
  and Scholz}]{vecerik2019practical}
Vecerik, M.; Sushkov, O.; Barker, D.; Roth{\"o}rl, T.; Hester, T.; and Scholz,
  J. 2019.
\newblock A practical approach to insertion with variable socket position using
  deep reinforcement learning.
\newblock In \emph{2019 International Conference on Robotics and Automation
  (ICRA)}, 754--760. IEEE.

\bibitem[{Warde{-}Farley et~al.(2019)Warde{-}Farley, de~Wiele, Kulkarni,
  Ionescu, Hansen, and Mnih}]{warde2018unsupervised}
Warde{-}Farley, D.; de~Wiele, T.~V.; Kulkarni, T.~D.; Ionescu, C.; Hansen, S.;
  and Mnih, V. 2019.
\newblock Unsupervised Control Through Non-Parametric Discriminative Rewards.
\newblock In \emph{7th International Conference on Learning Representations,
  {ICLR} 2019, New Orleans, LA, USA, May 6-9, 2019}. OpenReview.net.

\bibitem[{Xu et~al.(2020)Xu, Liu, Li, and Loy}]{xu2020knowledge}
Xu, G.; Liu, Z.; Li, X.; and Loy, C.~C. 2020.
\newblock Knowledge Distillation Meets Self-Supervision.
\newblock \emph{arXiv preprint arXiv:2006.07114}.

\bibitem[{Zhang, Yu, and Xu(2021)}]{zhang2021hierarchical}
Zhang, J.; Yu, H.; and Xu, W. 2021.
\newblock Hierarchical Reinforcement Learning By Discovering Intrinsic Options.
\newblock \emph{CoRR}, abs/2101.06521.

\end{thebibliography}

\newpage 
\onecolumn
\section*{Appendix}

\nobreakdash 
Here, we will provide additional experiments (Section \ref{appendix_additional_exp}), theoretical derivations (Section \ref{appendix_dddDerivation}),  implementation details (Section~\ref{appendix_implementation_dddDetails}), and more results with respect the experiments in the main text (Section \ref{appendix-more-results}).  

\section{Additional Experiments}
\label{appendix_additional_exp}

\nobreakdash 
In this section, we make the additional experiments to 
\begin{enumerate} 
	\item[] (\ref{comparsion-diyan-variants})
	compare our method with DIAYN  and its variant, 
	\item[] (\ref{appendix-infer})
	show the generalization when the dynamics and goal conditions are missing, 
	\item[]  (\ref{appendix_perception_loss})
	conduct the ablation study to analyze how the self-supervised loss over the perception-level affects the learned behaviors in terms of the generalization to new tasks,  
	\item[]  (\ref{app-goal-genera})
	explore the (automatic) goal distribution generation, 
	\item [] (\ref{appendix_training_underspecified})
	analyze the effect caused by the underspecified latent (support) space, 
	\item [] (\ref{appendix_training_unfixed})
	lift the Assumption 1 (assuming fixed initial state) in the main text, 
	\item [] (\ref{appendix_training_stochastic_mdp})
	evaluate the robustness of GPIM on stochastic MDP (with varying stochasticity). 
\end{enumerate}

\subsection{Comparison with DIAYN and its Variant}
\label{comparsion-diyan-variants}

In this section, we expect to clarify the difference and connection with DIAYN~\citep{eysenbach2018diversity} experimentally, and indicate the limitations of maximizing $\mathcal{I}(s; \omega)$ and maximizing $\mathcal{I}(s; g)$ separately in learning the goal-conditioned policy.

\begin{figure}[h]
	\centering
	\subfigure[DIAYN-Imitator: $\mathcal{I}(s; \omega)$.]{
		\begin{minipage}{0.23\linewidth}
			\centering
			\includegraphics[scale=0.06]{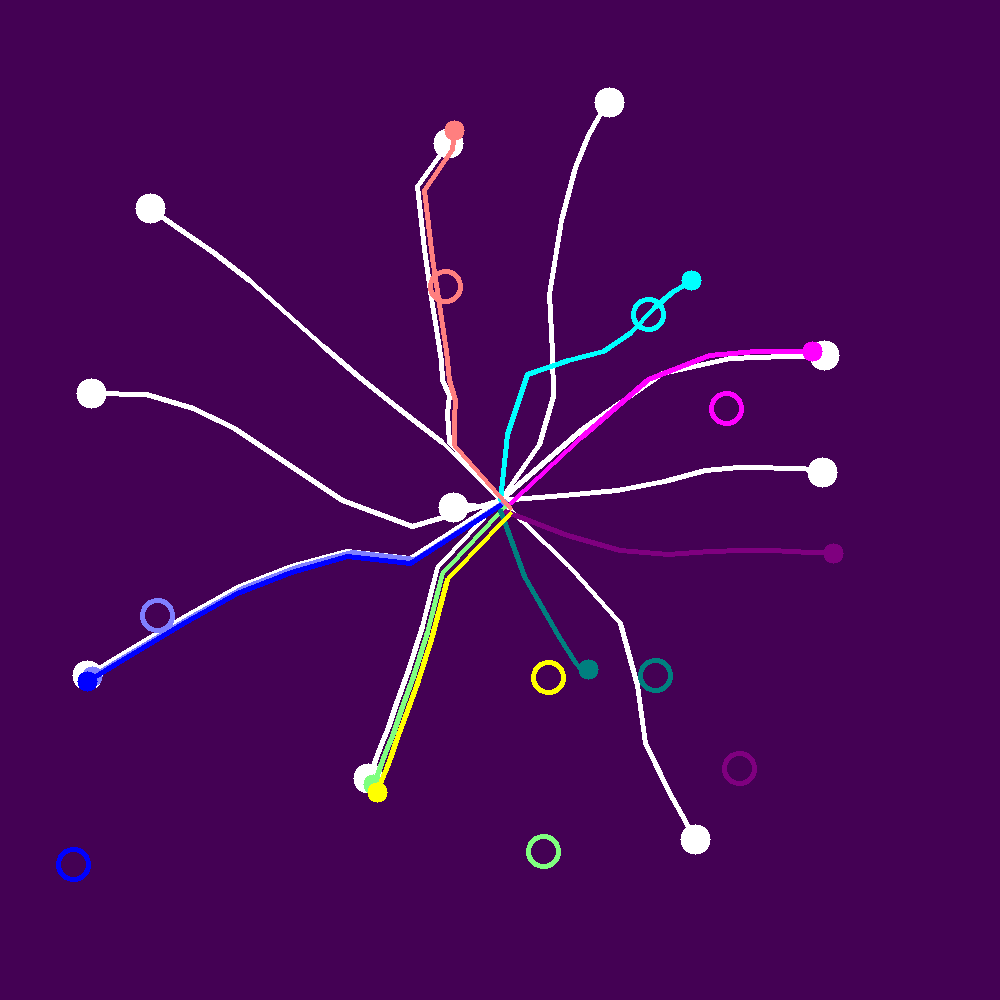}
		\end{minipage}
		\label{app-2-1}
	}
	\subfigure[DIAYN-Goal: $\mathcal{I}(s; g)$.]{
		\begin{minipage}{0.23\linewidth}
			\centering
			\includegraphics[scale=0.06]{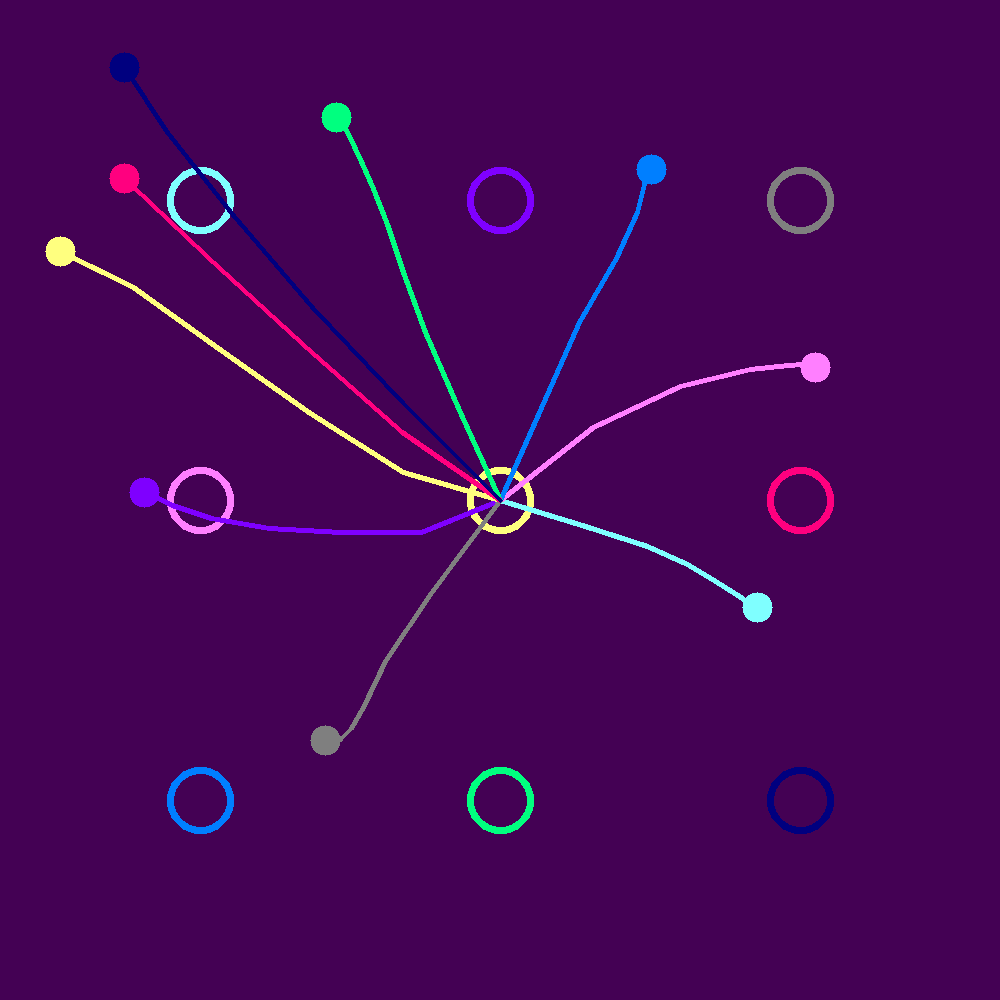}
		\end{minipage}
		\label{app-2-2}
	}
	\subfigure[GPIM: $\mathcal{I}(s; \omega)+\mathcal{I}(\tilde{s}; g)$]{
		\begin{minipage}{0.23\linewidth}
			\centering
			\includegraphics[scale=0.06]{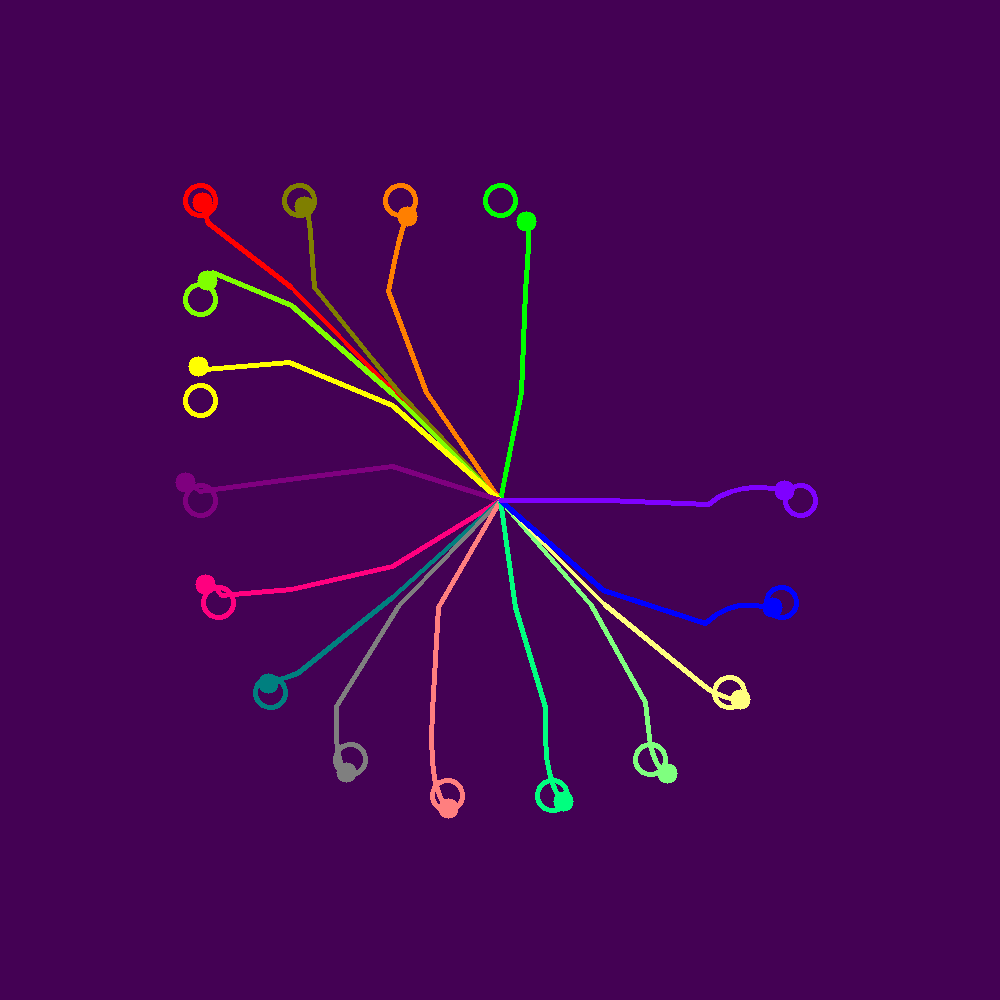}
		\end{minipage}
		\label{app-2-3}
	}
	\subfigure[The learned generative factors $z$ of goals in GPIM. ]{
		\begin{minipage}{0.23\linewidth}
			\centering
			\includegraphics[scale=0.3539]{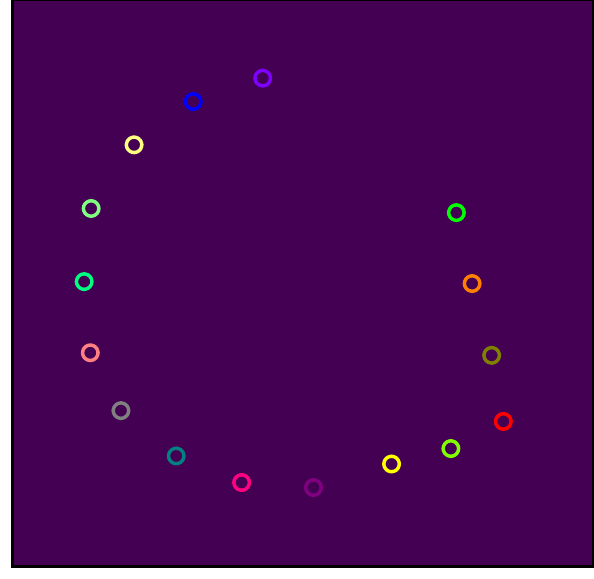}
		\end{minipage}
		\label{app-2-4}
	}
	\caption{ 2D navigation with (a) DIAYN-Imitator, (b) DIAYN-Goal, and (c) our proposed GPIM. In (a), (b) and (c), the open (colored) circle denotes the user-specified goal in test, and the (colored) line with a solid circle at the end is the corresponding behavior. Note that the white lines in~(a) are the learned skills in the training process, where we show 10 diverse skills.  We can see that neither DIAYN-Imitator nor DIAYN-Goal approaches the corresponding goals, while our model succeeds in reaching the goal as shown in~(c). In~(d), we also show the generative factors $z$ (see~Figure~\ref{pimu} in section~\ref{appendix_perception_loss}) by disentangling goals with the self-supervised loss over the perception-level in~(d), where the open circles with different colors in (d) correspond to the goals in~(c).}
	\label{app-2}
\end{figure}

In DIAYN, authors show the ability of the model to imitate an expert. Given the goal, DIAYN uses the learned discriminator to estimate which skill was most likely to have generated the goal $g$: 
\begin{equation}
\hat{\omega} = \arg \max_\omega q_\phi(\omega|g). 
\end{equation}
Here we call this model \emph{DIAYN-Imitator}. 
We also directly substitute the perceptually-specific goals for the latent variable in DIAYN's objective to learn a goal-conditioned policy. We call this model \emph{DIAYN-Goal}:
\begin{equation}
\max \ \  \mathcal{I}(s; g),
\end{equation}
where $g$ is sampled from the prior goal distribution $p(g)$. Please note that we do not adopt the prior non-parametric distance as in DISCERN~\citep{warde2018unsupervised} to calculate the reward. We obtain the reward as in vanilla DIAYN using the variational approximator $q_\phi(g|s)$.

Figure~\ref{app-2} shows the comparison of our GPIM with DIAYN variants, including DIAYN-Imitator and DIAYN-Goal, where the 2D navigation task is considered. 
As observed, DIAYN-Imitator can reach seen goals in training (white lines) but not unseen goals in test in Figure~\ref{app-2-1}, because it cannot effectively accomplish the interpolation between skills that are induced in training. 
And behaviors generated by DIAYN-Goal cannot guarantee consistency with the preset goals in Figure~\ref{app-2-2}. The main reason is that such objective only ensures that when $g$ (or $\omega$) is different, the states generated by $g$ (or $\omega$) are different.
However, there is no guarantee that $g$ (or $\omega$) and the state generated by the current $g$ (or $\omega$) keep semantically consistent. 
Our proposed GPIM method, capable of solving interpolation and consistency issues, exhibits the best performance in this 2D navigation task. 

Moreover, when the user-specified goals are heterogeneous to the states, the learned discriminator $q_\phi$ in DIAYN is unable to estimate which skill is capable of inducing given goals. 
Specifically, when goals are visual inputs, and states in training is feature vectors (e.g., joint angles), the learned discriminator is unable to choose the skills due to a lack of models for converting high-dimensional images into low-dimensional feature vectors. On the contrary, there are lots of off-the-shelf models to render low-dimensional feature vectors into perceptually-specific high-dimensional inputs~\citep{tobin2017domain, khirodkar2018vadra}.

\subsection{Generalization on the Gridworld Task} 
\label{appendix-infer}

Here we introduce an illustrative example of gridworld task and then show the generalization when the dynamics and goal conditions are missing in the gridworld task.

\begin{wrapfigure}{H}{0.385\textwidth}
	\centering
	\begin{center}
		\includegraphics[scale=0.15]{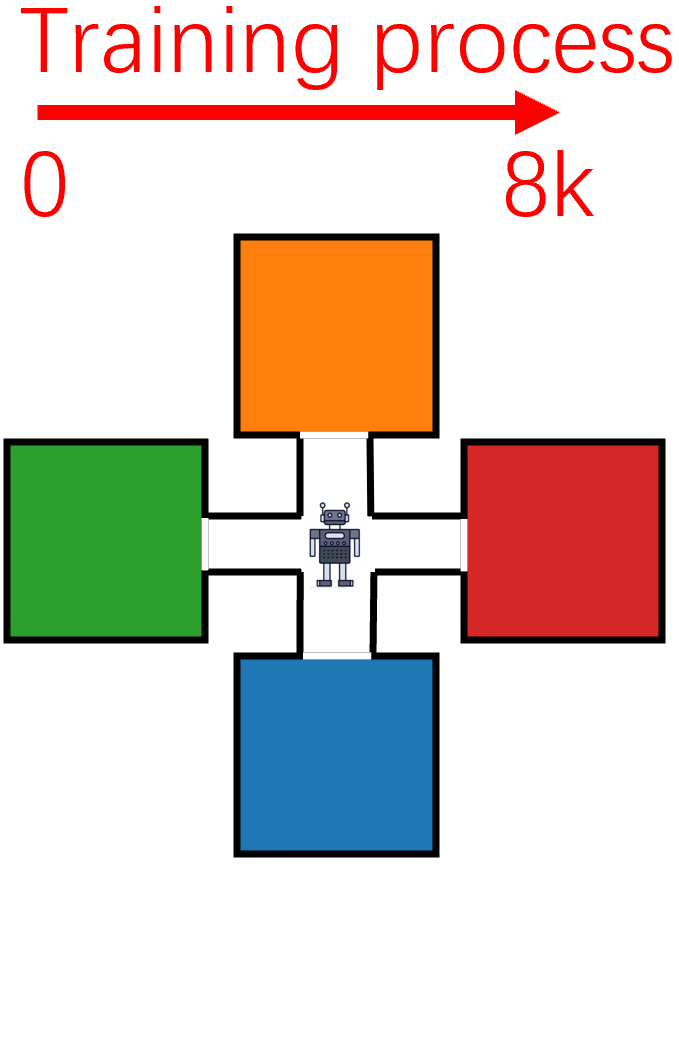} 
		\includegraphics[scale=0.453]{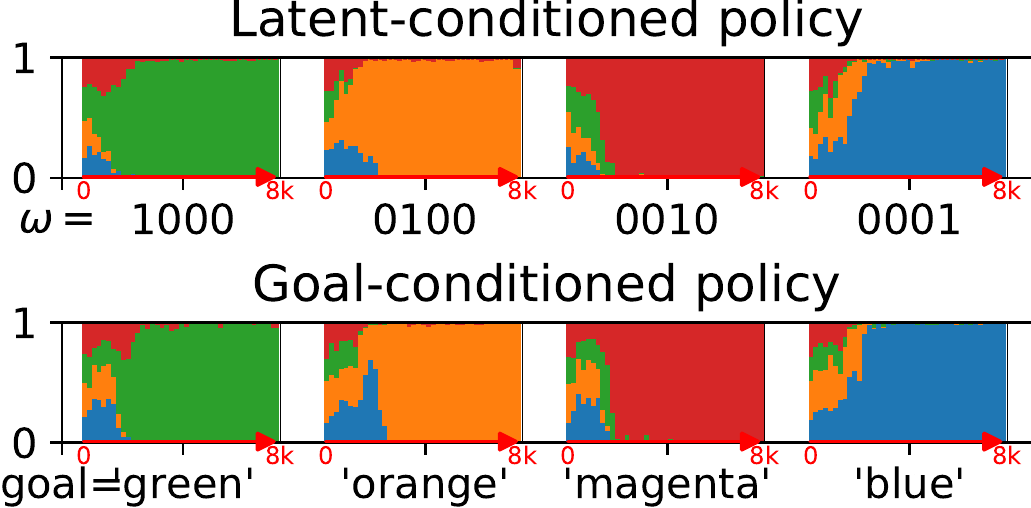}
	\end{center}
	\caption{Gridworld tasks. Y-axis is the percentage of different rooms that the robot arrives in.}
	\label{arrowe}
	\vspace{-7pt}
\end{wrapfigure}
\textbf{Illustrative example.} 
Here, we start with a simple gridworld example: the environment is shown on the left of Figure~\ref{arrowe}, where the goal for the agent is to navigate from the middle to the given colored room. 
By training the latent-conditioned policy and the discriminator, our method quickly acquire four skills ($\omega=1000; 0100; 0010; 0001$) to reach different rooms. Each time the agent arrives in a room induced by $\pi_\mu$, 
we train $\pi_\theta$ conditioned on the room's color (e.g., green), allowing the agent to be guided to the same room by the current reward function (e.g. $q_\phi$ conditioned on $\omega=1000$). 
The results are shown on the right of Figure~\ref{arrowe}, where the upper and lower subfigures show the learning processes of latent-conditioned policy $\pi_\mu$ and goal-conditioned policy $\pi_\theta$ respectively. 
It is concluded that the agent can automatically learn how to complete tasks given semantic goals in an unsupervised manner.

\begin{figure*}[h]
	\centering	
	\includegraphics[scale=0.202]{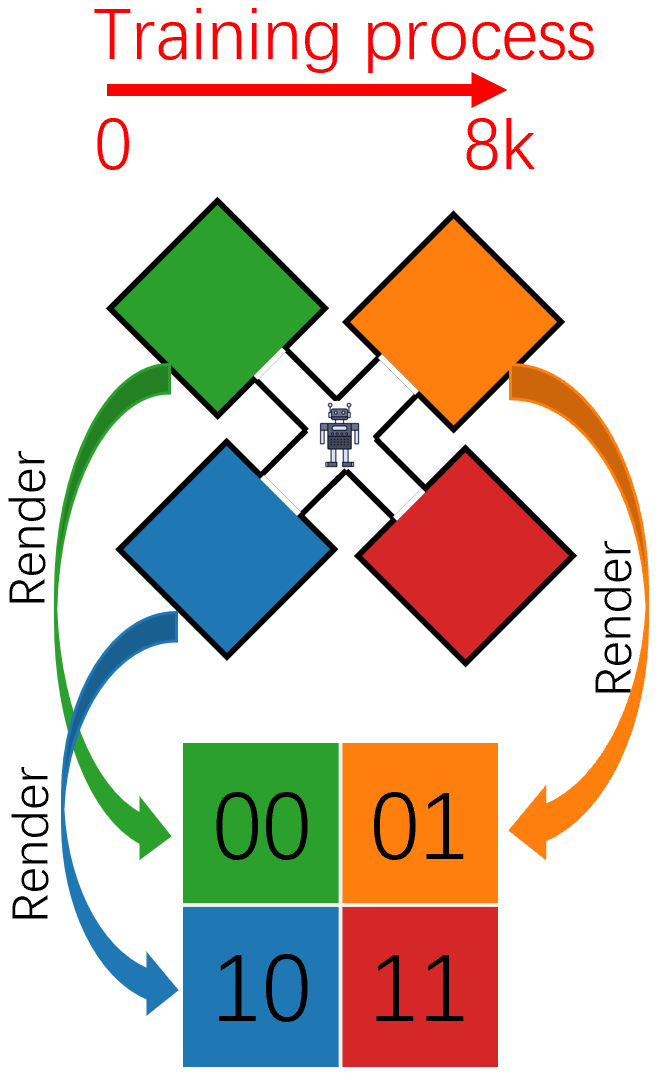} \ \ \ \   
	\includegraphics[scale=0.4057]{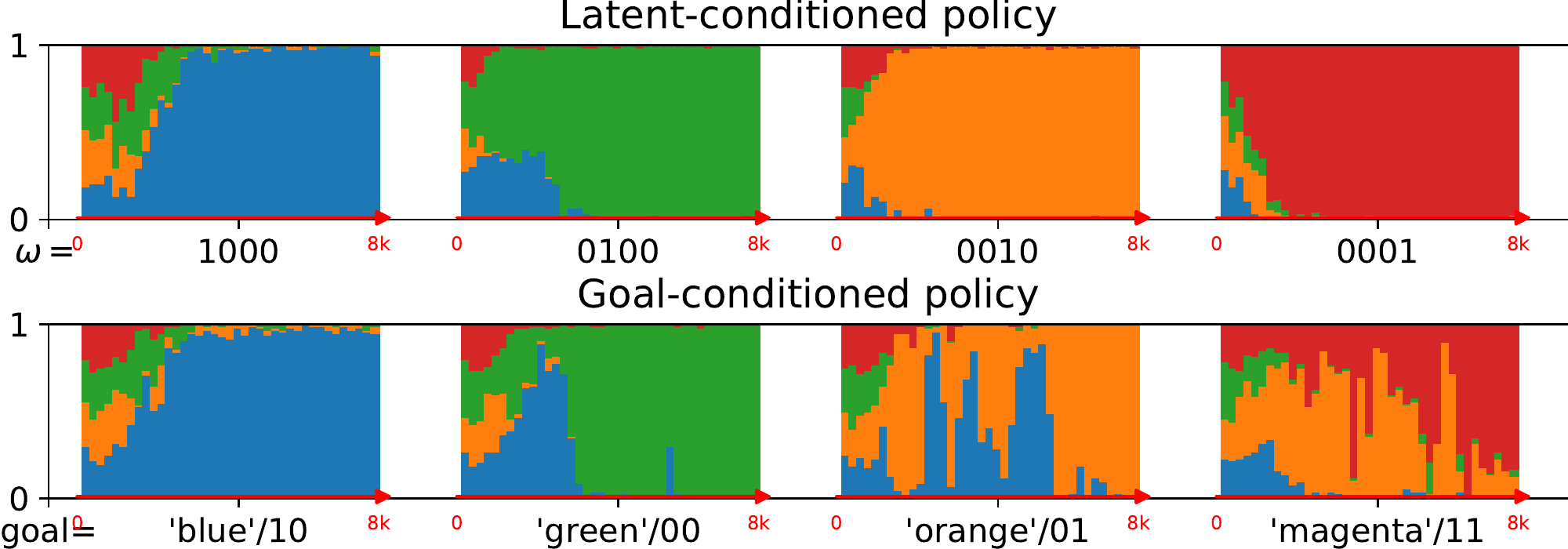}
	\caption{Gridworld tasks. The Y-axis is the percentage of different rooms that agent arrives in.} 
	\label{fig-app-infer}
	\vspace{-5pt}
\end{figure*}
\textbf{Generalization on the gridworld task.}
We further show the generalization when the dynamics and goal conditions are missing in the gridworld task, where there are four rooms in different colors: blue, green, orange and magenta. We consider the situation where we train the goal-conditioned policy in the blue room, the green room and the orange room, and test in the magenta room.

As shown in Figure~\ref{fig-app-infer}, we quickly acquire four skills for different rooms through the training at the abstract level. After the agent reaches at the colored rooms, we relabel the corresponding room as a two-bit representation: $f_\kappa(\textit{'blue'}) = 10$, $f_\kappa(\textit{'green'}) = 00$, and $f_\kappa(\textit{'orange'}) = 01$ ; we do not relabel the magenta room. 
We take the magenta room corresponding to $11$ as the test goal to verify the generalization ability. 

In lower part of Figure~\ref{fig-app-infer}, we show the learning process of the goal-conditioned policy. We can find that the blue room task is completed quickly, and the green and orange room tasks are learned relatively slowly, and the agent is still able to complete the (new) test task successfully (the magenta room). Compared to the task in Figure~\ref{arrowe} that relabel all the rooms as goals, the whole learning process in Figure~\ref{fig-app-infer} is much slower. 
We hypothesize that the main reason is that the agent needs to further infer the relationship between different goals. The lack of goal information (i.e., missing magenta $11$) leads to a lower efficiency and unstable training.

\begin{wrapfigure}{r}{0.385\textwidth}
	\centering
	\begin{center}
		\includegraphics[scale=0.75]{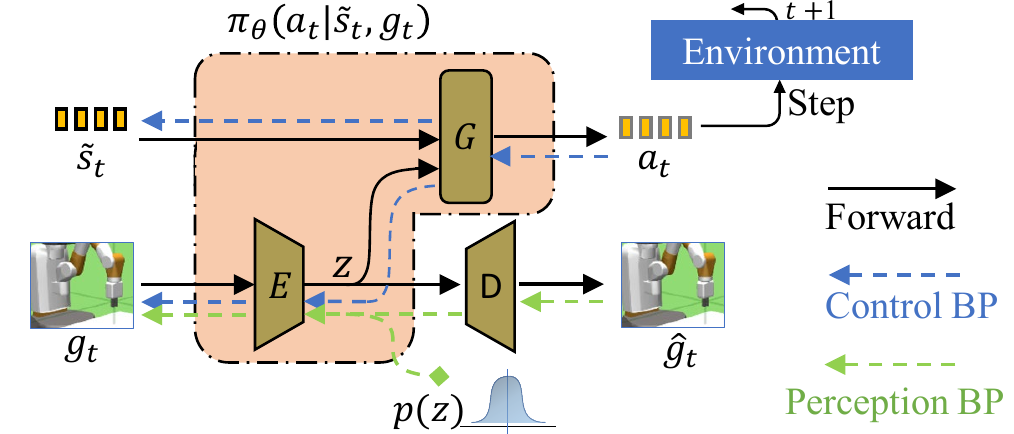}
	\end{center}
	\caption{Optimization for the goal-conditioned policy $\pi_\theta$ with the control-level loss (SAC with reward $\tilde{r}_t$) and perception-level loss.}
	\label{pimu}
\end{wrapfigure}
\subsection{Self-Supervised Loss over the Perception-Level}
\label{appendix_perception_loss}


As indicated by \citet{srinivas2020curl}, control algorithms built on top of useful semantic representations should be significantly more efficient. 
We further optimize the goal-conditioned policy $\pi_\theta$ with an associated self-supervised loss on the perception-level, so as to improve the learning efficiency and generalization ability of the goal-conditioned policy $\pi_\theta$ facing high-dimensional goals. 
As shown in Figure~\ref{pimu}, we decompose $\pi_\theta$ into two components: the encoder network E parameterized by $\vartheta_E$ and the generative network G parameterized by $\vartheta_G$. 
Hence, we optimize the overall objective function for the goal-conditioned policy $\pi_\theta$: 
\begin{equation}\label{eq12}
\max_{\vartheta_E,\vartheta_G,\vartheta_D} \ \  \mathbb{E}_{p_{m}(\cdot)} \left[\log q_\phi({\omega}|\tilde{s})\right] + Prec\_loss, 
\end{equation}
where the self-supervised loss on the perception-level $Prec\_loss$ includes the reconstruction loss and a prior matching loss (with an isotropic unit Gaussian prior $p(z) = N(0; I)$): 
$$
Prec\_loss \triangleq 
\underbrace{\alpha \cdot \mathbb{E}_{p(g)p_{\vartheta_E}(z|g)} \left[ p_{\vartheta_D}(g|z) \right]
}_{\text{reconstruction loss}} 
\underbrace{- \beta \cdot \mathbb{E}_{p(g)} \left[ \textit{KL}(q_{\vartheta_E}({z}| g) || p({z}) ) \right]
}_{\text{prior matching loss}}, 
$$ 
where 
$p(g) = p(\omega)p(s|\omega;\mu)p(g|s;\kappa)$, 
$\alpha$ and $\beta$ are two hyperparameters. 
It is worth noting that the update of the encoder E is based on gradients from both $\pi_\theta$ and \emph{Prec\_loss}.

\begin{figure}[h]
	\centering
	\subfigure[Ablation study on 2D Navigation (x-y goal). Top-subfig: the normalized distance to goals vs. the actor steps.]{
		\begin{minipage}{0.40\linewidth}
			\centering
			\includegraphics[scale=0.58]{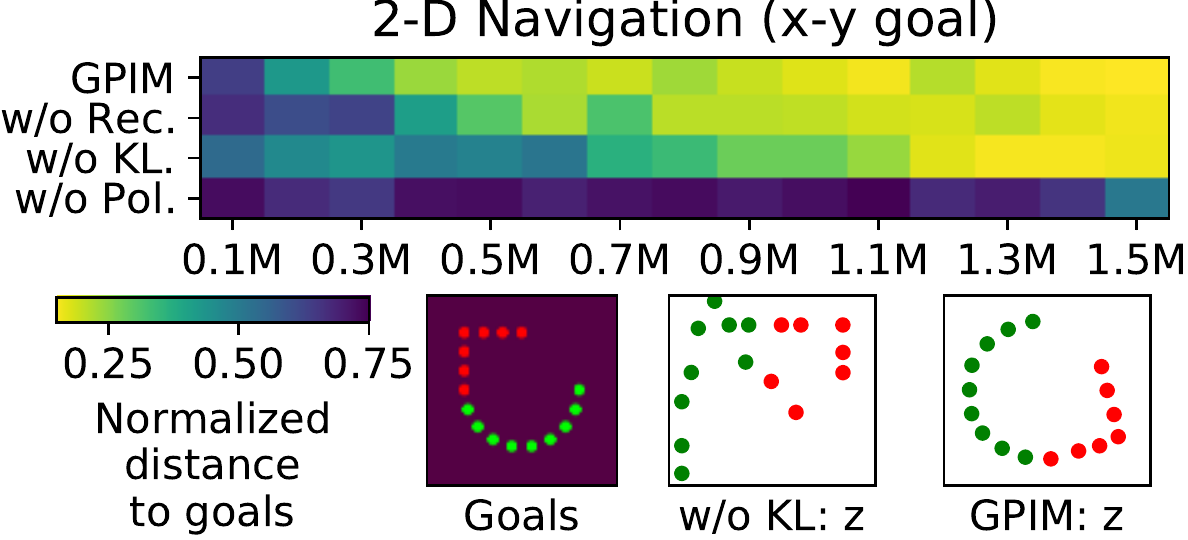}
		\end{minipage}
		\label{abst}
	}
	\ \ \ \ \ 
	\subfigure[Ablation study on mujoco tasks (half cheetah, swimmer and fetch): the normalized distance to goals vs. the actor steps.]{
		\begin{minipage}{0.50\linewidth}
			\centering
			\includegraphics[scale=0.40]{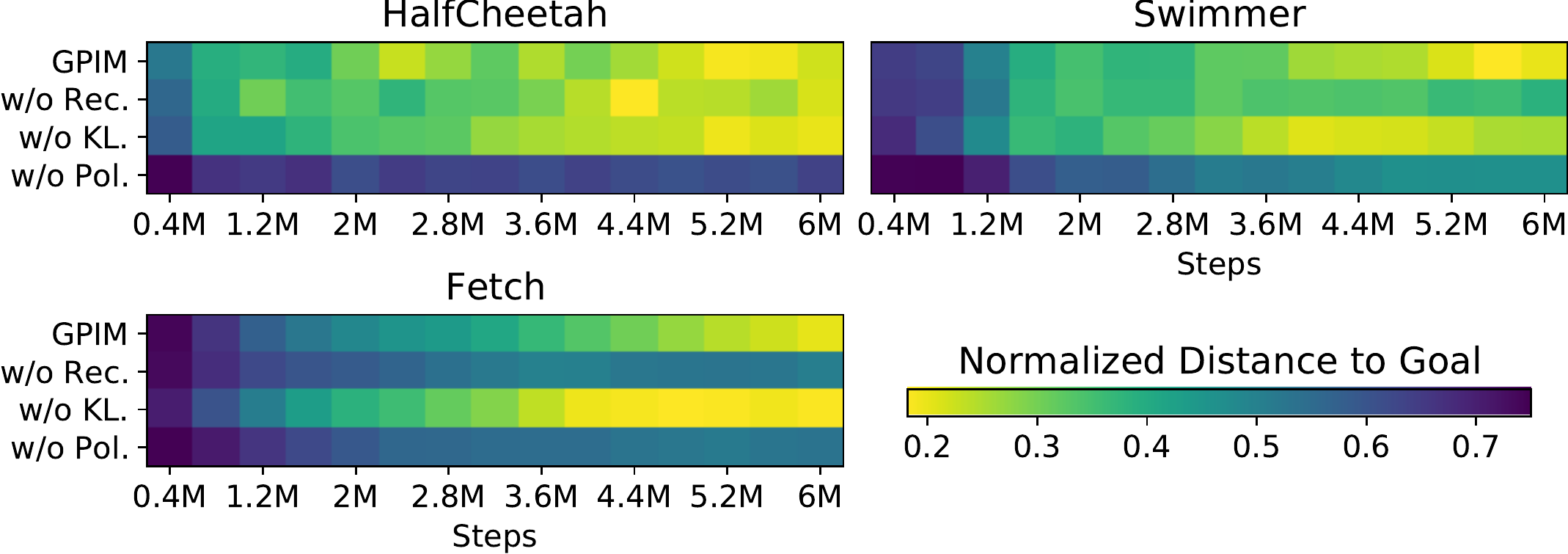}
		\end{minipage}
		\label{app-3-heatmap}
	}
	\caption{Ablation studies for the perception loss.}
\end{figure}
\textbf{Ablation study on 2D navigation.}
Here we conduct the ablation study on 2D navigation task to analyze how the perception-level loss in GPIM
affects the learned behaviors in terms of the generalization to new tasks. 
For convenience, we remove certain component from GPIM and define the new method as follows: 
\emph{w/o~Rec} - removing the reconstruction loss (i.e. $\alpha=0$); 
\emph{w/o~KL} - removing the KL loss (i.e. $\beta=0$); 
\emph{w/o~Pol} - removing the policy loss updating $\vartheta_E$ as shown in Figure~\ref{pimu}.

The performance is reported in Figure~\ref{abst}.
It is observed that \emph{w/o.~Pol} performs worse than all other methods, which is consistent with the performance of RIG that trains VAE and policy separately. 
The main reason is that the perception-level loss fails to figure out the required latent factors on given tasks. 
Moreover, although GPIM has a similar performance with the other three methods on 2D navigation task, GPIM has better interpretability to behaviors. 
As shown in Figure~\ref{abst} (bottom), considering a series of goals from the first red to the last green (\emph{left}) in a counterclockwise order, GPIM can successfully disentangle them and learn effective latent $z$ (\emph{right}), but \emph{w/o~KL} fails to keep the original order of goals (\emph{middle}).


\textbf{Ablation study on mujoco tasks.} 
We further study the impact of perception-level loss on mujoco tasks. 
The performance is reported in Figure~\ref{app-3-heatmap}. It is observed that \emph{w/o Pol} performs worse than all other methods, which is consistent with the performance in 2D navigation task. 
And we can find that when we remove the reconstruction loss ($\alpha = 0$), the performance of \emph{w/o~Rec} degrades in these three environments. The main reason is that the process of learning generative factors become more difficult without the supervised reconstruction loss. While in 2D navigation task, the reconstruction loss  has little impact on the performance. 
Even though that \emph{w/o~KL} has a similar performance with our full GPIM method, GPIM demonstrates better interpretability to behaviors as shown in Figure~\ref{app-2-4}.

\subsection{Automated Goal-Generation for Exploration} 
\label{app-goal-genera}

\begin{figure*}[h]
	\centering
	\subfigure[Decoded latent goals (RIG).]{
		\begin{minipage}{0.98\linewidth}
			\centering
			\includegraphics[scale=0.35]{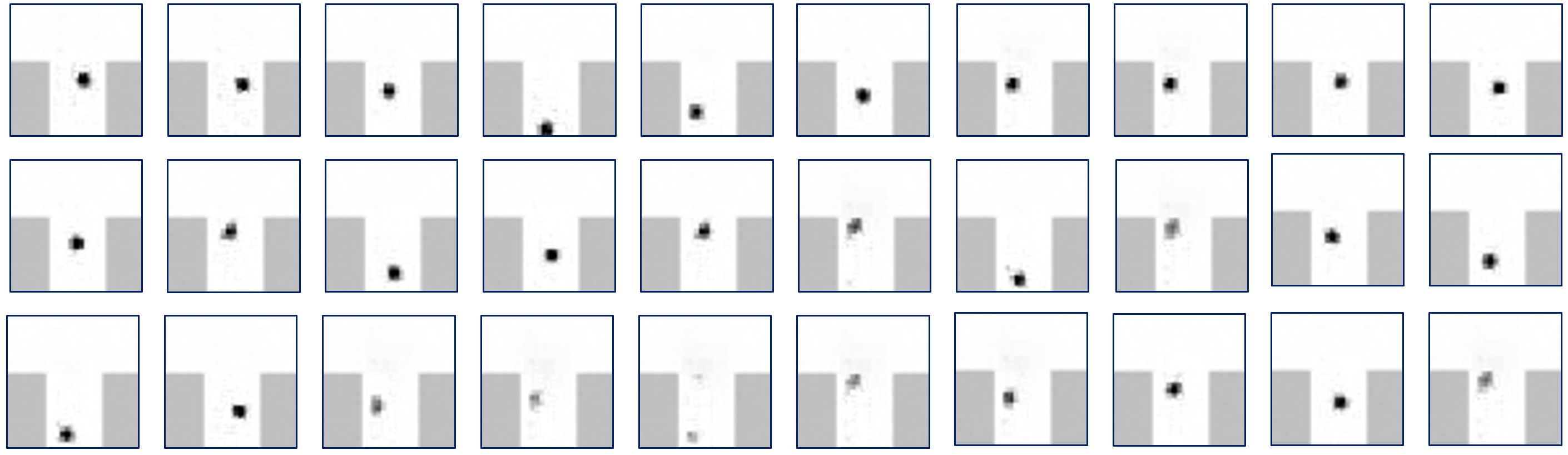}
		\end{minipage}
		\label{app-1-goals-rig}
	}
	\subfigure[Goals (DISCERN).]{
		\begin{minipage}{0.25\linewidth}
			\centering
			\includegraphics[scale=0.40]{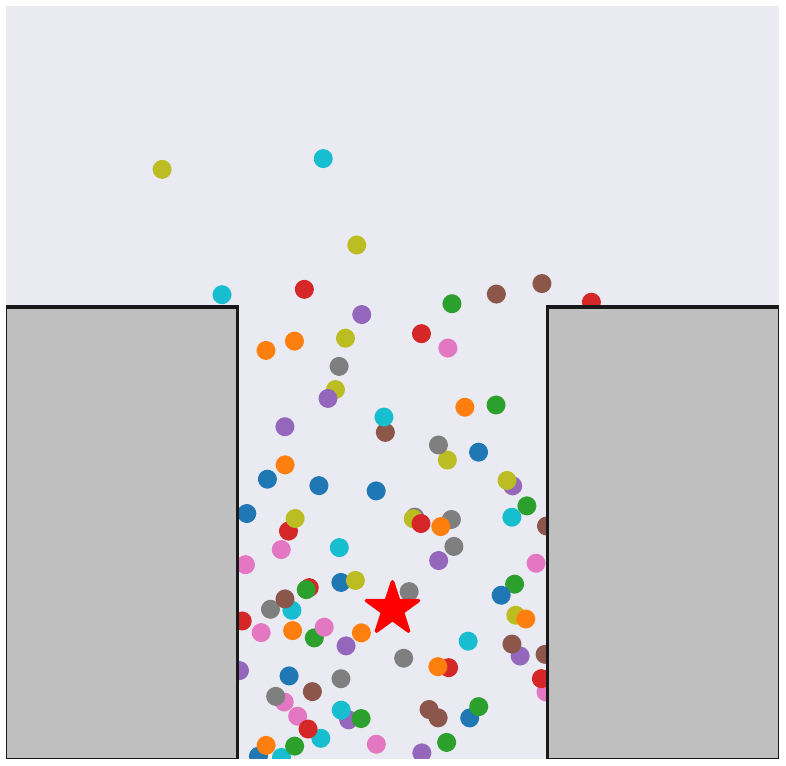}
		\end{minipage}
		\label{app-1-goals-discern}
	}
	\subfigure[Goals (GPIM).]{
		\begin{minipage}{0.25\linewidth}
			\centering
			\includegraphics[scale=0.40]{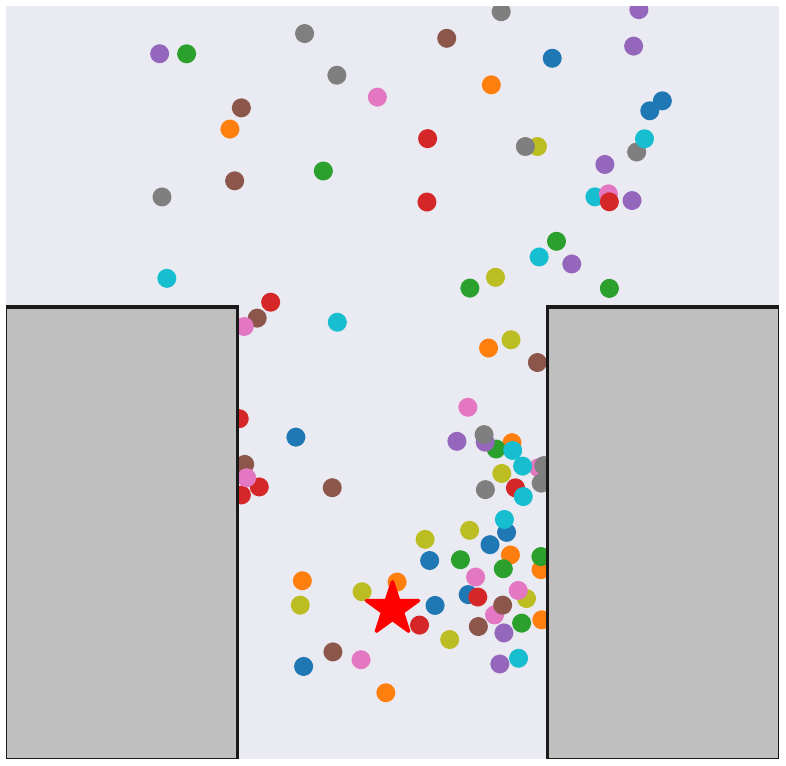}
		\end{minipage}
		\label{app-1-goals-gpim}
	}
	\subfigure[Performance.]{
		\begin{minipage}{0.45\linewidth}
			\centering
			\includegraphics[scale=0.435]{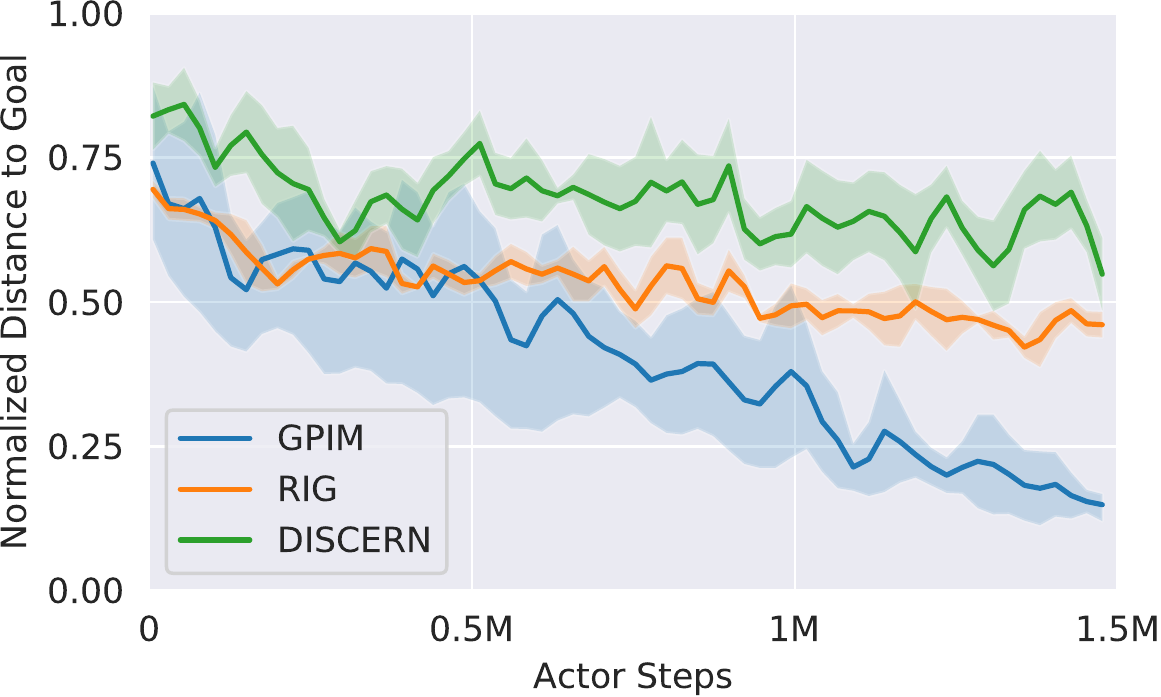}
		\end{minipage}
		\label{app-1-goalslearn}
	}
	\label{app-1-goalss}
	\caption{Distribution of sampled goals for the goal-conditioned policy, where the initial state is [5,2] (the red star) and the actor step is 20 for each rollout. (a) The goals (images) are obtained by decoding 30 sampled latent goals $z_g$ in VAE framework; (b) The goals (colored dots) are sampled from the agent's behaviors by random exploration; (c) The goals (colored dots) are relabeled from the states induced by our latent-conditioned policy. (d): Evaluation on reaching user-specified goals, where GPIM significantly outperforms baselines. }
\end{figure*}

In general, for unsupervised RL, we would like to ask the agent to carry out autonomous "practice" during training phase, where we do not know which particular goals will be provided in test phase. 
In order to maximize the state coverage, the ability to automatically explore the environment and discover diverse goals is crucial. In this section, we will further analyze the goal distribution of three methods (RIG~\citep{nair2018visual}, DISCERN~\citep{warde2018unsupervised}, GPIM) in a new 2D obstacle navigation task as shown in Figure~\ref{app-1-goals-rig}, Figure~\ref{app-1-goals-discern}, and Figure~\ref{app-1-goals-gpim}. The size of the environment is $10 \times 10$, the initial state is set as $[5,2]$, and there are two obstacles that prevent the agent from passing through each of which is $3 \times 6$ in size.

DISCERN samples goals during training by maintaining a fixed sized buffer $\mathcal{G}$ of past observations. We simply mimic the process of goal generation by taking random actions for 20 environment steps. As in Figure~\ref{app-1-goals-discern}, we generate $100$ goals with different colors. We can see that the majority of goals locate between the two obstacles, which limits the further exploration of the environment. 

RIG samples a representation (latent goals $z_g$) from the learned VAE prior, which represents a distribution over latent goals. The policy network takes the representation as a substitute for the user-specified goal. For a clear visualization of the distribution of the sampled latent goals, we further feed the sampled latent goals into the decoder to obtain the real goals in the user-specified goal space. The decoded latent goals are shown in Figure~\ref{app-1-goals-rig}, where we sample 30 goals. It is shown that the majority of goals are also between the two obstacles because the goals for training the VAE prior come from the same distribution as in DISCERN.

Our method, GPIM, obtains goals from the behaviors induced by the latent-conditioned policy. Maximizing $\mathcal{I}(s; \omega)$ encourages different skills to induce different states that are further relabeled to goals. This objective ensures that each skill individually is distinct and the skills collectively explore large parts of the state space~\citep{eysenbach2018diversity}. As shown in Figure~\ref{app-1-goals-gpim}, our method provides better coverage of the state space than DISCERN and RIG. 

Figure~\ref{app-1-goalslearn} shows the performance of the three methods, where we randomly sample goals from the whole state (or goal) space at test phase. We can see that our method significantly outperforms the baselines. The most common failure mode for prior methods is that the goal distribution collapses~\citep{pong2019skew}, causing that the agent can reach only a fraction of the state space, as shown in Figure~\ref{app-1-goals-rig} and \ref{app-1-goals-discern}. 

Exploration is a well-studied problem in the field of RL, and there are many proven approaches with different benefits to improve the exploration~\citep{campos2020ExploreDA, colas2018gep, Pitis2020MaximumEG}. 
Note that these benefits are orthogonal to those provided by our straightforward GPIM, and these approaches could be combined with GPIM for even greater effect. We leave combing our method with sophisticated exploration strategies to future work.

\subsection{Training with Underspecified Latent (Support) Space}
\label{appendix_training_underspecified}

Here we show that underspecified latent (support) space may induce a large performance gap for learning the goal-conditioned policy. 
Qualitatively, Theorem 1 explains the performance gap induced by $p(s|\omega,\mu*)$ and highlights that such gap is sensitive to the prior latent distribution $p(\omega)$ . We thus add additional experiments on the distribution of skills on 2D Navigation, Object manipulation and Fetch tasks, as shown in Table~\ref{app:tab:underspecified-omega}.

\begin{table}[htbp]
	\centering
	\caption{2D Nav.: taks 2D navigation; task object manipulation.}
	\begin{tabular}{llllll}
		\toprule
		& Discrete; 5 skills & Discrete; 10 skills & Discrete; 50 skills & Discrete; 100 skills & Continuous \\
		\midrule
		2D Nav. & 0.296 (+/- 0.091) & 0.152(+/- 0.066) & 0.137 (+/- 0.052) & 0.132 (+/- 0.037) & 0.122 (+/- 0.044) \\
		O-M & 0.535 (+/- 0.110) & 0.321 (+/- 0.082) & 0.199 (+/- 0.057) & 0.193 (+/- 0.064) & 0.182 (+/- 0.060) \\
		Fetch & 0.376 (+/- 0.175) & 0.320 (+/- 0.104) & 0.257 (+/- 0.089) & 0.162 (+/- 0.057) & 0.149 (+/- 0.033) \\
		\bottomrule
	\end{tabular}%
	\label{app:tab:underspecified-omega}%
\end{table}%

We found that the number of skills does affect the performance of the goal-conditioned policy. In simple tasks (2D navigation task), the effect is relatively small. But in complex environments, this effect cannot be ignored. We analyze the generated goal distributions in these experiments with different numbers of skills and found that goals generated by a small number of skills are extremely sparse, leading to the bias of the goal-conditioned policy. Nevertheless, such bias can be reduced, to some extend. As shown in 2D navigation task, learning with 10 skills can still produce comparable results. We attribute this result to the policy's generalization ability and the perception-level loss (Appendix~\ref{appendix_perception_loss}).

\subsection{Training with Unfixed Initial State}
\label{appendix_training_unfixed}

In Assumption (the main text), we assume that the initial state of the environment is fixed.
For granting valid training for each rollout of the goal-conditioned $\pi_\theta$, we need to keep the initial state fixed or filter out initial states (for goal-conditioned $\pi_\theta$, see Line 14 in Algorithm 1) that never lead to the relabeled goals, eg. constraining $\text{Distance}(\tilde{s}_0, s_0) \leq \epsilon$.

\begin{figure}[H]
	\centering
	\includegraphics[scale=0.1253]{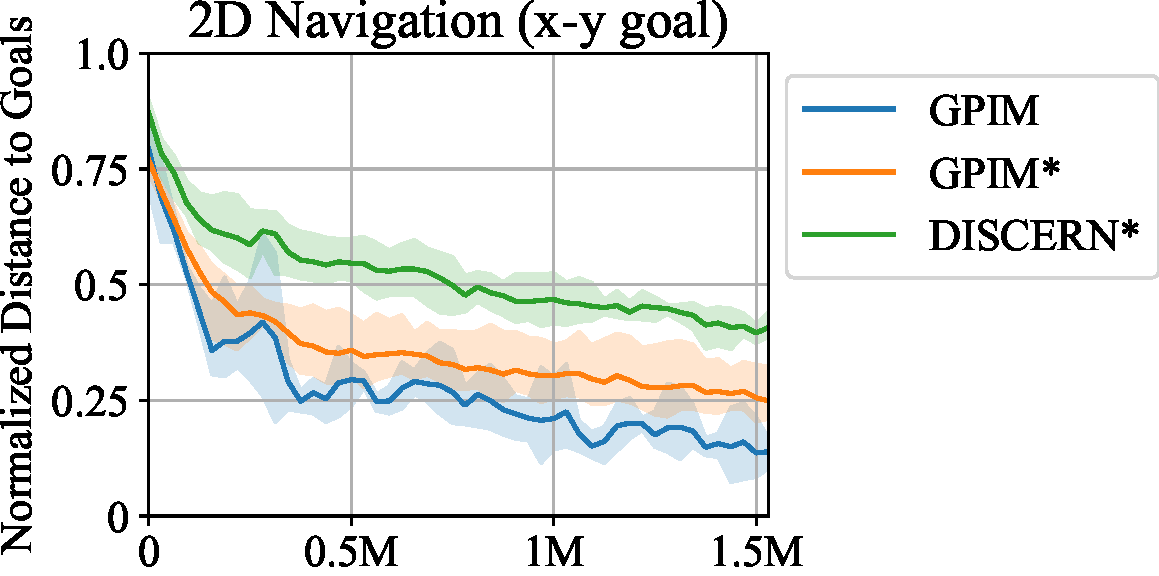} \ \ \ \ \ \
	\includegraphics[scale=0.1253]{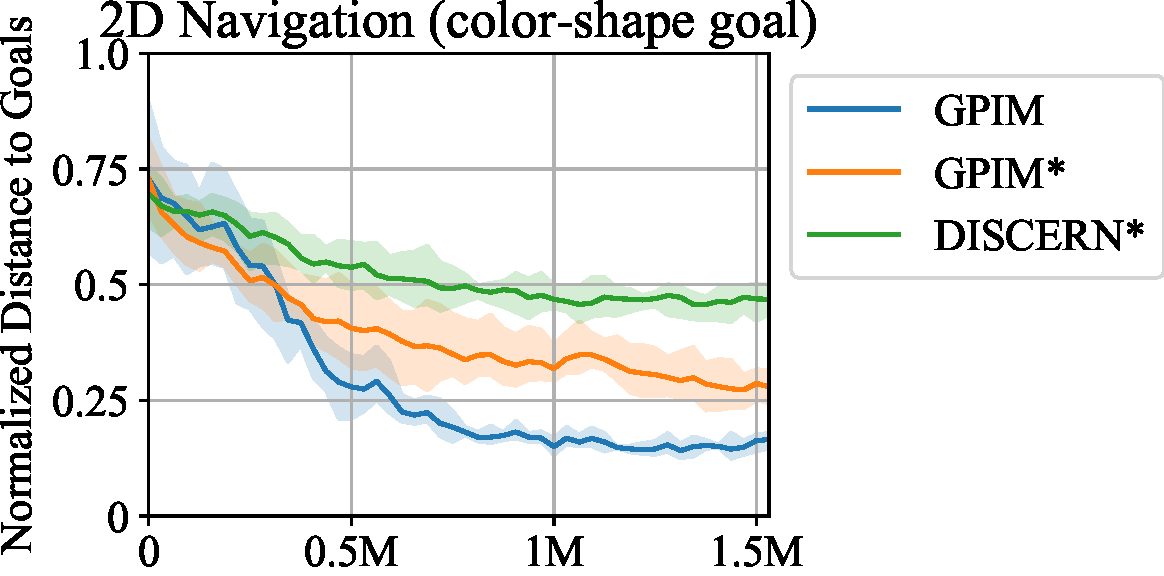}
	\caption{GPIM: training GPIM with the fixed initial state [0,0].
		GPIM*: training GPIM with the initial state distribution (uniformly sampled from {[0,0], [-0.25, -0.25], [-0.25, 0.25], [0.25, -0.25], [0.25, 0.25]}). 
		DISCERN*: training DISCERN with the initial state distribution (uniformly sampled from {[0,0], [-0.25, -0.25], [-0.25, 0.25], [0.25, -0.25], [0.25, 0.25]}). }
	\label{app:fig:nonfixed_state_2dnavi}
\end{figure}

However, although randomly sampling initial states for $\pi_\mu$ and $\pi_\theta$ from the initial distribution (more than one fixed state) can not guarantee that the coming training (for $\pi_\theta$) must be valid (goals are reachable), a large number of samples will ensure that there exist cases where initial states obtained by two consecutive resets is close, which can enable valid training. We also validate such setting (more than one fixed initial state) in 2D navigation (x-y goal) and 2D navigation (color-shape goal) tasks, as shown in Figure~\ref{app:fig:nonfixed_state_2dnavi}. We can observe that 1) such performance is reduced compared to the fixed starting point setting, and 2) the performance of our method outperforms that of DISCERN with the same initial state distribution (more than one fixed state).

\subsection{The Robustness of GPIM on Stochastic MDP}
\label{appendix_training_stochastic_mdp}

Here we show the robustness of our GPIM on stochastic MDP. 
We add random noise to the dynamics of the environment 2D navigation, so as to build the stochastic MDP: $p(s'|s,\pi(s)) \to p(s'|s,\pi(s)+\text{action-space.sample()} \cdot \Delta)$, where we sample an random action from the action space and then weigh the action and the action output from the policy with $\Delta$. Thus, we can use  $\Delta$ to represent the stochasticity of the environment.

\begin{figure}[H]
	\centering
	\includegraphics[scale=0.1253]{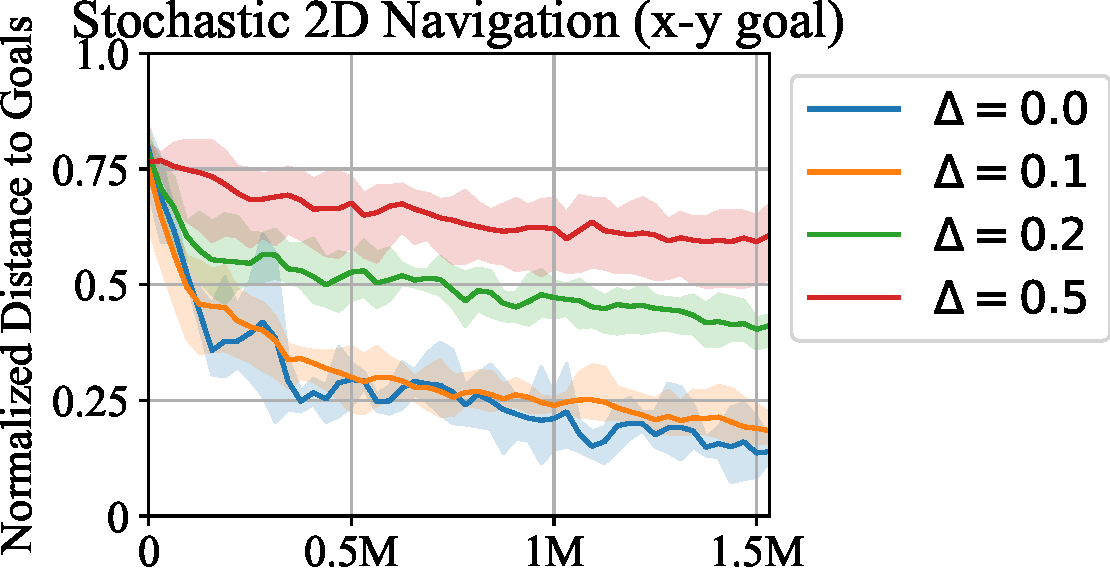} \ \ \ \ \ \
	\includegraphics[scale=0.1253]{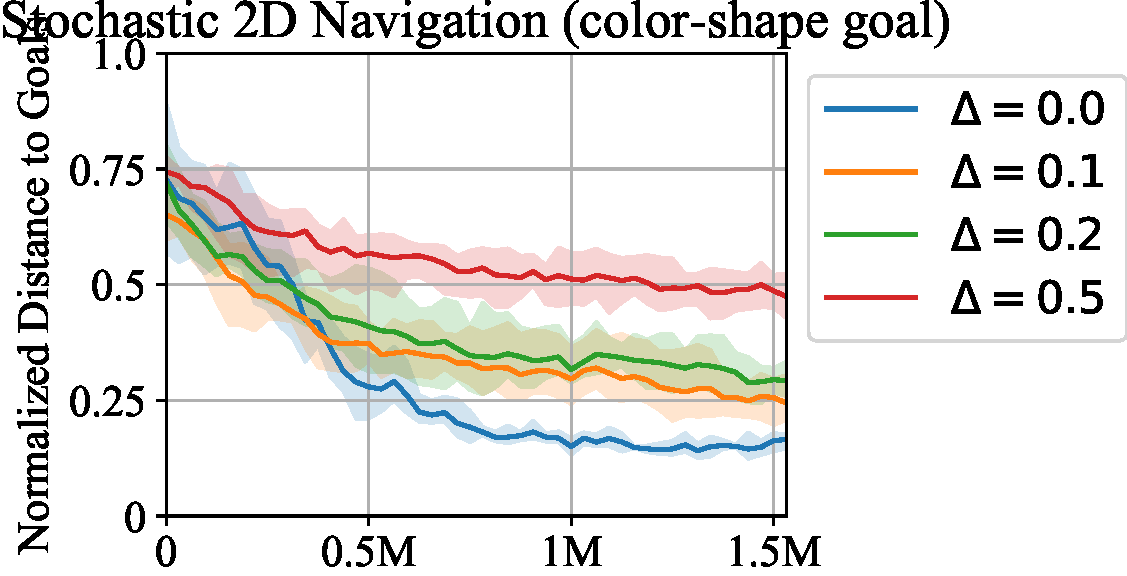}
	\caption{Evaluation on the stochastic MDP.}
	\label{app:fig:stochastic_2dnavi}
\end{figure}

We shown our results in the stochastic environments in Figure~\ref{app:fig:stochastic_2dnavi}. We can find that when  equals 0.1, the learning process is hardly affected. As the stochasticity of the environment increases, the performance of the method decreases. When D equals 0.5, the method barely completes the task.

\section{Derivation}
\label{appendix_dddDerivation}

\subsection{Derivation of the Variational Bound}
\label{appendix-derivation}
\begin{eqnarray}
&\ & \mathbb{E}_{p_m(\cdot)} \left[ \log p({\omega}|{s}) + \log p({\omega}|{\tilde{s}}) - 2 \log p({\omega}) \right] 
\nonumber \\ 
&=& \mathbb{E}_{p_m(\cdot)} \left[\log q_\phi({\omega}|{s})  + \log q_\phi({\omega}|{\tilde{s}}) - 2\log p({\omega}) \right] \nonumber \\
&\ & \  \quad \quad 
+ \underbrace{\mathbb{E}_{p_m(\cdot)} \left[ KL(p({\omega}|{s})||q_\phi({\omega}|{s})) +  KL(p({\omega}|{\tilde{s}})||q_\phi({\omega}|{\tilde{s}})) \right]}_{\geq 0 }
\nonumber \\ 
&\geq& \mathbb{E}_{p_m(\cdot)} \left[\log q_\phi({\omega}|{s})  + \log q_\phi({\omega}|{\tilde{s}}) - 2\log p({\omega}) \right] \nonumber 
\end{eqnarray}

\subsection{Proof of Theoretical Guarantees}
\label{appendix-derivation-analysis}

Note that $$\eta(\pi_\theta) = \mathbb{E}_{p'(g,\tilde{s}; \theta)}\left[ \log p'(g|\tilde{s}) - \log p'(g) \right],$$ 
and $$\hat{\eta}(\pi_\theta) = \mathbb{E}_{p(\omega, s, g, \tilde{s}; \mu^*, \kappa, \theta)}\left[ \log p(\omega|\tilde{s}) - \log p(\omega) \right],$$ where $\mu^* = \arg \max_{\mu} \mathcal{I}({s}; {\omega})$. For deterministic $\pi_\mu$, we have  $\mathbb{E}_{p(\omega,s;\mu^*)}\log p(s|\omega;\mu^*)=0$.

Thus, we have: 
\begin{align}
&\ \ \ \ \hat{\eta}(\pi_\theta) - \eta(\pi_\theta)  
\quad \quad \quad \quad  
\text{\textcolor{gray}{\# $p^*_{m}(\cdot) \triangleq p(\omega, s, g, \tilde{s}; \mu^*, \kappa, \theta)$}}
\nonumber \\
&= \mathbb{E}_{p_{m}^*(\cdot)}\left[ \log p(\omega|\tilde{s}) - \log p(\omega) \right] -  \mathbb{E}_{p'(g,\tilde{s}; \theta)}\left[ \log p'(g|\tilde{s}) - \log p'(g) \right] 
\nonumber \\
&= \underbrace{\mathbb{E}_{p_{m}^*(\cdot)}\left[ \log p(\omega|\tilde{s}) - \log p(\omega) - \log p'(g|\tilde{s}) + \log p'(g) \right] }_{\text{TERM-I}}
\nonumber \\
&\ \ \  + 
\underbrace{\mathbb{E}_{p_{m}^*(\cdot)}\left[ \log p'(g|\tilde{s}) - \log p'(g) \right] 
	- \mathbb{E}_{p'(g,\tilde{s}; \theta)}\left[ \log p'(g|\tilde{s}) - \log p'(g) \right] }_{\text{TERM-II}}.
\nonumber  
\end{align}

\textbf{Special case:} 
If we assume the prior goal distribution $p'(g)$ for optimizing $\eta(\pi_\theta)$ matches the goal distribution $p(g|\omega; \mu^*, \kappa)$ induced by $\pi_{\mu^*}$ and $f_\kappa$ for optimizing $\hat{\eta}(\pi_\theta)$, e.g., $\mathbb{E}_\omega\left[ p(g|\omega;\mu^*,\kappa)\right] = p'(g) $, thus we have $p_{m}^*(\cdot)\triangleq p(\omega, s; \mu^*) p(g, \tilde{s}|s; \kappa, \theta)=p(\omega, s; \mu^*) p'(g,\tilde{s}; \theta)$. 
By inserting this equation into $\hat{\eta}(\pi_\theta) - \eta(\pi_\theta)$, we obtain 
$$\text{TERM-II} = 0,$$ and 
\begin{align}
\text{TERM-I} 
&= \mathbb{E}_{p_{m}^*(\cdot)}\left[ \log p(\omega|\tilde{s}) - \log p(\omega) - \log p'(g|\tilde{s}) + \log p'(g) \right]  \nonumber \\
&= \mathbb{E}_{p_{m}^*(\cdot)}\left[ \log \left( \frac{p(\omega, \tilde{s})}{p(\tilde{s}) p(\omega)} \right)   - \log \left( \frac{p'(g, \tilde{s})}{p'(\tilde{s}) p'(g)} \right) \right] \nonumber \\
&= \mathbb{E}_{p_{m}^*(\cdot)}\left[ \log p(\tilde{s}|\omega) - \log p'(\tilde{s}|g) \underbrace{-\log p(\tilde{s}) + \log p'(\tilde{s})}_{=0} \right] \nonumber \\
&= \mathbb{E}_{p_{m}^*(\cdot)}\left[ \log p(\tilde{s}|\omega; \mu^*, \kappa, \theta) - \log p(\tilde{s}|g; \theta) \right] \nonumber \\
&= \mathbb{E}_{p_{m}^*(\cdot)}\left[ \log p(s,g,\tilde{s}|\omega; \mu^*, \kappa, \theta) - \log p(\tilde{s}|g; \theta) \right] \text{\textcolor{gray}{\quad \quad \# $\pi_{\mu^*}$ is deterministic }} \nonumber \\
&= \mathbb{E}_{p(\omega, s, g; \mu^*, \kappa)}\left[ \log p(s|\omega; \mu^*) + \log p(g|s; \kappa) \right] \nonumber\\
&= 0. \text{\textcolor{gray}{\quad \quad \quad \quad \quad \quad \quad \quad \quad \quad \quad \quad \quad \quad \quad \quad \quad \quad \quad \quad \# $\pi_{\mu^*}$ is deterministic}} \nonumber
\end{align}

\textbf{General case:}
If $\mathbb{E}_\omega \left[  p(g|\omega;\mu^*,\kappa) \right] \neq p'(g)$, e.g., $p_{m}^*(\cdot) \triangleq p(\omega, s; \mu^*) p(g, \tilde{s}|s; \kappa, \theta) \neq p(\omega, s; \mu^*) p'(g,\tilde{s}; \theta)$, we apply Holder’s
inequality and Pinsker’s inequality over $\text{TERM-II}$. Then the following inequality holds: 
\begin{align}
\text{TERM-II}  
&= 
\mathbb{E}_{p(\omega, s; \mu^*)} 
\left[ 
\mathbb{E}_{p(g, \tilde{s}|s; \kappa, \theta)} \left[ R(g,\tilde{s}) \right] - 
\mathbb{E}_{p'(g, \tilde{s}; \theta)} \left[ R(g,\tilde{s}) \right]
\right]
\nonumber \\
&\leq 
\mathbb{E}_{p(\omega, s; \mu^*)} 
\left[ 
\Vert R(g,\tilde{s}) \Vert_\infty \cdot \Vert p(g, \tilde{s}|s; \kappa, \theta) - p'(g, \tilde{s}; \theta) \Vert_1
\right]
\nonumber \\
&\leq 
\mathbb{E}_{p(\omega, s; \mu^*)} 
\left[ 
\left( \max_{g, \tilde{s}} R(g,\tilde{s}) \right) \cdot 2 \sqrt{\frac{1}{2} D_{\text{KL}}(p(g,\tilde{s}|s;\kappa,\theta) \Vert p'(g,\tilde{s};\theta)) } 
\right] 
\nonumber \\
&=
\mathbb{E}_{p(\omega, s; \mu^*)} 
\left[ 
\left( \max_{g, \tilde{s}} R(g,\tilde{s}) \right) \cdot 2 \sqrt{\frac{1}{2} D_{\text{KL}}(p(g|s;\kappa) \Vert p'(g)) } 
\right]
, \nonumber
\end{align}
where $R(g,\tilde{s}) = \log p'(g|\tilde{s}) - \log p'(g)$. 

Assuming 
the goal space is the same as the state space and relabeling $f_\kappa$ is bijective, we can ignore the relabeling term $p(g|s;\kappa)$ and replace $p'(g)$ with $p'(s)$. Thus, we have 
$$
\text{TERM-II} \leq \left( \max_{g, \tilde{s}} R(g,\tilde{s}) \right) \cdot  \mathbb{E}_{p(\omega)} \left[ 2 \sqrt{\frac{1}{2} D_{\text{KL}}(p(s|\omega;\mu^*) \Vert p'(s)) } \right]. 
$$

\section{Implementation Details}
\label{appendix_implementation_dddDetails}

\subsection{Relabeling over the Environment} 
\label{appendix-relabeling-on-the-env}
Applying our GPIM on 2D navigation (\emph{color-shape goal}) may cause that the color and shape of the goal that relabeled from a state that does not represent a valid goal in the reset environment (since valid goals are only ones that can be reached by the agent reach \emph{after} this reset). In Algorithm~1 (in the main text), the goals are claimed to be generated by $\pi_\mu$ while there may be no corresponding target in the reset environment for training $\pi_\theta$. Thus, at the step that we reset the environment for training $\pi_\theta$ (\emph{line 14}), we make this reset is associated with the generated goal by $\pi_\mu$ and $f_\kappa$. See Algorithm~\ref{alg:algorithm2-relabel-on-env} for the details.

\begin{algorithm}[h]
	\caption{Learning process of our proposed GPIM \textcolor{red}{(with relabeling over the environment)}}
	\small  
	\label{alg:algorithm2-relabel-on-env}
	\begin{multicols}{2}
		\begin{algorithmic}[1]
			\WHILE {not converged}
			\STATE \textcolor{gray}{\emph{\# Step I:} generate goals and reward functions.}
			\STATE Sample the latent variable: ${\omega} \sim p({\omega})$. 
			\STATE Reset Env. \& sample initial state: ${s}_0 \sim p_0({s})$.
			\FOR { $t = 0, 1, ..., T-1$ steps }
			\STATE Sample action: $ {a}_t \sim \pi_\mu ({a}_t | {s}_t, {\omega}) $. 
			\STATE Step environment: $ {s}_{t+1} \sim p({s}_{t+1} | {s}_t,{a}_t)$.
			\STATE Relabel: ${g}_{t} = f_\kappa({s}_{t+1})$. \ \ \ \  \textcolor{gray}{\emph{$\rhd$ Record.}} 
			\STATE Compute reward $r_t$ for policy $\pi_\mu$ using (5).
			\STATE Update{\scriptsize ~}policy{\scriptsize ~}$\pi_\mu${\scriptsize ~}to{\scriptsize ~}maximize{\scriptsize ~}$r_t${\scriptsize ~}with{\scriptsize ~}SAC.
			\STATE Update  discriminator ($q_\phi$) to maximize $\log q_\phi({\omega}|{s}_{t+1})$ with SGD.
			\ENDFOR
			
			\STATE \textcolor{gray}{\emph{\#{\scriptsize ~}Step{\scriptsize ~}II:}{\scriptsize ~}$\pi_\theta${\scriptsize ~}imitates{\scriptsize ~}$\pi_\mu${\scriptsize ~}with{\scriptsize ~}the{\scriptsize ~}relabeled{\scriptsize ~}goals and the associated rewards (for the same $\omega$). Here we only consider the static goal for~$\pi_\theta$.}
			\STATE \textcolor{red}{Reset Env. with the relabeled goals $g_T$} \& sample initial state: $\tilde{s}_0 \sim p_0(\tilde{s})$.
			\FOR { $t = 0, 1, ..., T-1$ steps }
			\STATE $g_t = g_T$ for fixed (\emph{static}) goals.  
			\STATE Sample action: $ {a}_t \sim \pi_\theta ({a}_t | \tilde{s}_t, {g}_{t})$.
			\STATE Step environment: $ {\tilde{s}}_{t+1} \sim p({\tilde{s}}_{t+1} | {\tilde{s}}_t,{a}_t)$.
			\STATE Compute reward $\tilde{r}_t$ for policy $\pi_\theta$ using (8).
			\STATE Update policy $\pi_\theta$  to maximize $ \tilde{r}_t$ with SAC. 
			\ENDFOR
			\ENDWHILE
		\end{algorithmic}
	\end{multicols}
\end{algorithm}


\subsection{Environment Details}
\label{appendix-tasks}

We introduce the details of environments and tasks here, including the environment setting of 2D navigation (\emph{x-y goal} and \emph{color-shape goal}), object manipulation, three atari games (seaquest, berzerk and montezuma revenge), and the mujoco tasks (swimmer, half cheetah and fetch).  

\textbf{2D navigation tasks:} 
In 2D navigation tasks, the agent moves in each of the four cardinal directions, where the states denote the 2D location of the agent. 
We consider the following two tasks: moving the agent to a specific coordinate named \emph{x-y goal}
and moving the agent to a specific object with certain color and shape named  \emph{color-shape goal}. 

\begin{itemize}
	\item \textbf{2D Navigation (\emph{x-y goal}):}
	The size of the environment is $10 \times 10$ (continuous state space) or $7 \times 7$ (discrete state space). The state is the location of the agent, and the goal is the location of the final location. 
	
	\item \textbf{2D Navigation (\emph{color-shape goal}):} The size of the environment is $10 \times 10$. The state consists of the locations of the agent and three objects with different color-shape pairs (one real target and two distractors). The goal is described by the color and shape of the real target, encoded with one-hot. 
\end{itemize}

\textbf{Object manipulation:} 
More complex manipulation considers a moving agent in 2D environment with one block for manipulation, and the other as a distractor. 
The agent first needs to reach the block and then move the block to the given location, where the block is described using color and shape. 
The size of the environment is $10 \times 10$. The state consists of the locations of the agent and two blocks with different color-shape pairs (one real target and one distractor). The goal consists of the one-hot encoding of the color-shape of the target block that needs to be moved, and the 2D coordinate of the final location of the movement. 

\begin{table}[htbp]
	\centering
	\vspace{-5pt}
	\caption{The repetition length of the action.}
	\begin{tabular}{l l}
		\toprule
		Environments & k \\
		\midrule
		Seaquest-ram-v0 & {2, 3, 4, 5} \\
		Berzerk-ram-v0 & {34, 36, 38, 40} \\
		MontezumaRevenge-ram-v0 & {2, 3, 4, 5} \\
		\bottomrule 
	\end{tabular}%
	\label{tab:ataris}%
	\vspace{-5pt}
\end{table}%

\textbf{Atari games:} 
We test the performance on three atari games: seaquest, berzerk, and montezuma revenge. 
In order to reduce the difficulty of training, we adopt the RAM-environment (i.e., Seaquest-ram-v0, Berzerk-ram-v0, and MontezumaRevenge-ram-v0), where each state represents a 128-dimensional vector. Each action repeatedly performs for a duration of k frames, where k is uniformly sampled from Table~\ref{tab:ataris}. 

\textbf{Mujoco tasks:} 
We consider to make diverse agents to fast imitate a given goal trajectory, including the imitation of behaviors of a swimmer, a half cheetah, and a fetch, where states in the trajectory denote positions of agents. 
Such experiments are conducted to demonstrate the effectiveness of our proposed method in learning behaviors over a continuous high-dimensional action space, which is more complicated in physics than the 2D map.

Note that the goals in all the experiments are images $50 \times 50 \times 3$ (3 channels, RGB) in size, except that the \emph{color-shape goal} is encoded with one-hot.

\subsection{Tracking Hand Movement}
\label{appendix-tracking-hand}
\begin{wrapfigure}{r}{0.30\textwidth}
	\centering
	\vspace{-33pt}
	\begin{center}
		\includegraphics[scale=0.23]{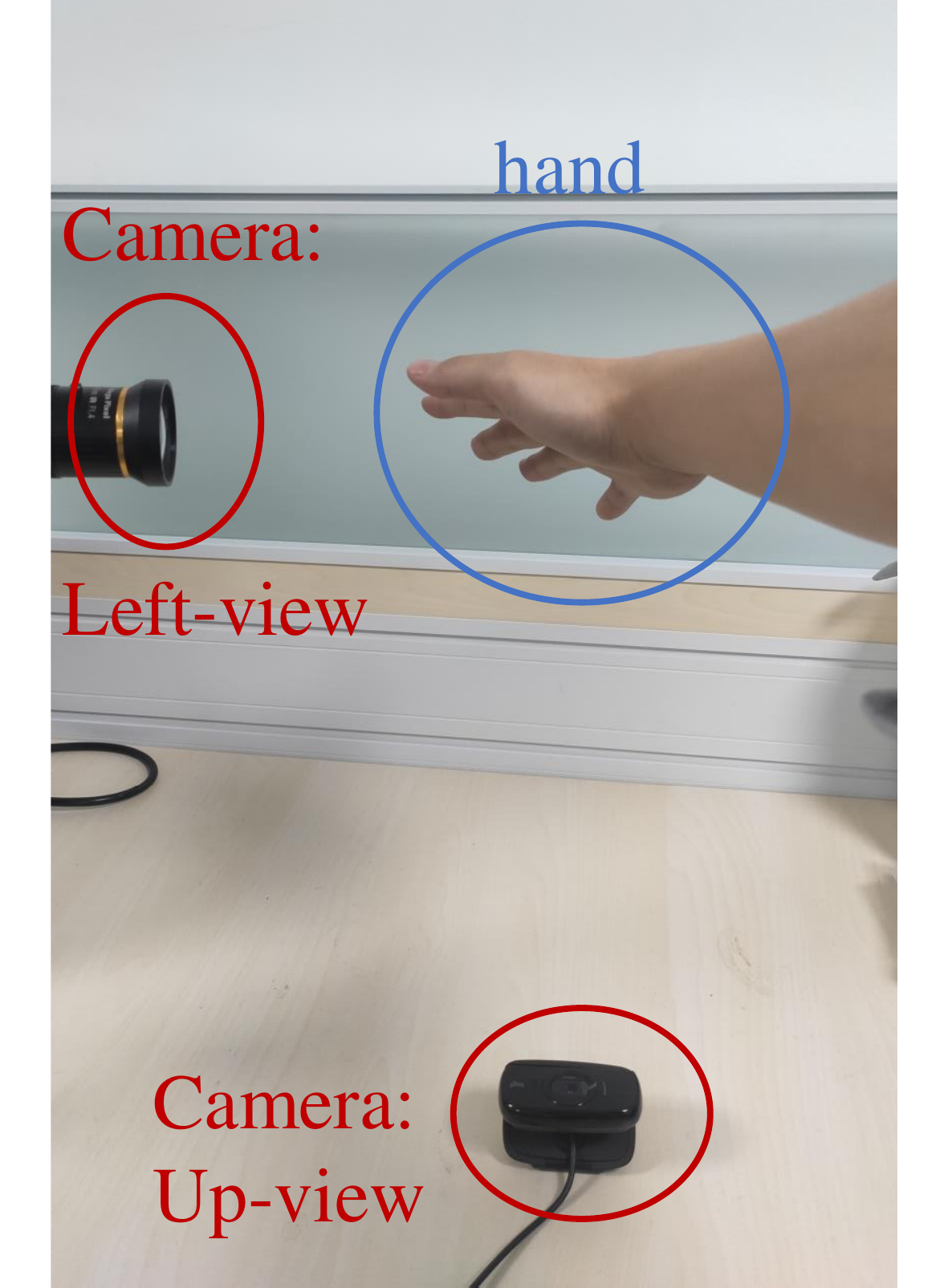}
	\end{center}
	\caption{We adopt~two cameras to~locate the hand for the movement tracking task.}
	\label{appendix-camera-hand}
	\vspace{-23pt}
\end{wrapfigure}
For the dynamic (time-varying) goals, we implement the hand movement tracking with MediaPipe~\citep{campillo2019mediapipe} to extracting features of hand from images. As shown in Figure~\ref{appendix-camera-hand}, we adopt two cameras to get the up-view and the left-view of hand, respectively. 
Then the two views are processed by MediaPipe to locate the hand, which allows us to get the position in three dimensions. Finally, we make the panda robot track the extracted 3D coordinate with the trained goal-conditioned policy $\pi_\theta$.

\subsection{Metrics, Network Architectures and Hyperparameters}
\label{appendix-hyperparameters}
Here we give a clear definition of our evaluation metric -- "normalized distance to goal":

\textbf{I:} 
When the goal is to reach the final state of the trajectory induced by $\pi_\mu$, the distance to goal is the L2-distance between the final state $\tilde{s}_{T}^k$ induced by $\pi_\theta(\cdot|\cdot, g^k)$ and the goal state $g^k$ randomly sampled from the goal (task) space: 
$$\textit{Dis} = \frac{1}{N} \sum_{k=1}^{N} L2(\tilde{s}_{T}^k, g^k),$$ 
where $N$ is the number of testing samples. We set $N = 50$ for 2D navigation, object manipulation and atari games (seaquest, berzerk and montezuma revenge). 

\textbf{II:} 
When the goal is to imitate the whole trajectory induced by $\pi_\mu$, the distance is the expectation of distance over the whole trajectory $\{\tilde{s}_0^k, \tilde{s}_1^k, ..., \tilde{s}_{T}^k\}$ induced by $\pi_\theta(\cdot|\cdot,g_t^k)$ and goal trajectory $\{g_0^k,g_1^k,...,g_{T-1}^k\}$ randomly sampled from the trajectory (task) space: 
$$\textit{Dis} = \frac{1}{N} \sum_{k=1}^{N}\left( \frac{1}{T} \sum_{t=1}^{T}L2(\tilde{s}_t^k, g_{t-1}^k) \right) ,$$ 
where $N$ is the number of testing samples. We set $N = 50$ for mujoco tasks (swimmer, half cheetah and fetch) . 

The term "normalized" means that the distance is divided by a scale factor. 

Note that, for three atari games (seaquest, berzerk, and montezuma revenge), the L2-distance for evaluation\footnote{The reward function for our baseline \textbf{L2 Distance} still calculates the L2-distance directly on the original state space, instead of the distance of the agents' positions after pixel matching here. } is the difference between the position of controllable agent and the target's position, where the position is obtained by matching the pixel on the imaged state.

In our implementation, we use two independent SAC architectures \citep{haarnoja2018soft} for latent-conditioned policy $\pi_\mu$ and goal-conditioned policy $\pi_\theta$. We find empirically that having two networks share a portion of the network structure will degrade the experimental performance. 
We adopt universal value function approximates (UVFAs)~\citep{schaul2015universal} for extra input (goals). 
For the latent-conditioned policy $\pi_\mu$, to pass latent variable $\omega$ to the Q function, value function and policy, as in DIAYN, we simply concatenate $\omega$ with the current state $s_t$ (and action $a_t$). 
For goal-conditioned policy $\pi_\theta$, we also concatenate  $g_t$ with current state $\tilde{s}_t$ (and action $a_t$). 
We update the $\vartheta_E$ using the gradients from both the \emph{Dis\_loss} and Q function's loss of the goal-conditioned policy $\pi_\theta$.  

The latent-distributions are provided in Table~\ref{tab:addlabel-new-laent-distribution} and the hyper-parameters are presented in Table~\ref{hyperparameters}.


\begin{table}[htbp]
	\centering
	\caption{The given latent distribution for each tasks in our experiments.}
	\begin{tabular}{llrr}
		\toprule
		Arrow & 2D navigation & \multicolumn{1}{l}{Object manipulation} & \multicolumn{1}{l}{Atari games} \\
		Discrete; 3 skills & Continuous; [-1, 1]$^2$ & \multicolumn{1}{l}{Continuous; [-1, 1]$^2$} & \multicolumn{1}{l}{Continuous; [-1, 1]$^2$} \\
		\midrule
		Half-cheetah & Fetch; Panda & \multicolumn{1}{l}{Maze (Figure 8)} & \multicolumn{1}{l}{2D navigation (Appendix 1.4)} \\
		Discrete; 50 skills & Continuous; [-1, 1]$^3$ & \multicolumn{1}{l}{Discrete; 10 skills} & \multicolumn{1}{l}{Discrete; 10 skills} \\
		\midrule
		Swimmer & Gridworld &       &  \\
		Discrete; 50 skills & Discrete; 4 skills &       &  \\
		\bottomrule
	\end{tabular}%
	\label{tab:addlabel-new-laent-distribution}%
\end{table}%

\begin{table*}[h]
	\centering
	\caption{Hyper-parameters}
	\begin{tabular}{lll c}
		\toprule
		\multicolumn{3}{c }{Hyper-parameter} & value \\
		\midrule
		\multicolumn{3}{c }{Batch Size} & 256 \\
		\multicolumn{3}{c }{Discount Factor} & 0.99 \\
		\multicolumn{3}{c }{Buffer Size} & 10000 \\
		\multicolumn{3}{c }{Smooth coefficient} & 0.05 \\
		\multicolumn{3}{c }{Temperature} & 0.2 \\
		\midrule
		\multirow{9}[8]{*}{Learning Rate} & \multirow{2}[2]{*}{2D Navigation} & x-y goal & 0.001 \\
		&       & color-shape goal & 0.001 \\
		\cmidrule{2-4}          & \multicolumn{2}{l }{Object Manipulation} & 0.001 \\
		\cmidrule{2-4}          & \multirow{3}[2]{*}{Mujoco tasks} & Swimmer & 0.0001 \\
		&       & HalfCheetah & 0.0001 \\
		&       & Fetch & 0.0001 \\
		\cmidrule{2-4}          & \multirow{3}[2]{*}{Atari games} & Seaquest & 0.0003 \\
		&       & Berzerk & 0.0003 \\
		&       & Montezuma Revenge & 0.0003 \\
		\midrule
		\multirow{9}[7]{*}{Path Length} & \multirow{2}[2]{*}{2D Navigation} & x-y goal & 20 \\
		&       & color-shape goal & 20 \\
		\cmidrule{2-4}          & \multicolumn{2}{l }{Object Manipulation} & 20 \\
		\cmidrule{2-4}          & \multirow{3}[2]{*}{Mujoco tasks} & Swimmer & 50 \\
		&       & HalfCheetah & 50 \\
		&       & Fetch & 100 \\
		\cmidrule{2-4}          & \multirow{3}[1]{*}{Atari games} & Seaquest & 25 \\
		&       & Berzerk & 25 \\
		&       & Montezuma Revenge & 25 \\
		\midrule
		\multirow{9}[7]{*}{Hidden Size} & \multirow{2}[1]{*}{2D Navigation} & x-y goal & 128 \\
		&       & color-shape goal & 128 \\
		\cmidrule{2-4}          & \multicolumn{2}{l }{Object Manipulation} & 128 \\
		\cmidrule{2-4}          & \multirow{3}[2]{*}{Mujoco tasks} & Swimmer & 256 \\
		&       & HalfCheetah & 256 \\
		&       & Fetch & 256 \\
		\cmidrule{2-4}          & \multirow{3}[2]{*}{Atari games} & Seaquest & 256 \\
		&       & Berzerk & 256 \\
		&       & Montezuma Revenge & 256 \\
		\midrule
		\multicolumn{1}{c}{\multirow{9}[8]{*}{Dimension of Generative Factor}} & \multirow{2}[2]{*}{2D Navigation} & x-y goal & 2 \\
		&       & color-shape goal & 2 \\
		\cmidrule{2-4}          & \multicolumn{2}{l }{Object Manipulation} & 4 \\
		\cmidrule{2-4}          & \multirow{3}[2]{*}{Mujoco tasks} & Swimmer & 3 \\
		&       & HalfCheetah & 4 \\
		&       & Fetch & 4 \\
		\cmidrule{2-4}          & \multirow{3}[2]{*}{Atari games} & Seaquest & 16 \\
		&       & Berzerk & 16 \\
		&       & Montezuma Revenge & 16 \\
		\midrule
		\multicolumn{3}{c }{$\alpha$} & 1 \\
		\multicolumn{3}{c }{$\gamma$} & 5 \\
		\multicolumn{3}{c }{$\delta_{pixel}$} & 255 \\
		\bottomrule
	\end{tabular}%
	\label{hyperparameters}%
\end{table*}%

\section{More Results}
\label{appendix-more-results}

\subsection{Comparison on the Atari tasks}

\begin{figure*}[h]
	\centering
	\includegraphics[scale=0.365]{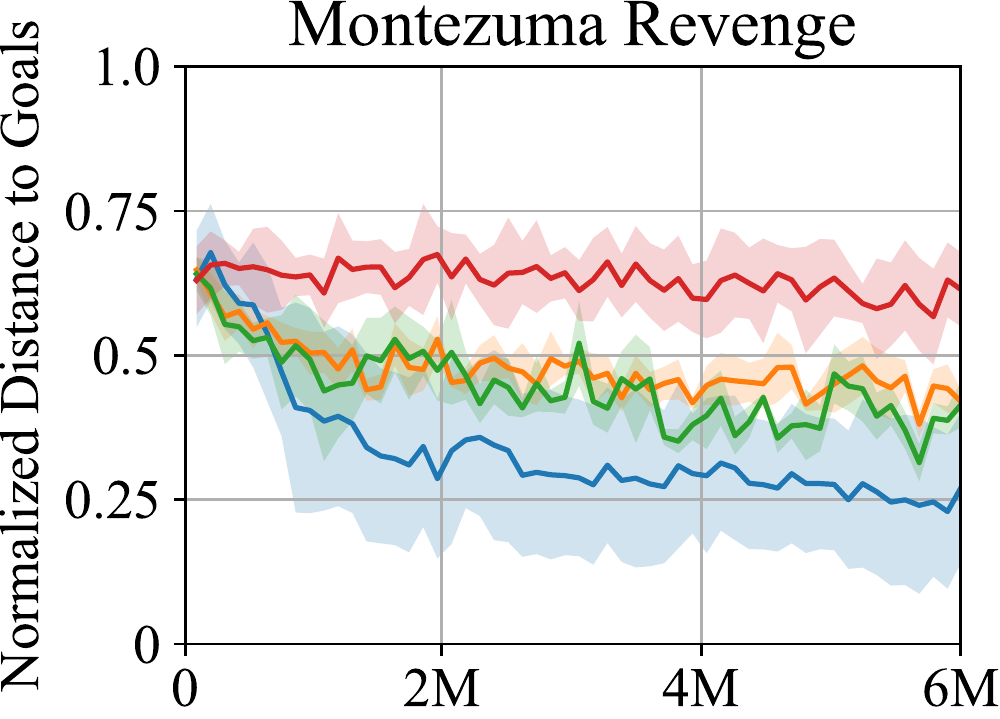}
	\includegraphics[scale=0.365]{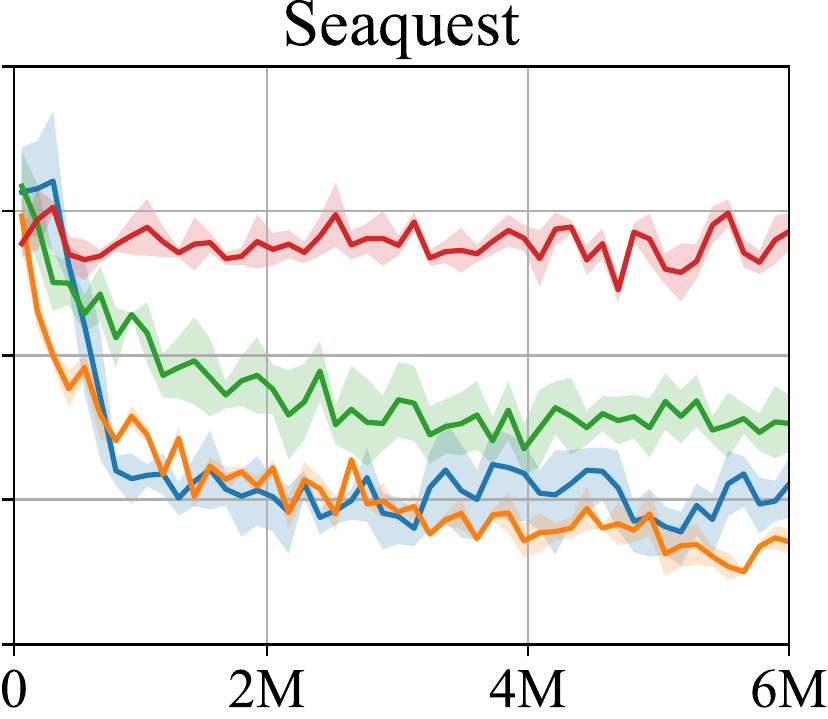}
	\includegraphics[scale=0.365]{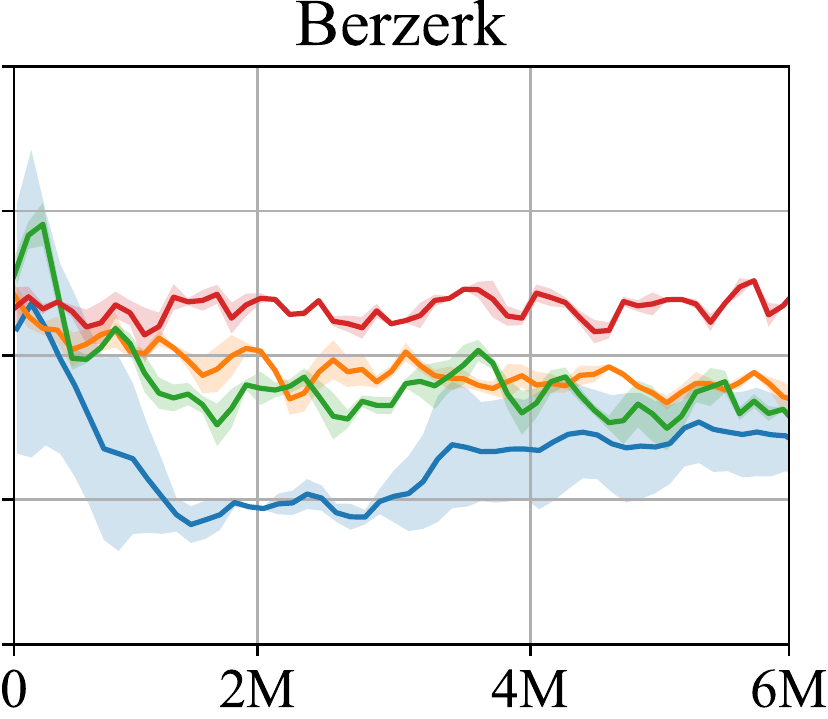}
	\includegraphics[scale=0.321]{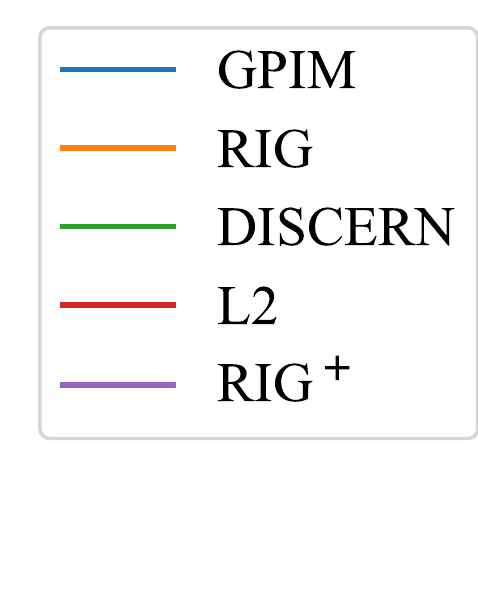}
	\caption{Performance (normalized distance to goals vs. actor steps) of our GPIM and baselines.}
	\label{app:additional:fig-exp3-Atari}
\end{figure*}

In Figure~\ref{app:additional:fig-exp3-Atari}, we provide more results on the  Atari tasks.

\subsection{Learned Behaviors on temporally-extended tasks}
\label{appendix-temporally-extended}

\begin{figure}[H]
	\centering
	\includegraphics[scale=0.53]{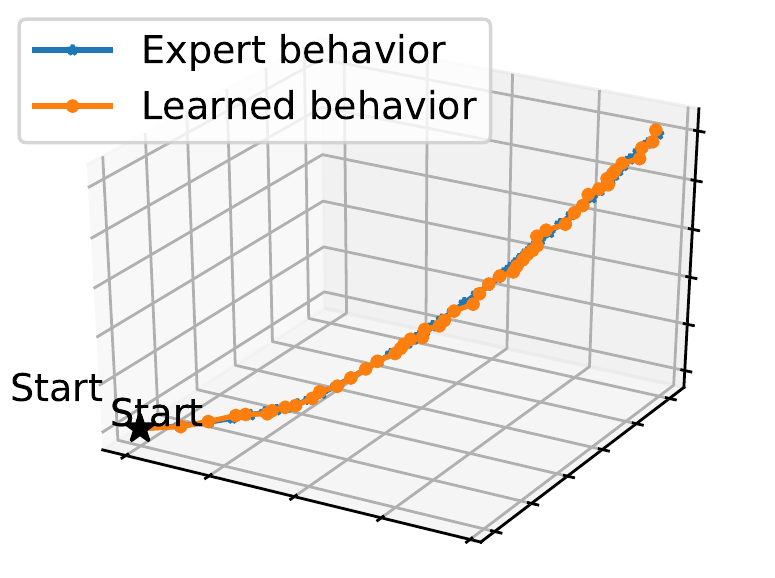}
	\includegraphics[scale=0.53]{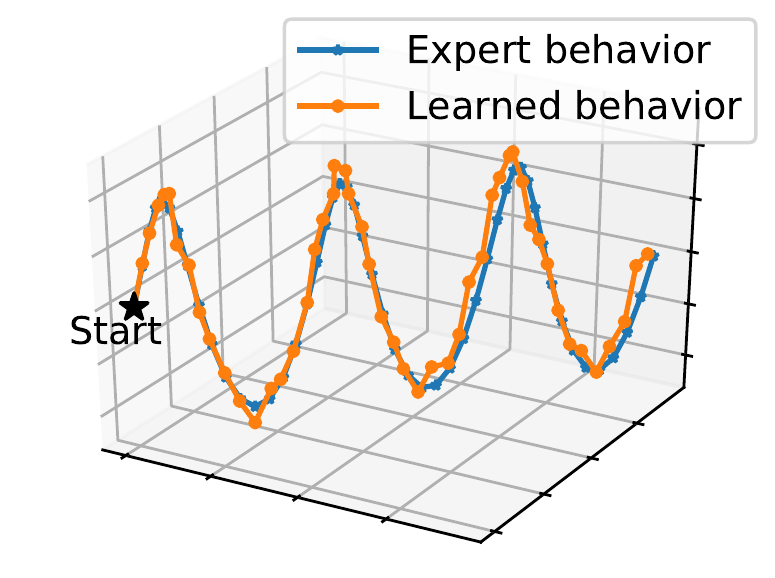}
	\\
	\includegraphics[scale=0.53]{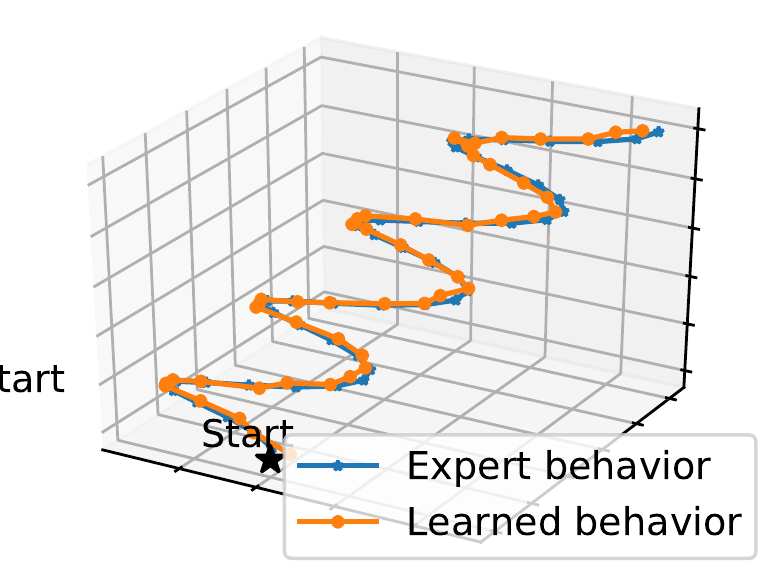}
	\includegraphics[scale=0.53]{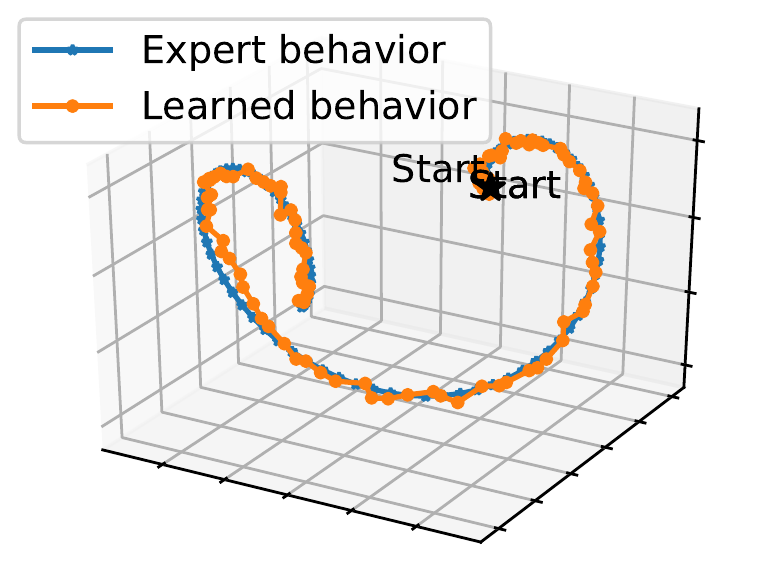}
	\caption{Expert behaviors and learned behaviors. The four expert trajectories are described parametrically as: $(x, y, z) = (\log_{10}(t+1)+t/50, \sin(t)/5+t/5, t/5)$, $(x, y, z) = (t/5, \cos(t)/5-1/5+t/5, \sin(t)/5)$, $(x, y, z) = (\cos(t)/5+t/50-1/1, \sin(t)/5+t/5, t/5)$, and $(x, y, z) = (\sin(t)-sin(2t)/2, -t/5, \cos(t)/2-cos(2t)/2)$. }
	\label{app-fetch-temporal}
\end{figure}

More experimental results are given in Figure~\ref{app-fetch-temporal} to show the imitation on several temporally-extended tasks.  

\subsection{Learned Behaviors from GPIM}
\label{appendix-behaviors}

\begin{figure*}[h!]
	\centering
	\subfigure[2D navigation (discrete x-y goal; continous x-y goal; color-shape goal)]{
		\begin{minipage}{0.99\linewidth}
			\centering
			\includegraphics[scale=0.08]{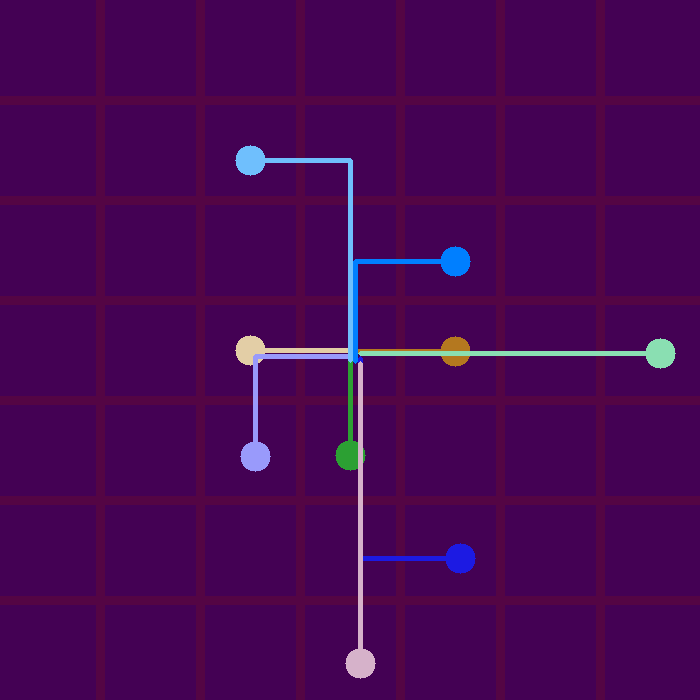}
			\includegraphics[scale=0.08]{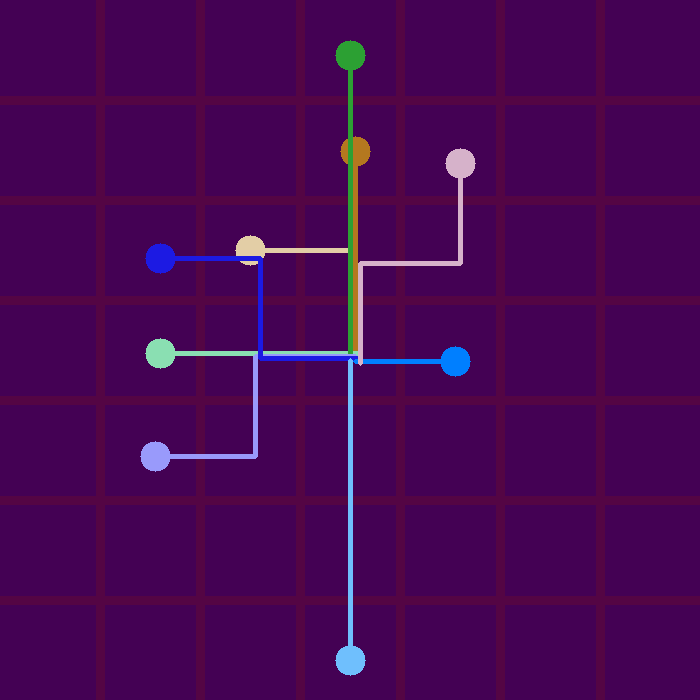}
			\includegraphics[scale=0.056]{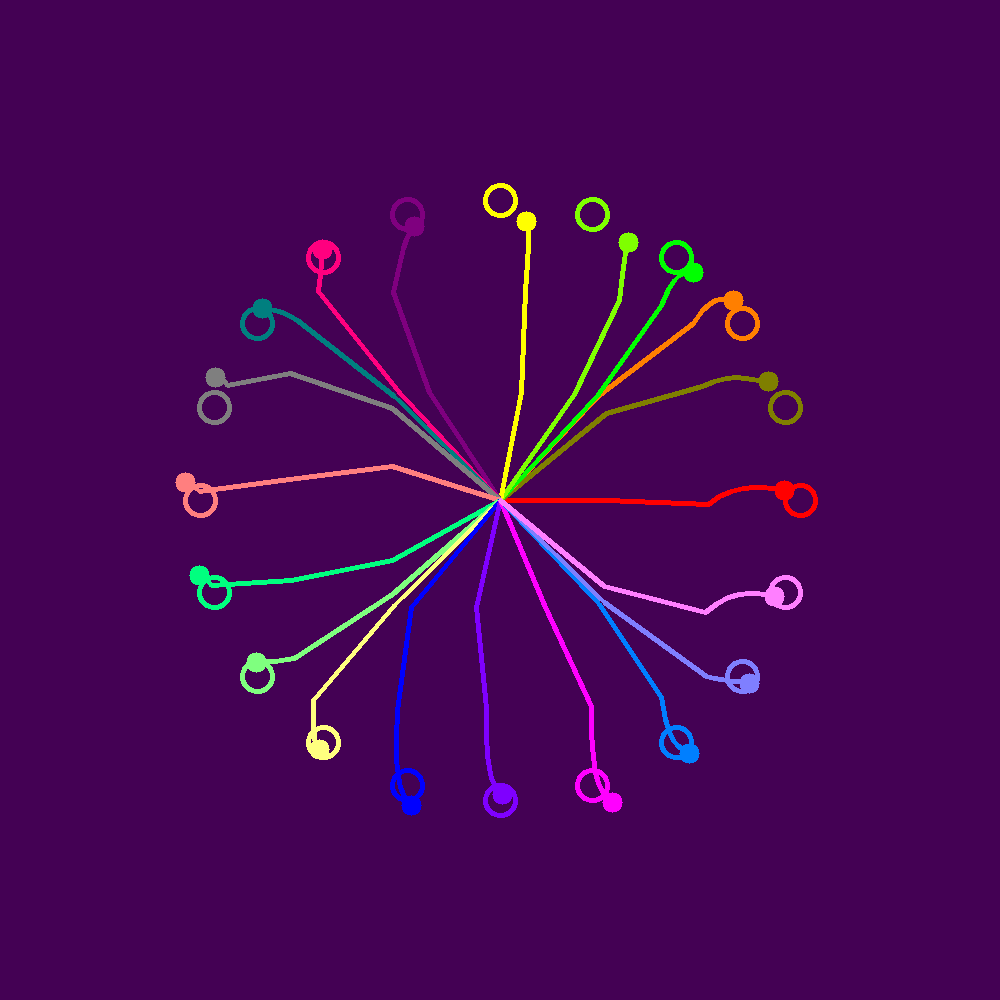}
			\includegraphics[scale=0.056]{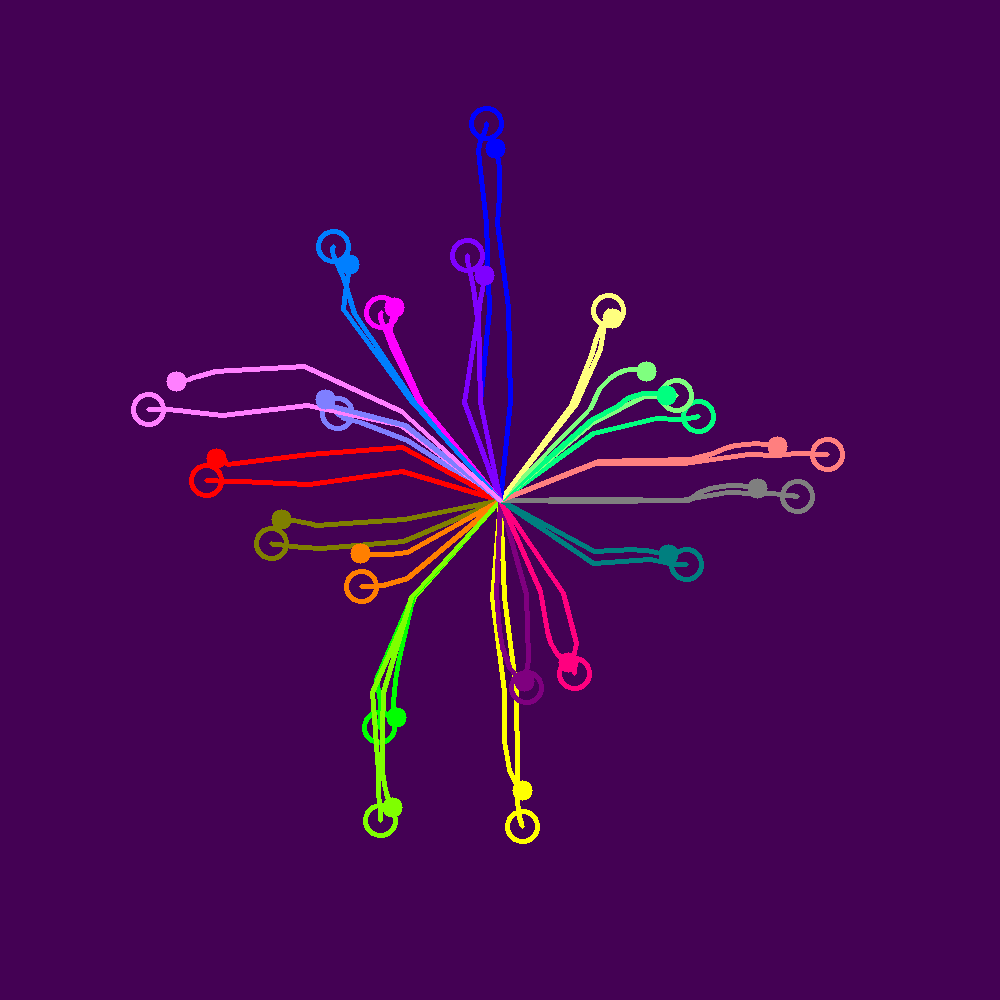}
			\includegraphics[scale=0.22]{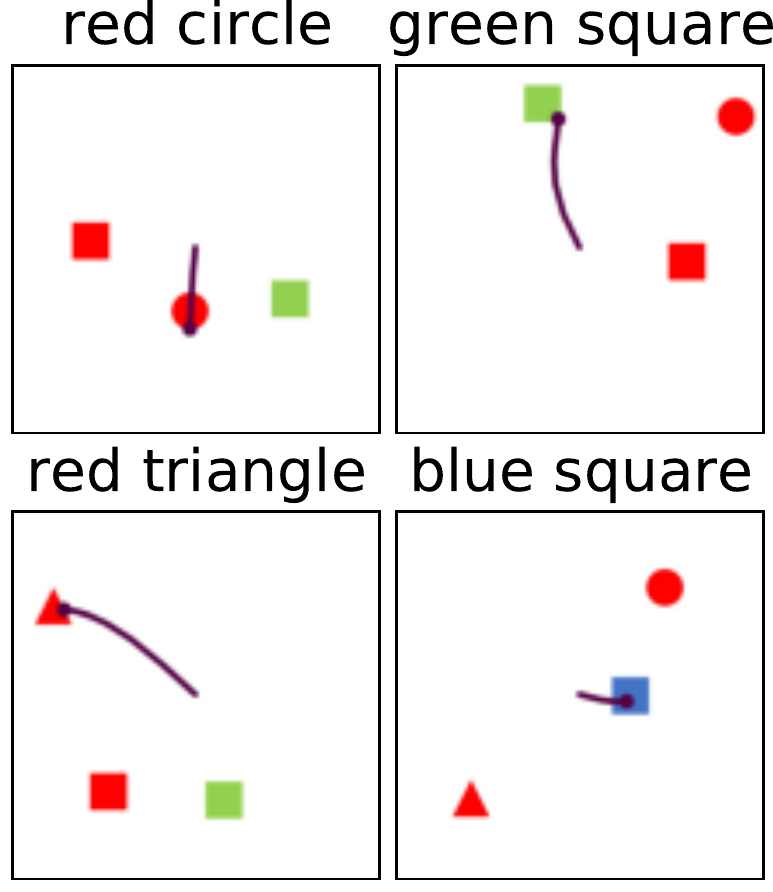}
			\includegraphics[scale=0.22]{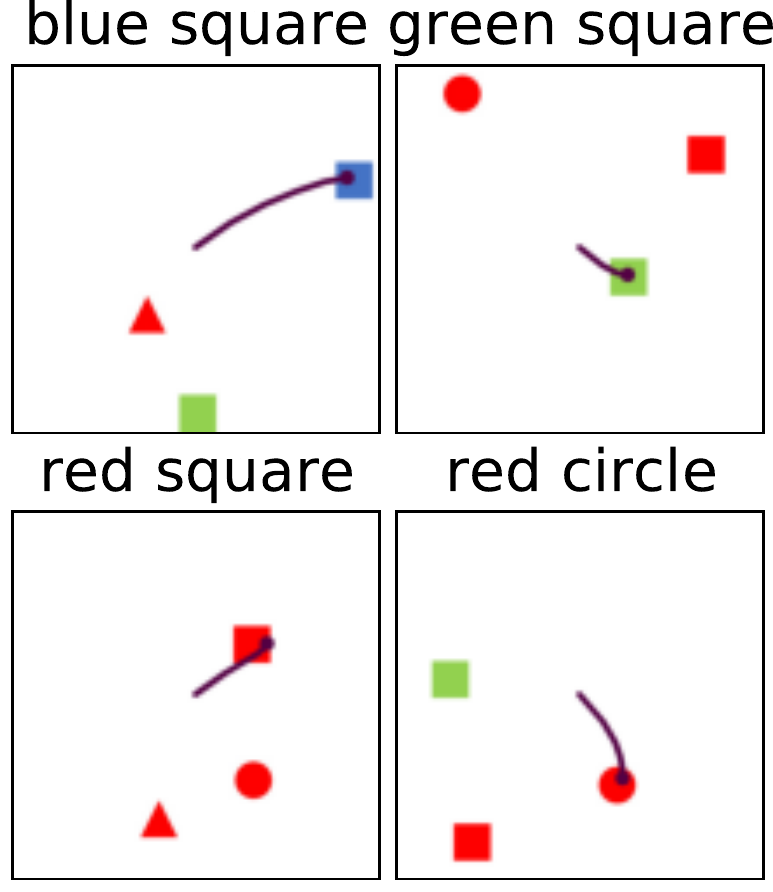}
		\end{minipage}
		\label{app-4-1}
	}
	\subfigure[Object manipulation]{
		\begin{minipage}{0.99\linewidth}
			\centering
			\includegraphics[scale=0.22]{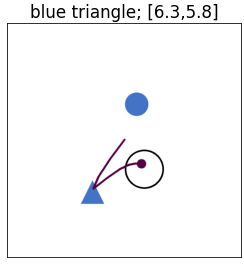}
			\includegraphics[scale=0.22]{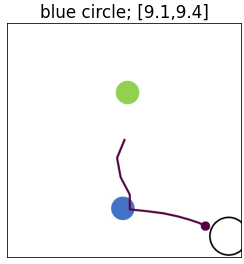}
			\includegraphics[scale=0.22]{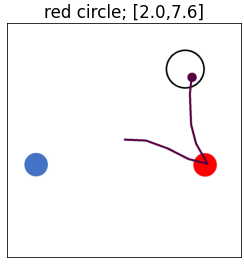}
			\includegraphics[scale=0.22]{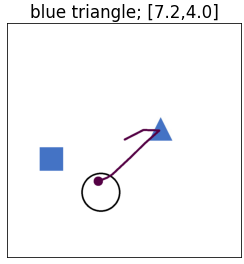}
			\includegraphics[scale=0.22]{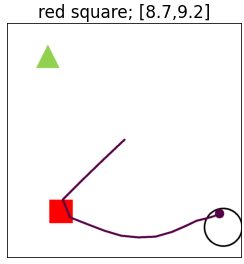}
			\includegraphics[scale=0.22]{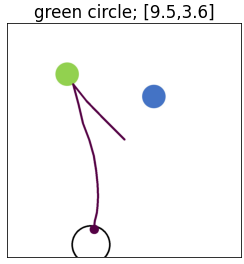}
		\end{minipage}
		\label{app-4-2}
	}
	\subfigure[Swimmer]{
		\begin{minipage}{0.48\linewidth}
			\centering
			\includegraphics[scale=0.18]{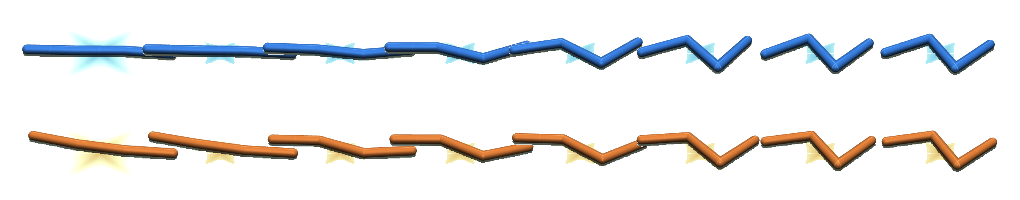}
			\includegraphics[scale=0.18]{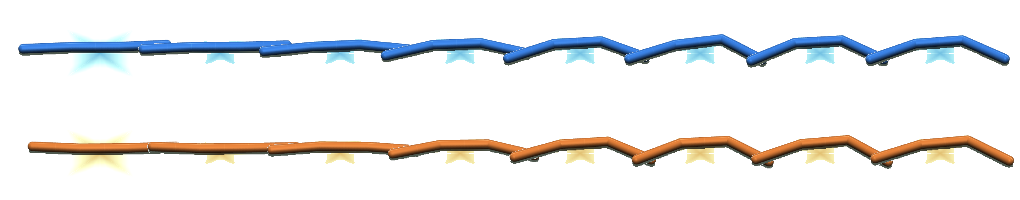}
		\end{minipage}
		\label{app-4-3}
	}
	\subfigure[Half cheetah]{
		\begin{minipage}{0.48\linewidth}
			\centering
			\includegraphics[scale=0.18]{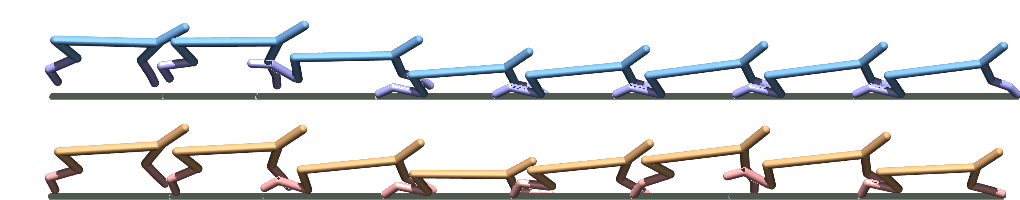}
			\includegraphics[scale=0.18]{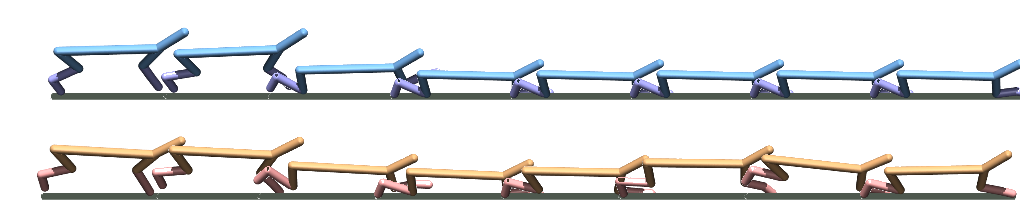}
		\end{minipage}
		\label{app-4-4}
	}
	\subfigure[Fetch]{
		\begin{minipage}{0.90\linewidth}
			\centering
			\includegraphics[scale=0.38]{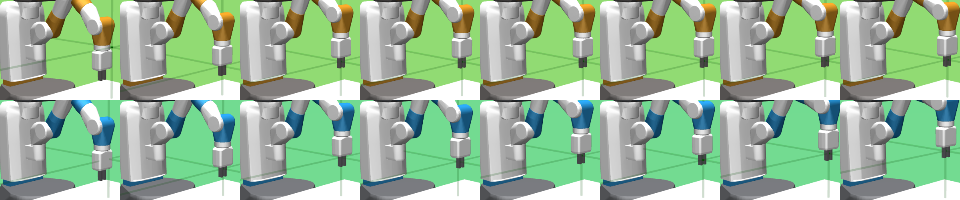}
			\includegraphics[scale=0.38]{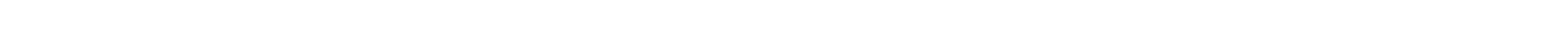}
			\includegraphics[scale=0.38]{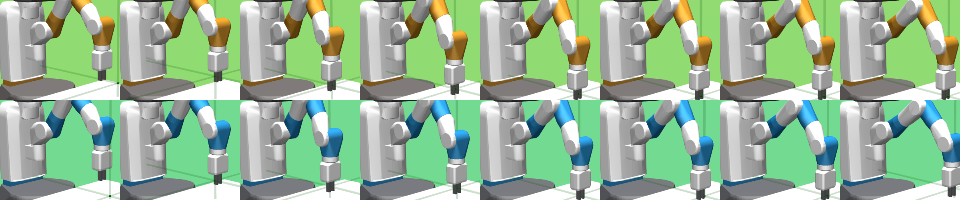}
		\end{minipage}
		\label{app-4-5}
	}
	\subfigure[Seaquest]{
		\begin{minipage}{1.0\linewidth}
			\centering
			\includegraphics[scale=0.19]{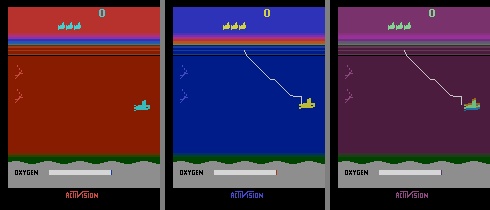}
			\includegraphics[scale=0.19]{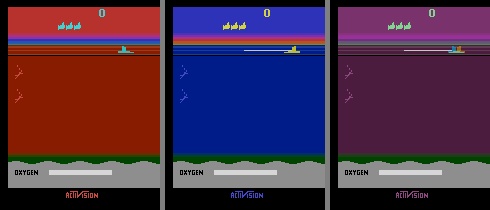}
			\includegraphics[scale=0.19]{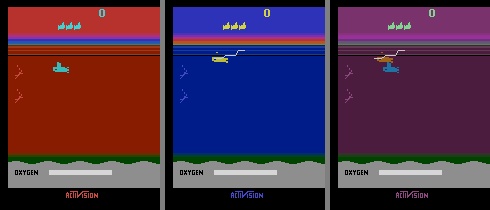}
			\includegraphics[scale=0.19]{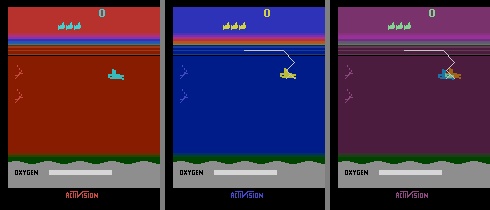}
		\end{minipage}
		\label{app-4-6}
	}
	\subfigure[Berzerk]{
		\begin{minipage}{1.0\linewidth}
			\centering
			\includegraphics[scale=0.19]{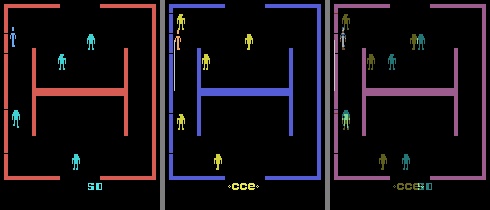}
			\includegraphics[scale=0.19]{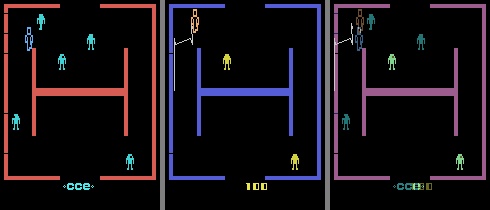}
			\includegraphics[scale=0.19]{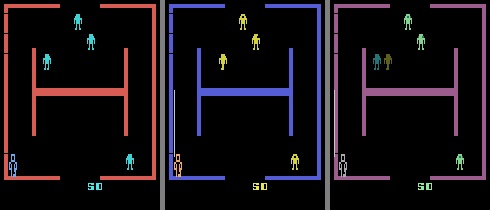}
			\includegraphics[scale=0.19]{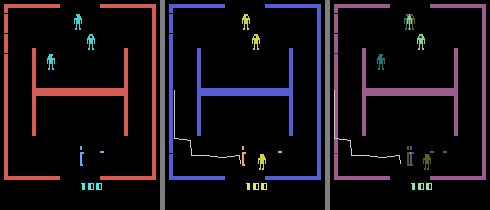}
		\end{minipage}
		\label{app-4-7}
	}
	\subfigure[Montezuma revenge]{
		\begin{minipage}{1.0\linewidth}
			\centering
			\includegraphics[scale=0.19]{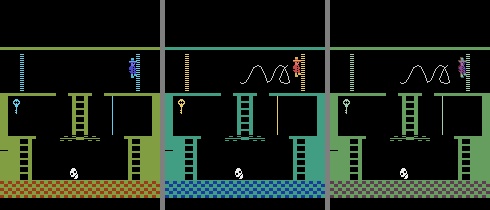}
			\includegraphics[scale=0.19]{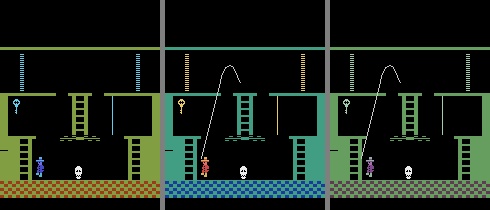}
			\includegraphics[scale=0.19]{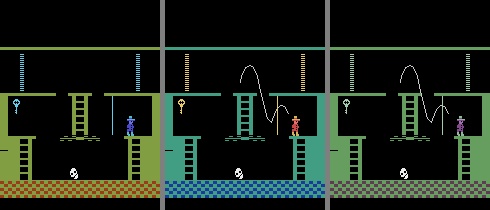}
			\includegraphics[scale=0.19]{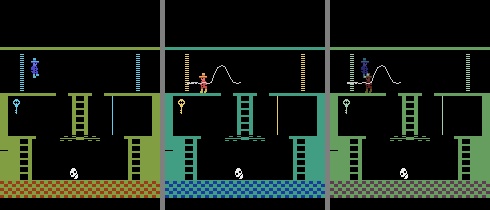}
		\end{minipage}
		\label{app-4-8}
	}
	\caption{Discovered goal-conditioned behaviors. (f-h): The left subfigure shows the expert behaviors (goals); The middle subfigure shows the learned behaviors by GPIM; The right subfigure is the stacked view of goals and GPIM behaviors.}
	\label{app-4}
\end{figure*}

More experimental results are given in Figure~\ref{app-4} to show the learned behaviors on 2D navigation, object manipulation, three atari games (seaquest, berzerk and montezuma revenge), and the mujoco tasks (swimmer, half cheetah and fetch). 
Videos are available under \href{https://sites.google.com/view/gpim}{https://sites.google.com/view/gpim}.

\section*{Broader Impact}

The main concern in the research area of unsupervised RL is how to find numerous diverse goals as well as associated reward functions. 
Particularly, in practical application, the goal and state are often heterogeneous data with high variability of data types and formats, which further aggravates the difficulty of reward learning. 
Our model addresses these problems by introducing a latent-conditioned policy and the relabeling procedure, scoring current states with latent variables, instead of goals. Although the manifold spaces of goals and latent variables are inconsistent, we show that this relabeling procedure can still effectively learn policy.


However, by autonomous exploration of the environment, the agent is likely to generate some useful behaviors as well as the majority of useless skills. 
Of particular concern is that the use of autonomous exploration is likely to generate a strategy that will induce dire consequences, such as a collision skill in an autonomous driving environment. 
How to generate useful behaviors for user-specified tasks and how to use these induced skills are also open problems. 
An alternative solution is to use an extrinsic reward or the offline data (IRL, offline RL) to guide the exploration. While another issue comes from the exploration-exploitation trade-off. 
We would encourage further work to understand the limitations of GPIM interacting with the environment autonomously. We would also encourage research to understand the risks arising from autonomous robot learning. 

Another limitation is that our learning framework needs an extra latent-conditioned policy training for relabeling goals and providing a reward function. This may require twice as much interaction time with the environment as learning a single policy network. For practical application, a simulation platform is preferred for such  expensive training. 
Thus, one promising direction is to mitigate the difference between the simulation platform and the actual environment. We also encourage researchers to pursue the effective methods for transfer learning.

\end{document}